\documentclass{article}

\usepackage{PRIMEarxiv}

\usepackage[utf8]{inputenc} % allow utf-8 input
\usepackage[T1]{fontenc}    % use 8-bit T1 fonts
\usepackage{hyperref}       % hyperlinks
\usepackage{url}            % simple URL typesetting
\usepackage{booktabs}       % professional-quality tables
\usepackage{amsfonts}       % blackboard math symbols
\usepackage{nicefrac}       % compact symbols for 1/2, etc.
\usepackage{microtype}      % microtypography
\usepackage{lipsum}
\usepackage{fancyhdr}       % header
\usepackage{graphicx}       % graphics
\graphicspath{{media/}}     % organize your images and other figures under media/ folder

\usepackage{natbib}
\usepackage{float}
\usepackage{subfig}
\usepackage{amsmath}
\usepackage[linesnumbered, ruled]{algorithm2e}
\usepackage{tabularx}
\usepackage{lineno}
\usepackage{changepage}

\hypersetup{draft=false} % reserve hyper when draft

\newlength{\extralength}
\makeatletter
\newcommand{\appendixtitles}[1]{\gdef\@appendixtitles{#1}} % define \appendixtitles

\makeatother

\title{Dynamics and Control of Vision-Aided Multi-UAV-tethered Netted System Capturing Non-Cooperative Target
\thanks{\textit{\underline{Citation}}: 
\textbf{Authors. Title. Pages.... DOI:000000/11111.}} 
}

\author{
  Runhan Liu \\
  School of Astronautics \\
  Harbin Institute of Technology \\
   \And
  Hui Ren \\
  School of Astronautics \\
  Harbin Institute of Technology \\
   \And
  Wei Fan \\
  School of Astronautics \\
  Harbin Institute of Technology \\
}

\begin{document}
\maketitle

\begin{abstract}
As the number of Unmanned Aerial Vehicles (UAVs) operating in low-altitude airspace continues to increase, non-cooperative targets pose growing challenges to low-altitude operations. To address this issue, this paper proposes a multi-UAV-tethered netted system as a non-lethal solution for capturing non-cooperative targets. To validate the proposed system, we develop mySim, a multibody dynamics-based UAV simulation environment that integrates high-precision physics modeling, vision-based motion tracking, and reinforcement learning-driven control strategies. In mySim, the spring-damper model is employed to simulate the dynamic behavior of the tethered net, while the dynamics of the entire system is modeled using multibody dynamics (MBD) to achieve accurate representations of system interactions. The motion of the UAVs and the target are estimated using VINS-MONO and DETR, and the system autonomously executes the capture strategy through MAPPO. Simulation results demonstrate that mySim accurately simulates dynamics and control of the system, successfully enabling the multi-UAV-tethered netted system to capture both non-propelled and maneuvering non-cooperative targets. By providing a high-precision simulation platform that integrates dynamics modeling with perception and learning-based control, mySim enables efficient testing and optimization of UAV-based control policies before real-world deployment. This approach offers significant advantages for simulating complex UAVs coordination tasks and has the potential to be applied to the design of other UAV-based systems.
\end{abstract}

% keywords can be removed

\section{Introduction}\label{sec:main_intro}

In recent years, the demand for multirotor UAV has steadily increased, making them a prevalent solution in various domains. Single multirotor UAV have been widely applied in crop health monitoring \citep{ecke2022uav}, remote sensing \citep{alvarez2021uav}, disaster relief \citep{daud2022applications}, and battlefield reconnaissance \citep{nohel2023area}. On a larger scale, multi-UAV systems have been utilized for formation flight \citep{dong2018theory}, cooperative path planning \citep{he2019research}, perception and communication \citep{khan2020emerging}, search and mapping operations \citep{queralta2020collaborative}, and more. Additionally, UAVs equipped with robotic arm and other modular attachments have been developed to form multirotor aerial manipulators \citep{kim2016vision,kim2018cooperative}, further expanding their application scope.

\begin{figure}[H]
\centering
% \begin{adjustwidth}{-\extralength}{0cm}
                % \includegraphics[width=8cm]{resources/results/multi-drone_model.jpg}
                \subfloat[\centering]{\includegraphics[width=8cm]{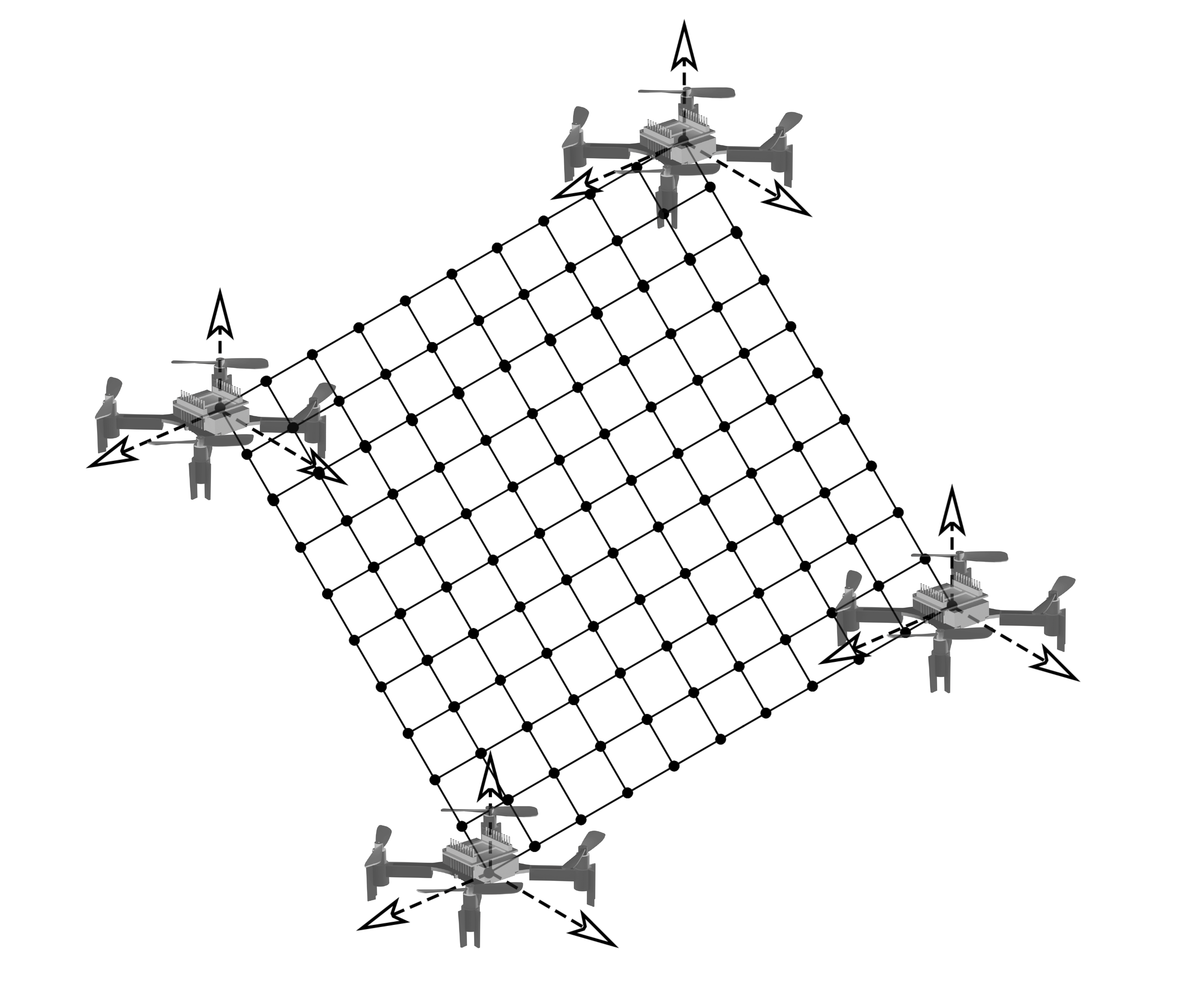}}
                \hfill
                \subfloat[\centering]{\includegraphics[width=8cm]{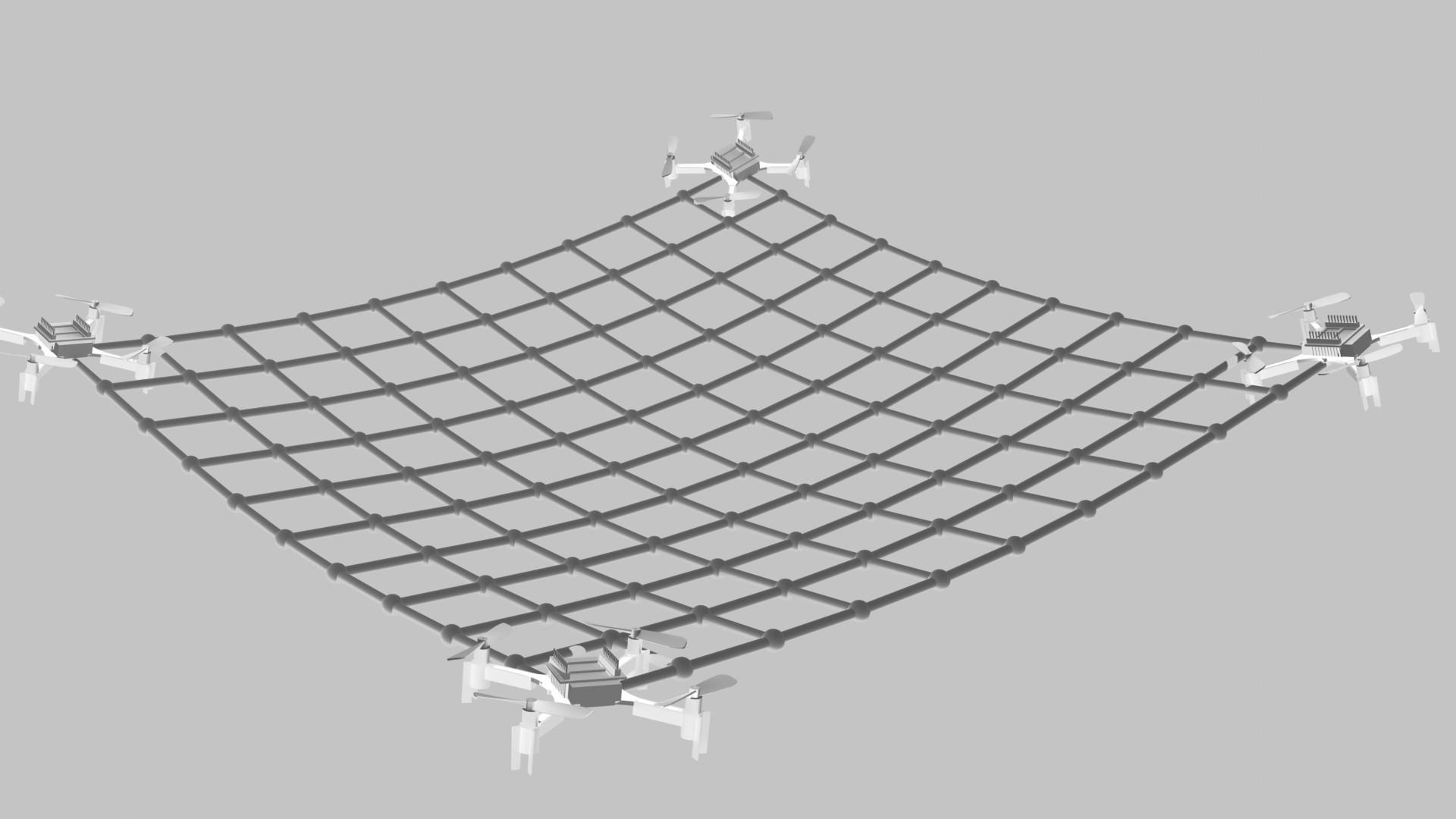}}\\
% \end{adjustwidth}
\caption{Schematic diagram and visualization of the multi-UAV-tethered netted system for the capture task, where the net is composed of rope modules from Fig.~\ref{fig:rope_model}, with UAVs connected to the net via Marker.
\label{fig:multi-drone_vis}}
\end{figure}

As the number of UAVs operating in low-altitude airspace continues to grow rapidly \citep{xu2020recent}, which is increasingly influenced by non-cooperative targets such as birds or other UAVs \citep{chen2019classification}, the need for effective systems to safeguard low-altitude operations has become more pressing. To support the growth of the low-altitude economy, systems capable of capturing or deterring non-cooperative targets without causing harm are required. To address these needs, this paper proposes the multi-UAV-tethered netted system, illustrated in Fig.~\ref{fig:multi-drone_vis}, as a non-lethal means of capturing non-cooperative targets.

When designing new UAV systems or developing UAV-based tasks, establishing a corresponding simulation environment is essential. Several UAV simulators have been developed, each offering unique features and advantages. RotorS \citep{Furrer2016} is a widely used UAV simulator built on the Gazebo \citep{1389727} platform, providing a modular framework for multirotor simulation and control. It has been instrumental in UAV dynamics research and algorithm testing. CrazyS \citep{Silano2018MED,Silano2019ROSVolume4} extends RotorS by tailoring its functionalities for the Crazyflie nano quadcopter, making it particularly useful for research on swarming and control algorithms. PX4-SITL \citep{lorenzmeierPX4PX4AutopilotV11522024}, is a software-in-the-loop simulator that integrates the PX4 autopilot, enabling realistic flight dynamics and sensor simulations. PyBullet-Drones \citep{panerati2021learning}, built on the Bullet physics engine \citep{coumans2021}, provides a lightweight and efficient simulation environment that balances computational efficiency and accuracy, making it suitable for reinforcement learning and control applications. Airsim \citep{shah2018airsim}, developed by Microsoft, leverages Unreal Engine to provide high-precision environments for AI research and autonomous vehicle development. MuJoCo \citep{todorov2012mujoco}, a physics engine known for its efficiency, is frequently employed for UAV simulation tasks that involve articulated mechanisms and complex interactions. Pegasus Simulator \citep{10556959} integrates NVIDIA Omniverse \citep{makoviychuk2021isaac} and Isaac Sim \citep{mittal2023orbit}, offering real-time multirotor simulation with photorealistic rendering. It supports both PX4 and ArduPilot, making it versatile for various UAV research applications. Flightmare \citep{song2020flightmare} combines Unity’s rendering capabilities with RotorS’ physics modeling, facilitating learning applications such as autonomous drone racing and high-speed flight. FlightGoggles \citep{guerra2019flightgoggles}, designed for vision-based navigation research, provides a photorealistic sensor simulation environment that aids in developing and testing UAV perception algorithms. Additionally, numerous other platforms utilize OpenGL, Unity, Unreal Engine, NVIDIA Isaac Sim, and OmniverseRTX for rendering in UAV simulations. These diverse simulation environments enable researchers to develop, test, and refine UAV algorithms efficiently under various conditions.

However, for complex multibody systems involving multiple UAVs or intricate simulation environments, existing physics engines predominantly simulate flexible bodies using Position-Based \citep{muller2007position} or Particle-Based \citep{bell2005particle} techniques instead of the widely used methods in engineering like the absolute nodal coordinates formulation (ANCF) and spring-damper models. Previous studies on the tethered net dynamics for space debris capture have primarily utilized two modeling approaches. The spring-damper model has been widely adopted due to its computational efficiency and straightforward implementation \citep{botta2016simulation, botta2017contact}. This method discretizes the net into a set of mass nodes connected by spring-damper elements, effectively capturing large deformations while maintaining high computational efficiency. On the other hand, the absolute nodal coordinate formulation (ANCF) has been explored for more precise modeling of net flexibility and large deformation behavior \citep{gerstmayr2006analysis}. ANCF represents net elements using absolute position coordinates and their gradients, allowing for a more continuous and physically accurate description of net deformation and contact dynamics. While this approach improves simulation accuracy, it significantly increases the degrees of freedom and computational cost. A comparative study evaluated both approaches in space debris capture scenarios \citep{shan2020analysis}. The results indicate that, while ANCF provides superior accuracy in modeling net flexibility, its high computational cost makes it impractical for large-scale, real-time simulations. Conversely, the spring-damper model demonstrates adequate accuracy with significantly reduced computational overhead, making it a more viable option for multi-UAV-tethered netted system simulations. Therefore, due to its computational efficiency and suitability for large-scale multibody simulations, the spring-damper model is chosen in this study for simulating multi-UAV-tethered netted system.

On the other hand, commercial multibody dynamics-based (MBD) software, such as Adams, RecurDyn, Chrono \citep{mazhar2013chrono,tasora2016chrono}, and MWorks, often lack comprehensive visual information modules, limiting their ability to integrate advanced vision-based odometry algorithms. For instance, OpenVINS \citep{geneva2020openvins} is a filter-based visual-inertial estimator that fuses inertial measurements with sparse visual feature tracks, providing robust state estimation for UAV applications. ORB-SLAM3 \citep{campos2021orb} extends monocular and stereo SLAM capabilities to multi-map and visual-inertial settings, offering improved localization and mapping accuracy. DM-VIO \citep{stumberg22dmvio} enhances visual-inertial odometry by jointly optimizing IMU and visual data, improving robustness in dynamic environments. In addition to the lack of support for vision-based algorithms, these MBD tools also lack interfaces with modern reinforcement learning (RL) frameworks, which are essential for developing and evaluating UAV control strategies. Stable Baselines3 \citep{raffin2021stable}, for example, provides a widely adopted RL framework in PyTorch, offering implementations of key algorithms such as PPO, DDPG, and SAC for UAV training and optimization. The absence of such integration capabilities in traditional MBD software highlights the need for more comprehensive simulation environments that bridge vision-based state estimation and learning-based control strategies.

To address these limitations and enable the multi-UAV-tethered netted system capture of non-cooperative targets, we propose the development of a multibody dynamics-based simulation environment that integrates high-precision physics modeling, vision-based motion tracking, and reinforcement learning capabilities.

To implement the multi-UAV-tethered netted system for the capture task, as illustrated in Fig.~\ref{fig:multi-drone_vis}, we developed a custom simulator, mySim, with the following key features:

\begin{itemize}
\item	Accuracy: Built on multibody dynamics, mySim achieves precision comparable to commercial simulation software in validated modules;
\item	Flexibility: The system's structural configuration can be easily modified via programmatic descriptions due to its MBD-based architecture;
\item	Motion Tracking Support: By integrating visual perception modules and vision-based detection algorithms, mySim enables motion tracking of both UAVs and non-cooperative targets.
\end{itemize}

This paper is organized as follows, Sect.~\ref{sec:main_intro} discusses the background and related work on UAV-based tasks. Sect.~\ref{sec:main_meth} presents the implementation of the mySim simulation environment, including dynamics modeling and control design. Sect.~\ref{sec:main_res} validates each module's correctness and demonstrates the feasibility of the multi-UAV-tethered netted system capture strategy. Sect.~\ref{sec:main_dis} provides a discussion and future outlook.

\section{Methodology}\label{sec:main_meth}

\begin{figure}[h]
    \centering
                \includegraphics[width=12cm]{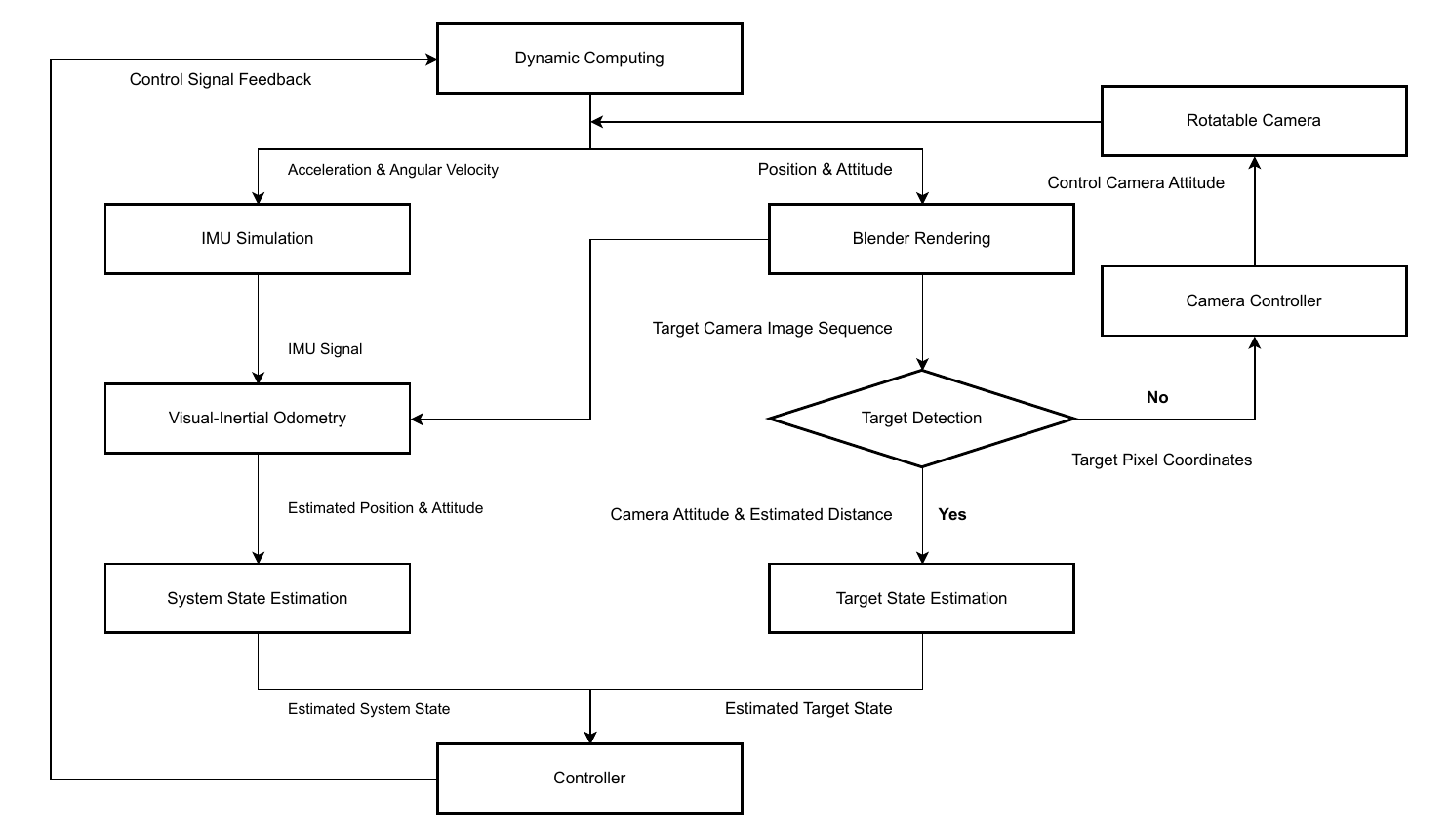}
    \caption{Flowchart of the overall solution framework}
    \label{fig:entire}
\end{figure}

To address the challenge of capturing non-cooperative target, based on the multi-UAV-tethered netted system as shown in Fig.~\ref{fig:multi-drone_vis}, this section presents the complete solution, as illustrated in Fig.~\ref{fig:entire}. The proposed framework is built upon the multibody dynamics-based simulation environment mySim, which integrates a post-processing module to incorporate visual information. By leveraging both the dynamics computation results and visual data, the system achieves fundamental perception and control functionalities.

Sect.~\ref{sec:rope_collision} introduces the dynamics modules for rope modeling and collision modeling. Sect.~\ref{sec:uav_design} details the design of the UAV and its onboard camera system, along with the corresponding controller design. Sect.~\ref{sec:perception} describes the perception module.

\subsection{Design of Rope and Collision Modules in a Multibody Dynamics-based Simulation Environment}\label{sec:rope_collision}

\begin{figure}[h]
    \centering
                \includegraphics[width=10cm]{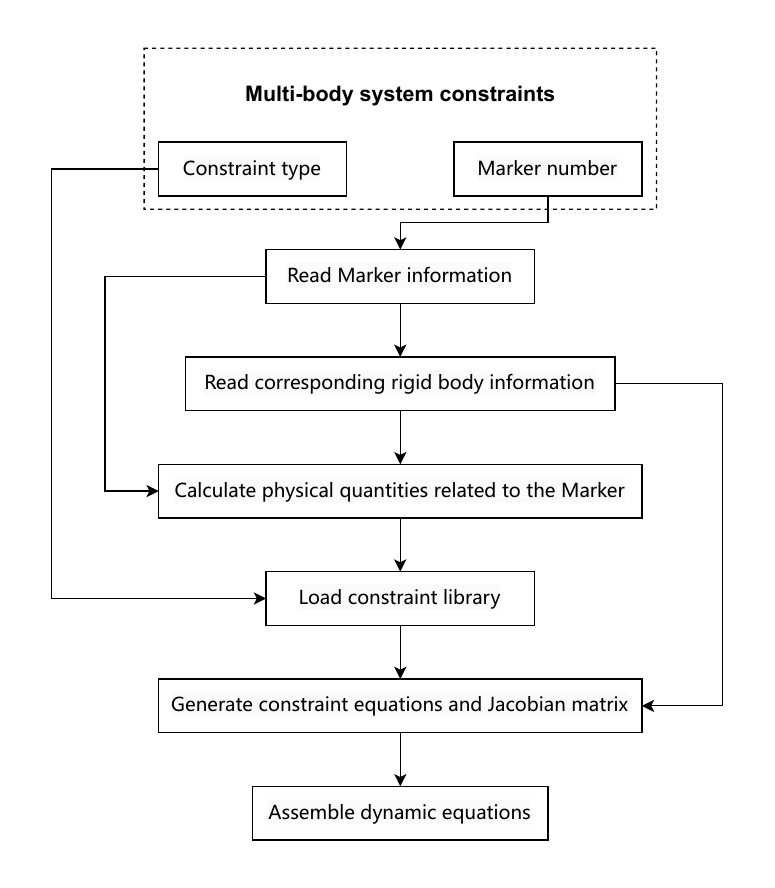}
    \caption{Flowchart of dynamics equation construction based on marker technology}
    \label{fig:multibody_marker}
\end{figure}

\begin{figure}[h]
    \centering
                \includegraphics[width=10cm]{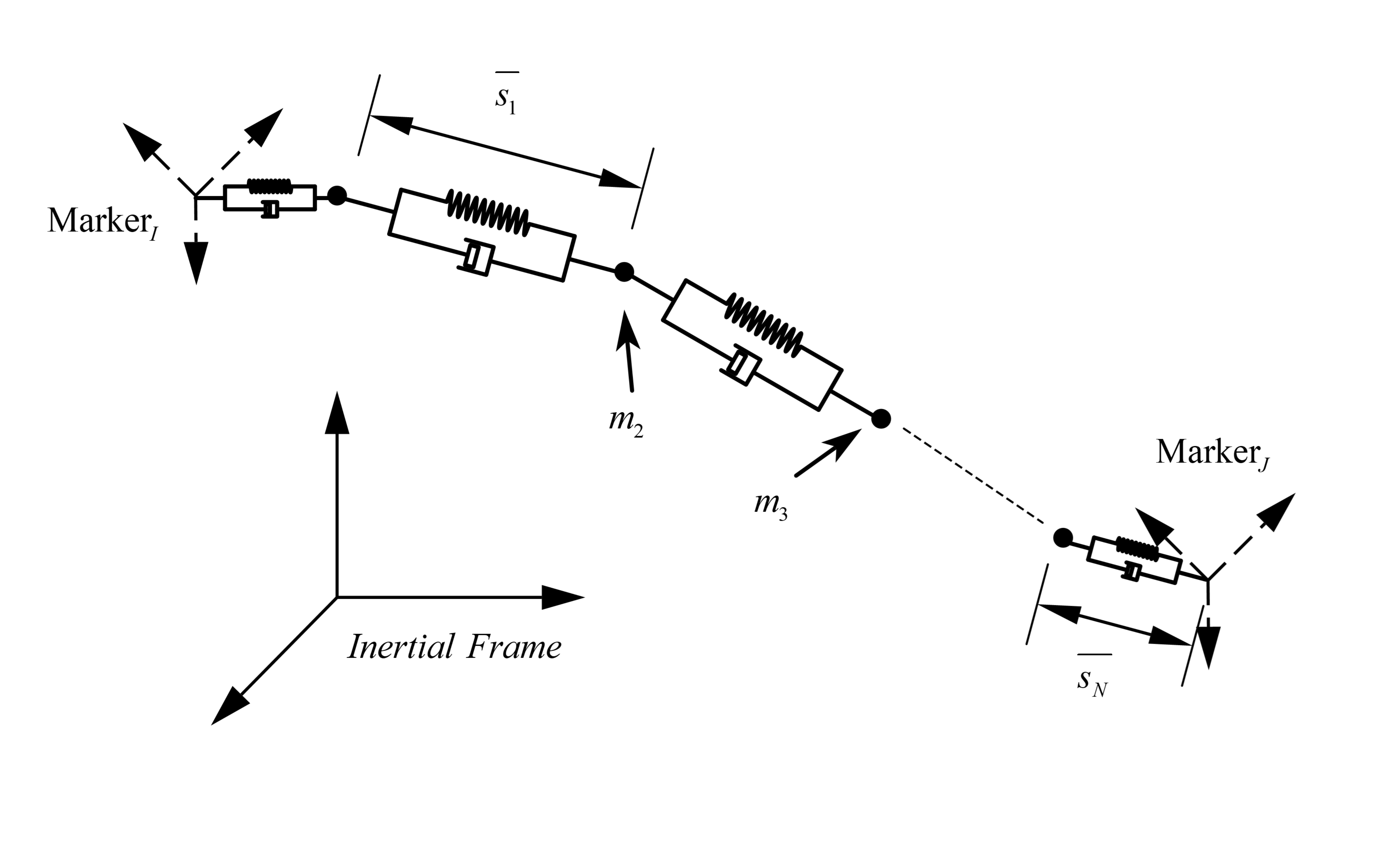}
    \caption{Schematic diagram of parameterized rope module based on spring-damper model, the ends of the rope can interact with other modules in the multibody system through markers.}
    \label{fig:rope_model}
\end{figure}

\begin{figure}[h]
    \centering
                \includegraphics[width=10cm]{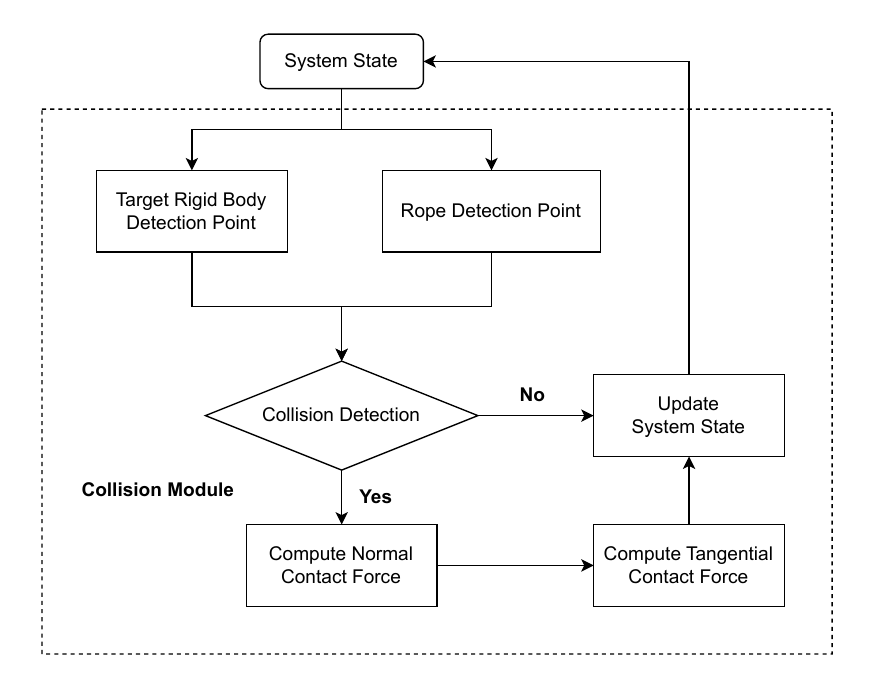}
    \caption{Flowchart of collision detection module}
    \label{fig:collision}
\end{figure}

\begin{figure}[h]
    \centering
% \begin{adjustwidth}{-\extralength}{0cm}
\subfloat[\centering]{\includegraphics[width=7cm]{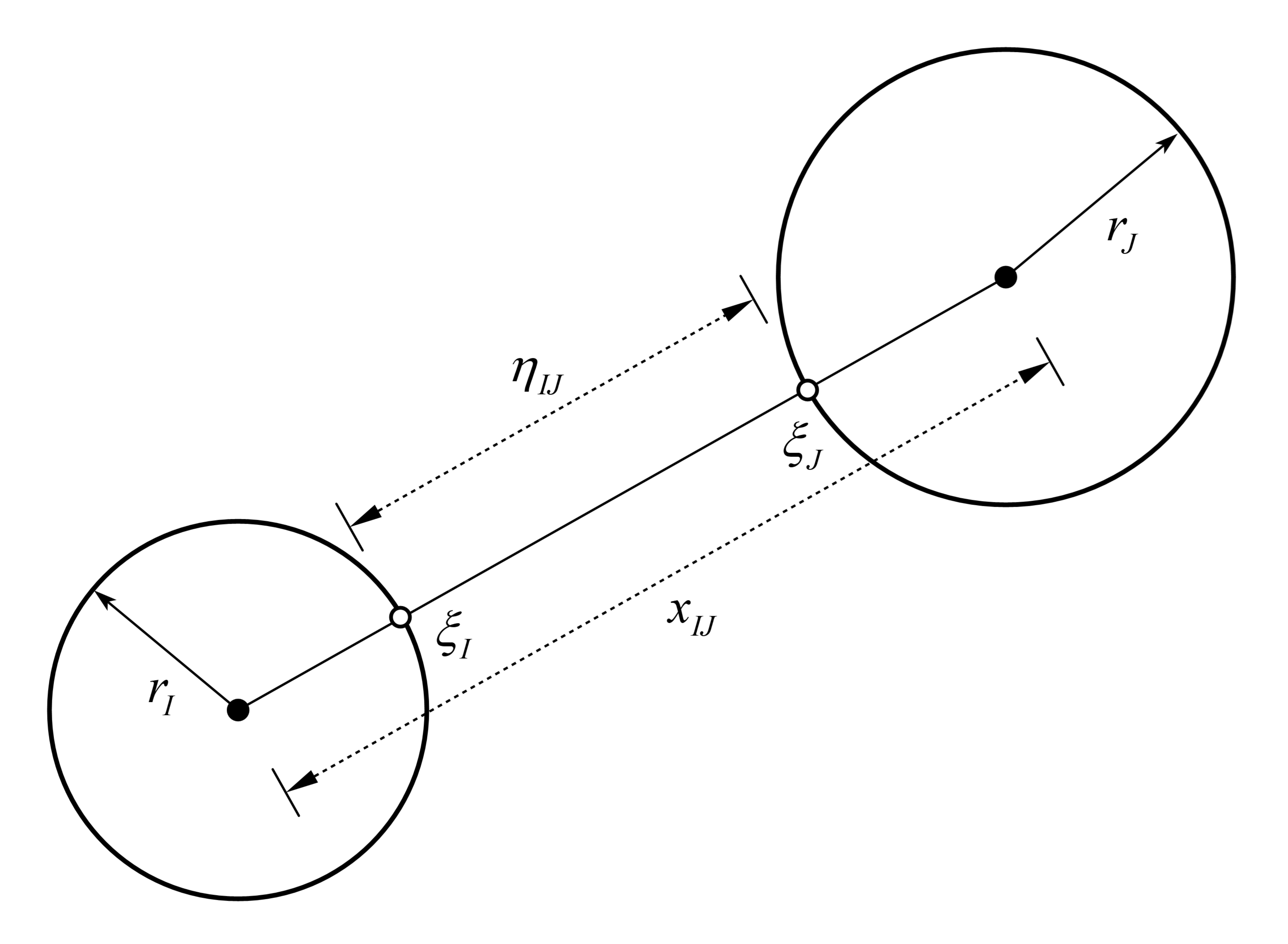}}
\hfill
\subfloat[\centering]{\includegraphics[width=4.5cm]{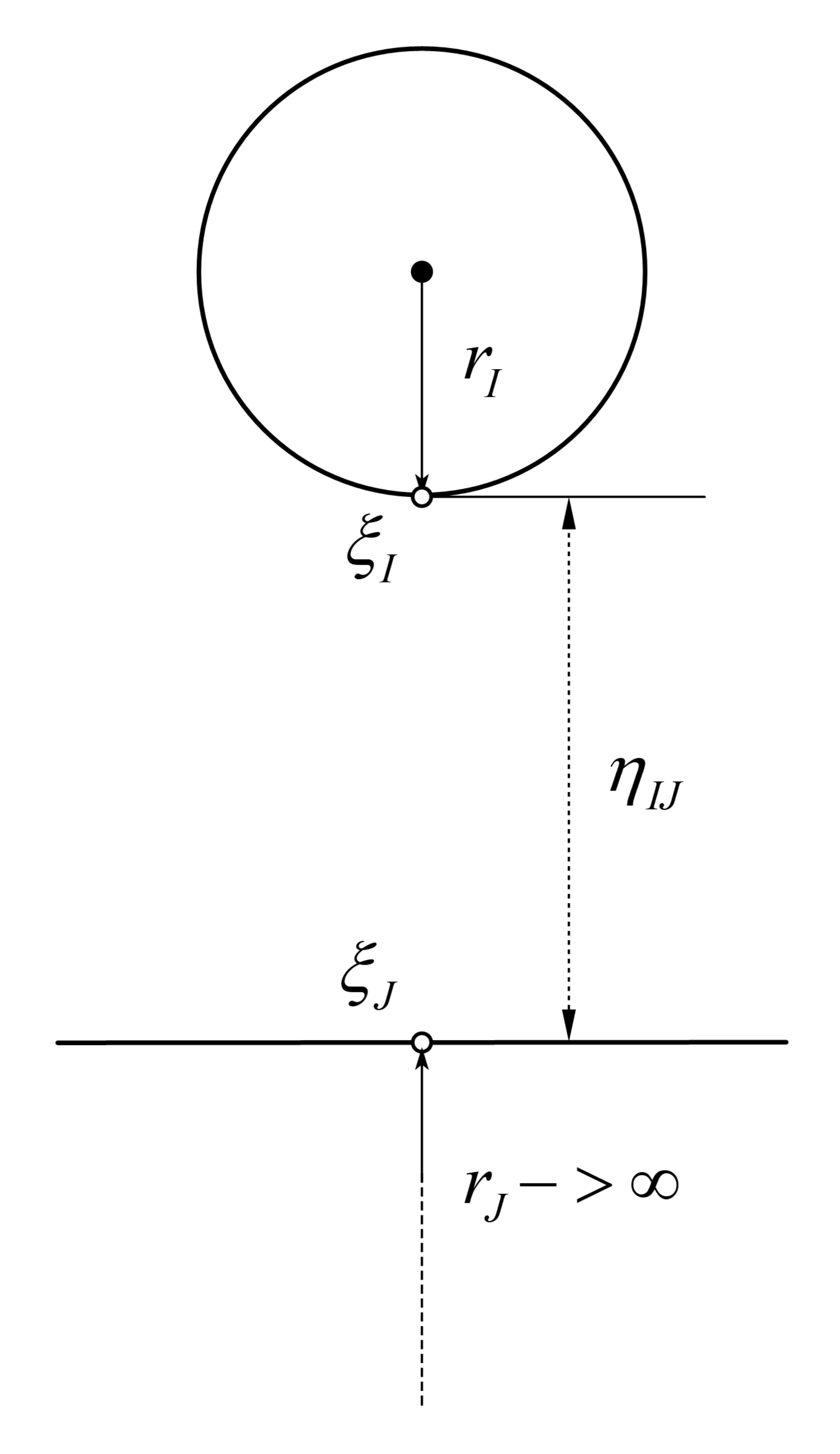}}
\hfill
\subfloat[\centering]{\includegraphics[width=4.5cm]{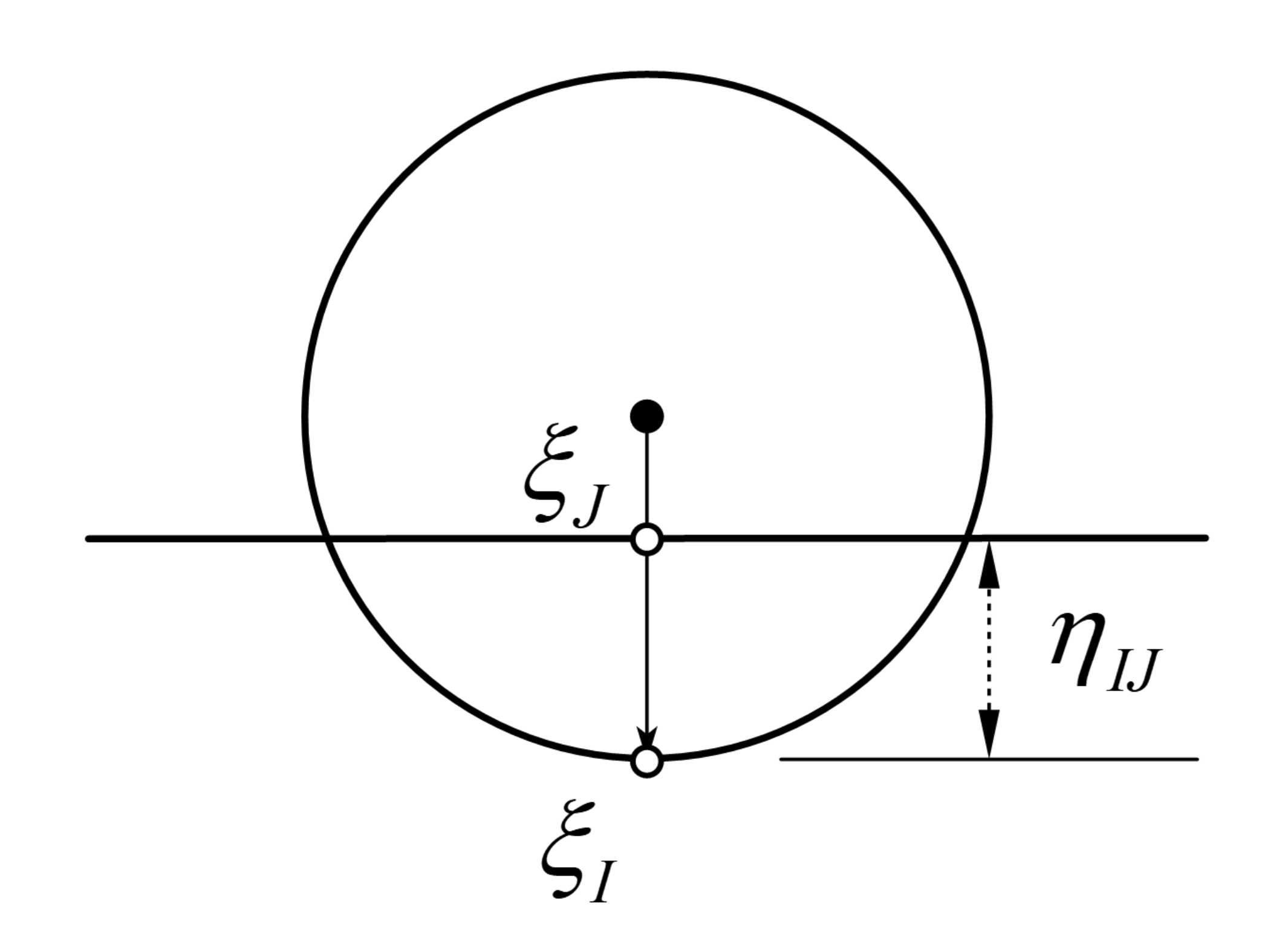}}

% \end{adjustwidth}
    \caption{Schematic diagram of collision detection module for the two bodies, where (\textbf{b}) shows the condition before collision, and (\textbf{c}) indicates the situation colliding.}
    \label{fig:collision_demonstration}
\end{figure}

\begin{figure}[h]
    \centering
                \includegraphics[width=10cm]{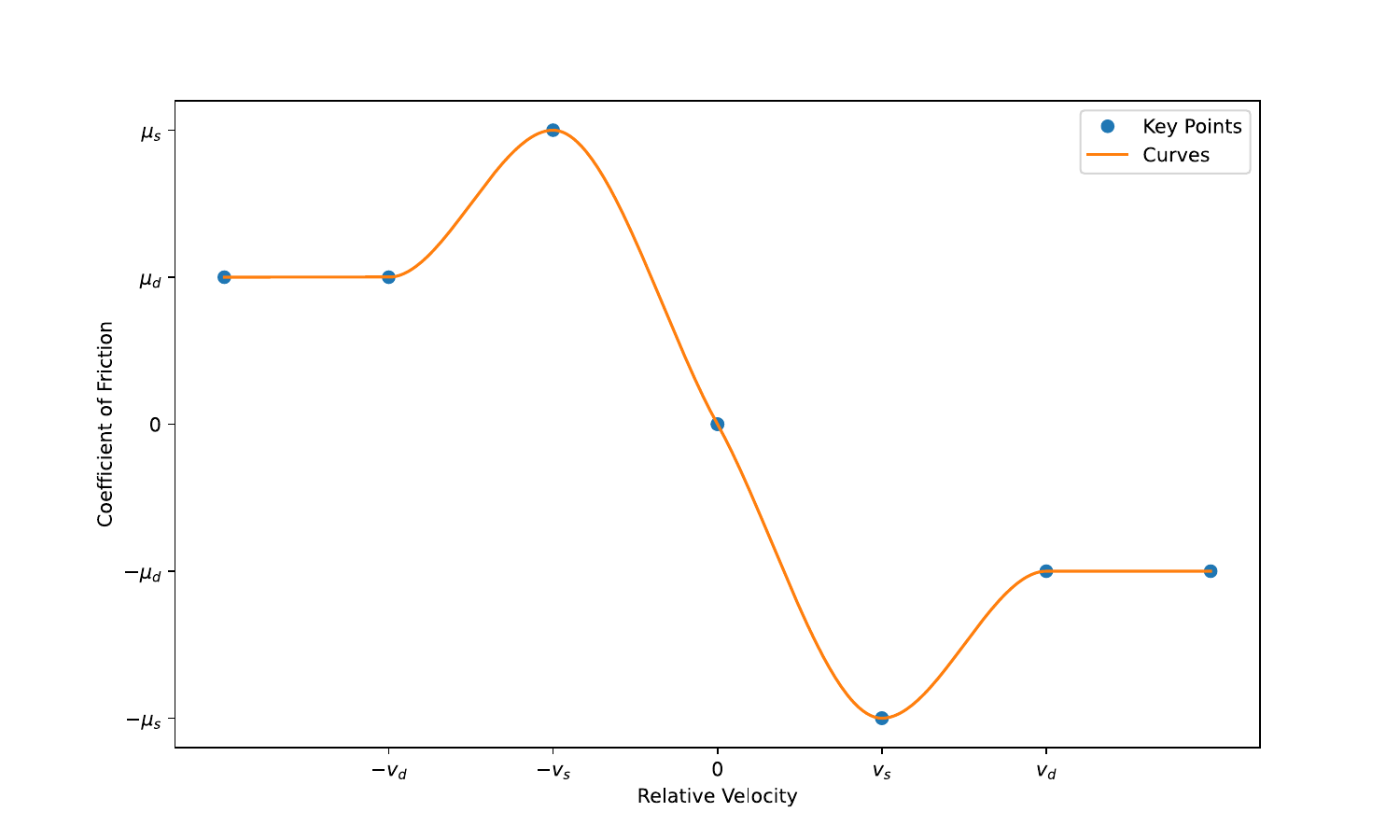}
    \caption{Schematic diagram of friction coefficient $\mu$ and relative velocity coefficient $v_t$.}
    \label{fig:collsion miu-v}
\end{figure}

In a smooth dynamical system without contact or collision, for a multibody system using marker technology, the flow diagram can be described in Fig.~\ref{fig:multibody_marker}, the dynamic equations can be expressed as a system of differential algebraic equations as Eq.~\eqref{eq:MBDwithoutCollision}

\begin{linenomath*} 
\begin{equation}\label{eq:MBDwithoutCollision}
    \begin{aligned}
          \mathbf{M}\dot{\mathbf{v}}+\mathbf{C}_{\mathbf{q}}^{T}(\mathbf{q},t)\lambda-\mathbf{F}(\mathbf{q},\mathbf{v},t)&=\mathbf{0} \\
        \mathbf{C}(\mathbf{q},t)&=\mathbf{0}
    \end{aligned}
\end{equation}
\end{linenomath*}
where $\mathbf{M}$ is the system's mass matrix, $\mathbf{q}$ represents the generalized coordinates,  $\mathbf{C}$ represents the constraint equation, and $\mathbf{F}$ denotes the generalized forces, with $\lambda$ representing the Lagrange multipliers.

A spring-damping model is used to simulate the rope module in the rope net, and the spring-damping model of the rope module can be established as shown in Fig.~\ref{fig:rope_model}.

Let the total length of the rope in its initial state be $L$, therefore, the natural length of the rope between two mass points is then:

\begin{linenomath}
    \begin{equation}
        \overline{s}_i=\frac{L}{N}
    \end{equation}
\end{linenomath}
where $N$ represents the number of concentrated masses. To facilitate the connection between the rope module and external modules, the lengths at both ends of the rope are set to:

\begin{linenomath}
    \begin{equation}
        \overline{s}_0=\overline{s}_N=\frac{\overline{s}_i}{2}
    \end{equation}
\end{linenomath}

The mass of the rope $m$ is given by:

\begin{linenomath}
    \begin{equation}
        m=\sum_{i}m_i
    \end{equation}
\end{linenomath}
where $m_i$ is the mass of each point mass.

The axial stiffness of the rope $EA$ and the linear tensile damping coefficient $d$ are defined as:
\begin{linenomath}
    \begin{equation}
        k_i=\frac{EA}{s_i}
    \end{equation}
\end{linenomath}

\begin{linenomath}
    \begin{equation}
        d_i=\frac{d}{s_i}
    \end{equation}
\end{linenomath}
where $k_i$ is the stiffness and $d_i$ is the damping coefficient.

The elastic force $f_i^k$ and the damping force $f_i^d$ of the concentrated mass rope are calculated using the following formulas:

\begin{linenomath}
    \begin{equation}
        f_i^k=k_i(s_i-\overline{s}_i)
    \end{equation}
\end{linenomath}

\begin{linenomath}
    \begin{equation}
        f_i^d=d_i{\frac{\mathrm d}{\mathrm{d}t}}(s_i-\overline{s}_i)
    \end{equation}
\end{linenomath}
From these, the internal forces of the rope can be derived as
\begin{linenomath}
    \begin{equation}
        F_i=f_i^k+f_i^d
    \end{equation}
\end{linenomath}

For dynamical systems considering collisions, when a contact event occurs, the contact forces affect the multibody system. In this case,the flow diagram can be described in Fig.~\ref{fig:collision}, Eq.~\eqref{eq:MBDwithoutCollision} is modified as Eq.~\eqref{eq:MBDwithCollision}

\begin{linenomath*} 
\begin{equation}\label{eq:MBDwithCollision}
    \begin{aligned}
          \mathbf{M}\dot{\mathbf{v}}+\mathbf{C}_{\mathbf{q}}^{T}(\mathbf{q},t)\lambda-\mathbf{F}(\mathbf{q},\mathbf{v},t)-\mathbf f&=\mathbf{0} \\
        \mathbf{C}(\mathbf{q},t)&=\mathbf{0}
    \end{aligned}
\end{equation}
\end{linenomath*}
where $\mathbf{f}$ represents the generalized contact forces that act on the contact points within the system.

For the penalty function method, the contact force between two colliding bodies is calculated using an impact function as Eq.~\eqref{eq:compact_force}. The contact process is simulated by considering the elasticity and damping of the contact surfaces of the colliding bodies, employing a non-linear equivalent spring-damping model.

For each colliding body, a convex sphere with radius $r_N$ is centered on the centroid. The collision detection module for the two bodies is shown in Fig.~\ref{fig:collision_demonstration}.

For collision bodies $I$ and $J$, their radius are denoted as $r_I$ and $r_J$. The potential contact points of the two bodies are $\xi_I$ and $\xi_J$, and the distance between their centers is $x_{IJ}$. The contact state variable is a Boolean, defined as:
\begin{linenomath}
    \begin{equation}
        \Delta=(x_{IJ}\leq r_I+r_J)
    \end{equation}
\end{linenomath}
where the contact state is true, the contact force and frictional force are calculated, otherwise, when the contact state is false, both the contact force and frictional force are zero. Similarly, the distance between the two colliding bodies can be determined:
\begin{linenomath}
    \begin{equation}
        \eta_{IJ}=\left\Vert  (\xi_I,\xi_J) \right\Vert
    \end{equation}
\end{linenomath}

The normal contact force of collision on body $I$ is given by:
\begin{linenomath}
    \begin{equation}\label{eq:compact_force}
        f_I^c=\max{\left[0,k(\eta_{IJ})^n-\mathrm{step}(\eta_{IJ}+r_I,\eta_{IJ}-p,d,0,0)\frac{\mathrm d\eta_{IJ}}{dt} \right]}
    \end{equation}
\end{linenomath}
where $p$ represents the maximum penetration depth, $d$ is the damping coefficient, and $n$ is the stiffness index. The function $\mathrm{step}(x,x_0,h_0,x_1,h_1)$ is expressed as:
\begin{linenomath*}
    \begin{equation}
        \mathrm{step}=\begin{cases}
            h_0 &x\leq x_0\\h &x_0<x<x_1\\h_1 & x\geq x_1
        \end{cases}
    \end{equation}
\end{linenomath*}
where $h$ is obtained through Hermite interpolation.

Based on the Eq.~\eqref{eq:compact_force}, the contact frictional force can be expressed as:
\begin{linenomath}
    \begin{equation}
        f_I^f=-\mu(v_t)f_I^c\mathrm{sign}(v_t)
    \end{equation}
\end{linenomath}
where the friction coefficient $\mu(v_t)$  is given by Eq.~\eqref{eq:collsion miu-v}, the relationship between the friction coefficient $\mu$ and relative velocity $v_t$ can be depicted in Fig.~\ref{fig:collsion miu-v}.
\begin{linenomath*}
    \begin{equation}\label{eq:collsion miu-v}
        \mu(\nu_{t})=
\begin{cases}
\mu_d\,\mathrm{sign}(\nu_t) &\left|\nu_t\right|>\nu_d \\
-\left\{\mu_d+(\mu_s-\mu_d)\left(\frac{\left|\nu_t\right|-\nu_s}{\nu_d-\nu_s}\right)^2 \left[3-2\left(\frac{\left|\nu_t\right|-\nu_s}{\nu_d-\nu_s}\right)\right]\right\}\,\mathrm{sign}(\nu_t) &\nu_s\leq\left|\nu_t\right|\leq\nu_d \\
2\mu_d\left[3\left(\frac{\nu_t+\nu_d}{2\nu_d}\right)^2-2\left(\frac{\nu_t+\nu_d}{2\nu_d}\right)^3-\frac12\right] &\left|\nu_t\right|<\nu_s
\end{cases}
    \end{equation}
\end{linenomath*}
where  $v_s$ and $v_d$ represent the critical coefficients for static and dynamic friction, respectively, while $\mu_s$ and $\mu_d$ denote the static and dynamic friction coefficients.

\subsection{Modeling of UAV Dynamics and Controller Design}\label{sec:uav_design}

\begin{figure}[h]
    \centering
        \subfloat[\centering]{\includegraphics[width=8cm]{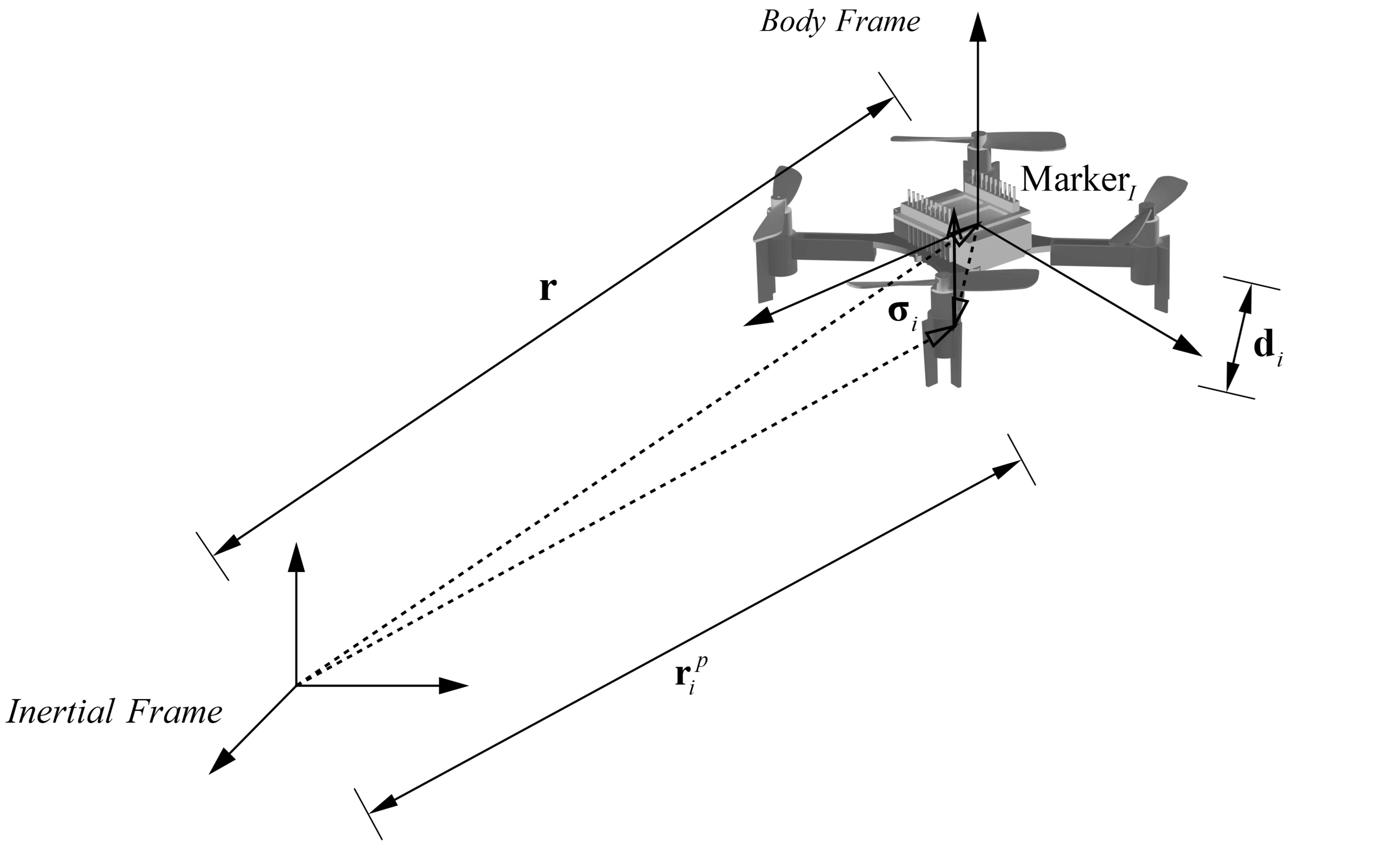}}
        \hfill
        \subfloat[\centering]{\includegraphics[width=5cm]{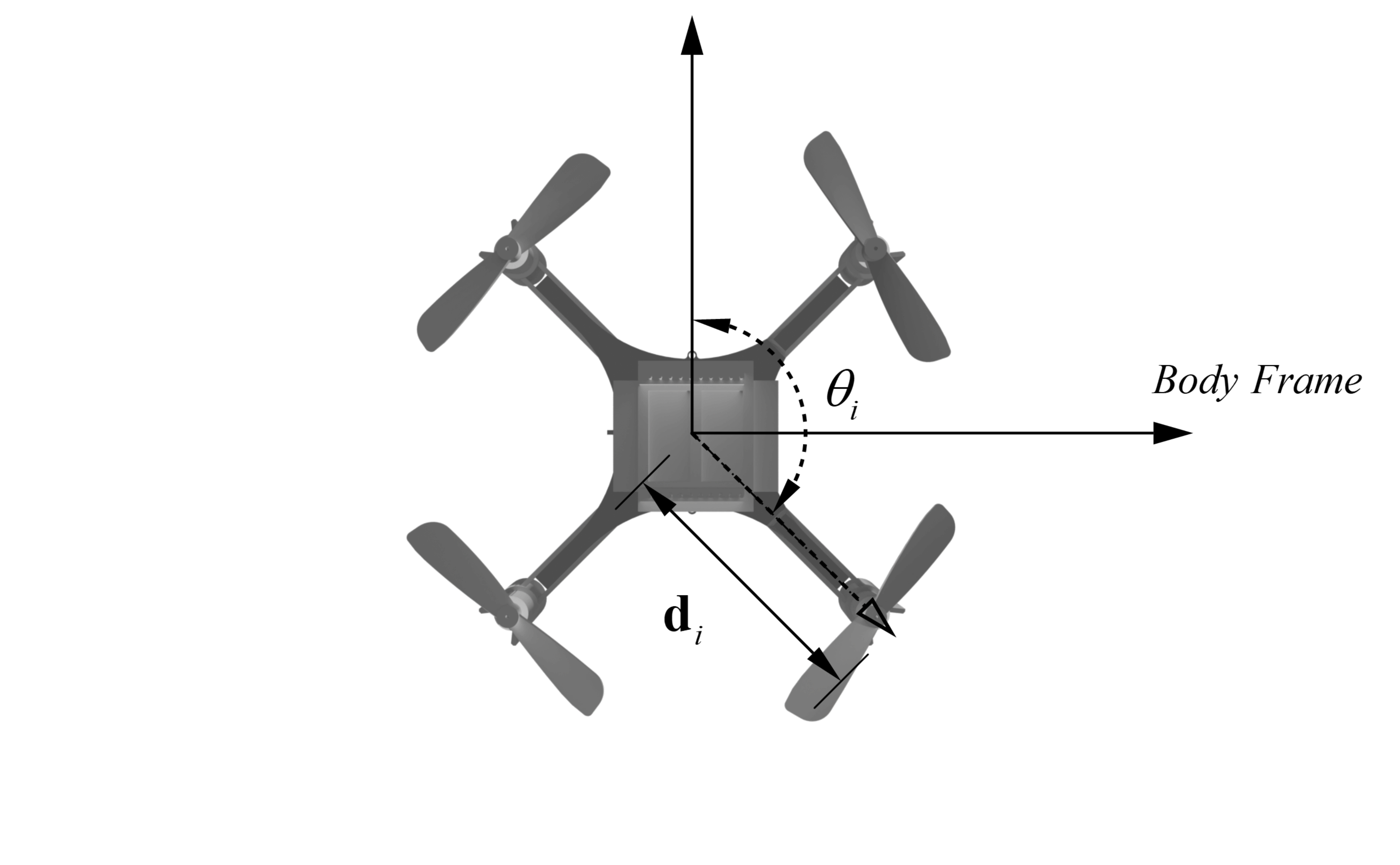}}\\
                
    \caption{Schematic diagram of UAV dynamics model, (\textbf{a}) is presented from a side view perspective, and (\textbf{b}) is presented from a top-down view perspective, the UAV's center can interact with other modules in the multibody system through markers.}
    \label{fig:uav_model}
\end{figure}

\begin{figure}[h]
    \centering
                \includegraphics[width=10cm]{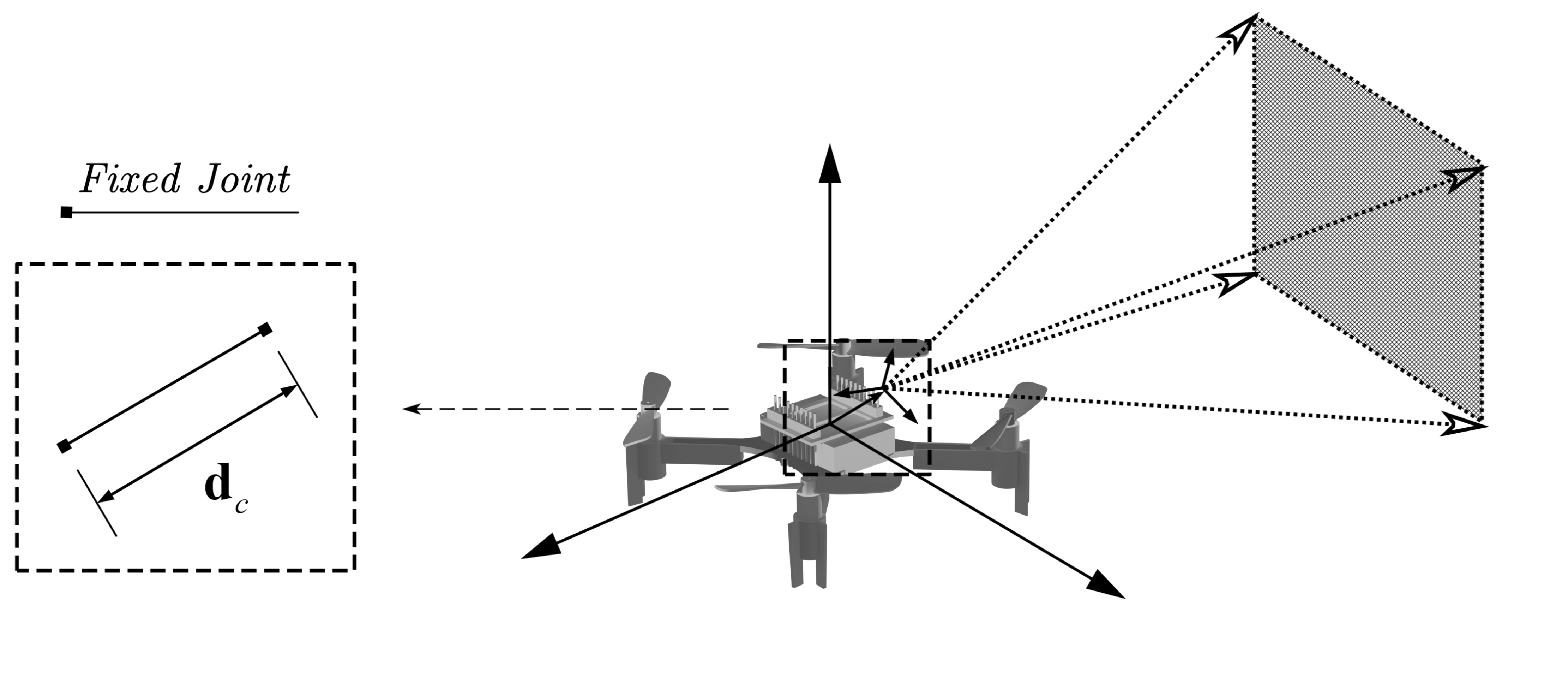}
    \caption{Schematic diagram of simulating the monocular camera for visual information acquisition, where the camera is fixed relative to the UAV's center $\mathbf d_c$ through a fixed joint, as shown in Eq.~\eqref{eq:appendix_fixed}, by Marker technology.}
    \label{fig:camera_fixed}
\end{figure}

\begin{figure}[h]
    \centering
                \includegraphics[width=10cm]{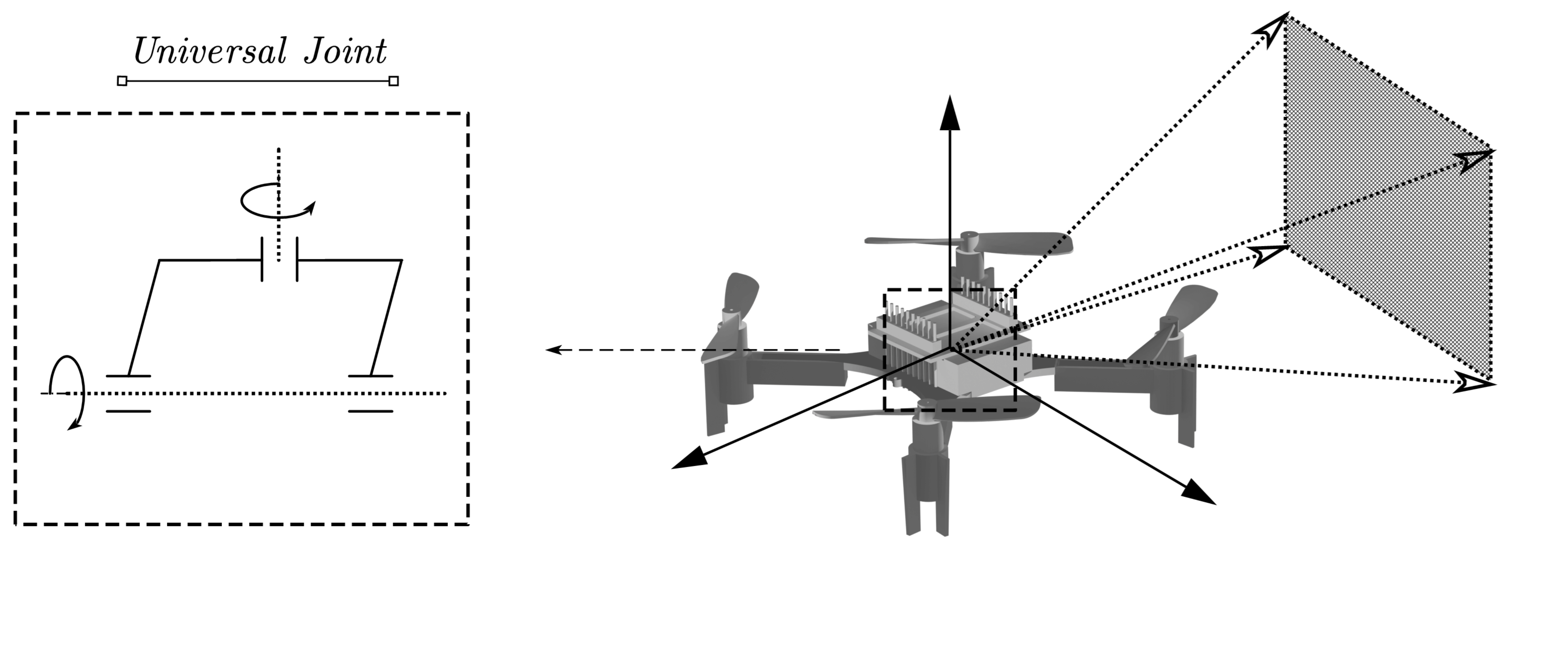}
    \caption{Schematic diagram of simulating a rotatable camera on a two-axis gimbal for visual information acquisition, where the camera at the UAV's center is connected to the camera through a universal joint, as shown in Eq.~\eqref{eq:appendix_universal}, by Marker technology.}
    \label{fig:camera_universal}
\end{figure}

\begin{figure}[h]
    \centering
                \includegraphics[width=13.5cm]{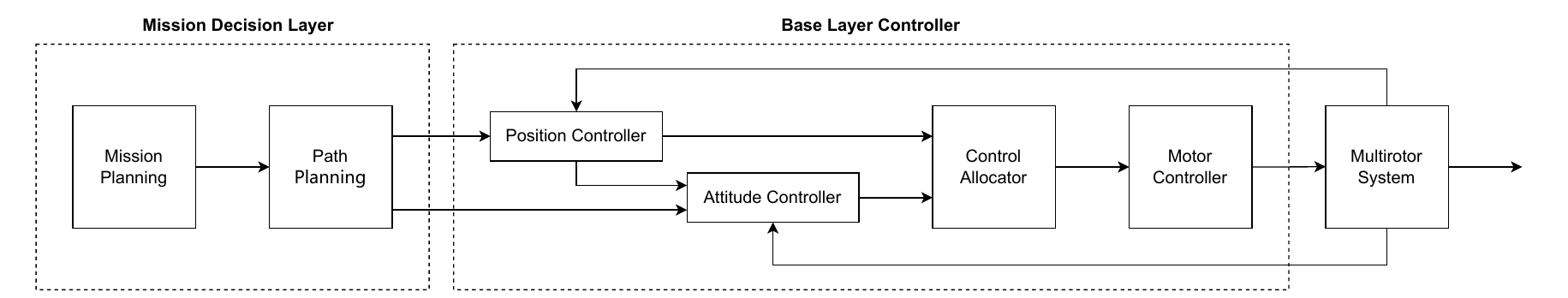}
    \caption{Schematic diagram of entire controller, where the base layer controller adopts an inner-outer loop control structure. The outer loop controller is a position controller, and the inner loop controller is an attitude controller. The controllers generate the desired motor speeds.}
    \label{fig:entire_controller_pd}
\end{figure}

\begin{figure}[h]
    \centering
                \includegraphics[width=10cm]{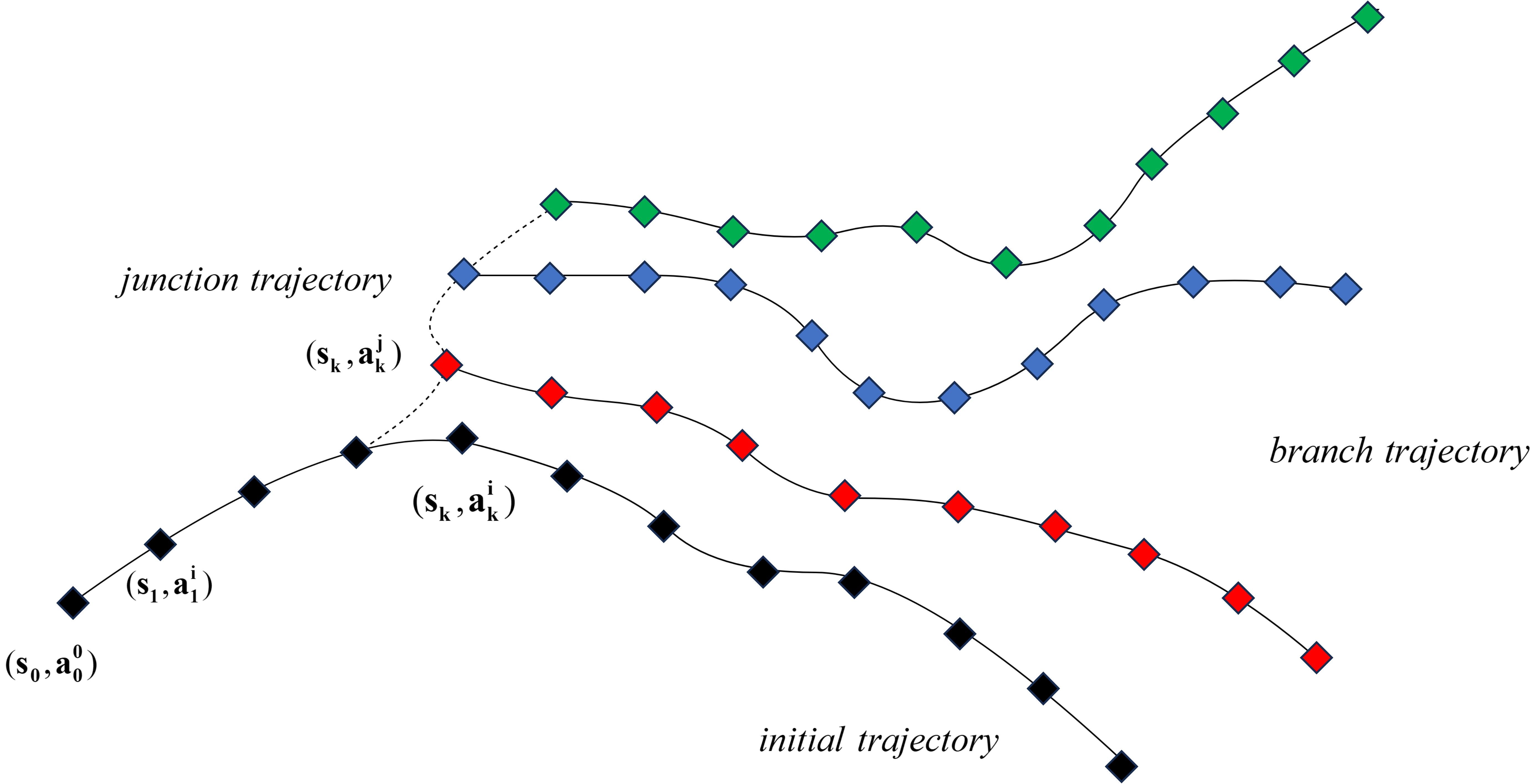}
    \caption{Schematic diagram of trajectory, during a complete simulation process, based on the observation variables $\mathbf s_n$ output at each time step from the simulation environment, the controller outputs a new action $\mathbf a_n^m$, continuing until the task is completed. }
    \label{fig:ppo_demonstration}
\end{figure}

\begin{figure}[h]
    \centering
    \begin{adjustwidth}{-\extralength}{0cm}
        \subfloat[\centering]{\includegraphics[width=9cm]{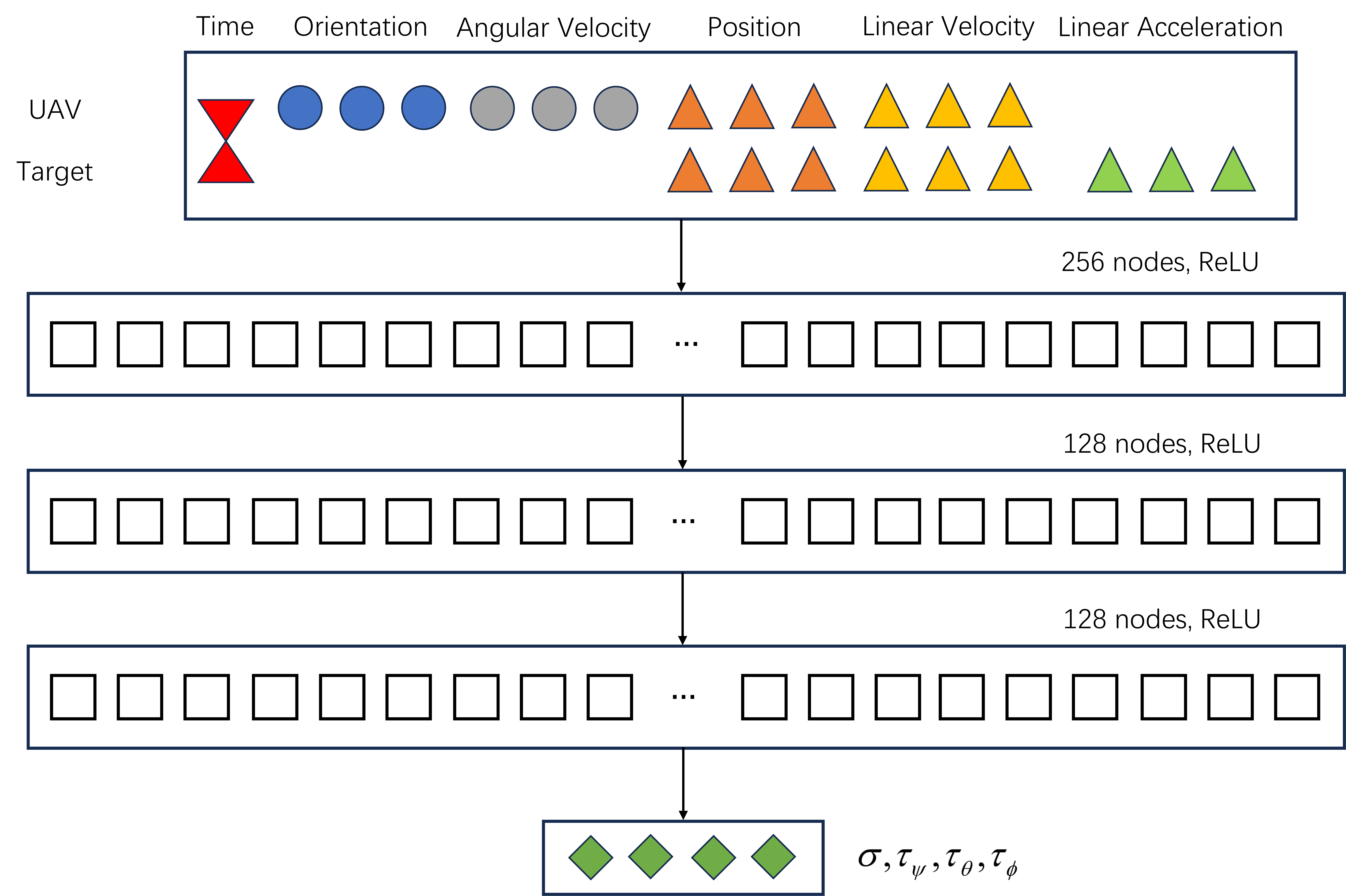}}
        \hfill
        \subfloat[\centering]{\includegraphics[width=9cm]{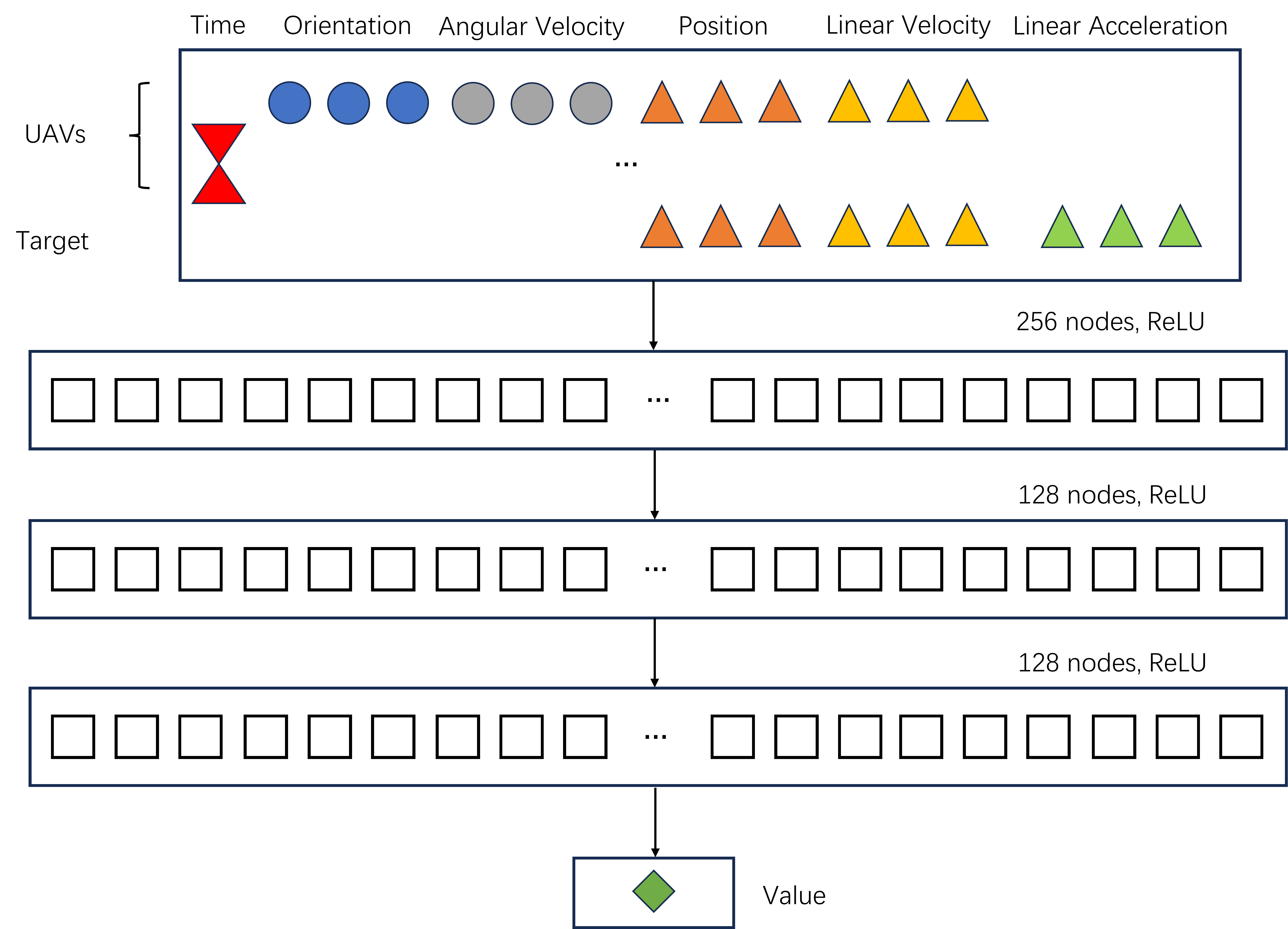}}\\
    \end{adjustwidth}            
    \caption{Schematic diagram of network for each UAV, The policy network (\textbf{a}) outputs the action $\mathbf a_i$  based on the observation $\mathbf s_i$ and is continuously updated using reinforcement learning algorithms. The value network (\textbf{b}) estimates the state value for each state  to assist in the training and optimization of the policy network.}
    \label{fig:mappo_LNN}
\end{figure}

The dynamic model of the multirotor UAV can be established as shown in Fig.~\ref{fig:uav_model}, and the vehicle's dynamic model can be presented with the vehicle's coordinate system position given as:
\begin{linenomath}
    \begin{equation}
        \mathbf r=\left[x,y,z\right]^{T}
    \end{equation}
\end{linenomath}

Let $l_i$ denote the distance from the UAV's center to the center of each rotor, and let $\theta_i$ represent the offset angle of each rotor relative to the front of the vehicle. The positions of the individual rotors on the UAV can be described as:
\begin{linenomath}
    \begin{equation}
        \mathbf r_i^p=\mathbf r+\mathbf A \mathbf d_i
    \end{equation}
\end{linenomath}
where $\mathbf A$ denotes the transformation from the vehicle's coordinate system to the inertial frame, and $\mathbf d_i$ can be expressed as:
\begin{linenomath}
    \begin{equation}
        \mathbf d_i=l_i\left[\sin\theta_i,\cos\theta_i,0\right]
    \end{equation}
\end{linenomath}

Obviously, the velocity and acceleration in the vehicle's coordinate system is:
\begin{linenomath*}
    \begin{equation}
    \begin{aligned}
        \mathbf v=&\dot{\mathbf r}=\left[\dot x,\dot y,\dot z\right]^{T}\\
        \mathbf a=&\ddot{\mathbf r}=\frac{\mathbf A \boldsymbol{\sigma}+\mathbf F^{\mathrm{ext}}}{m}
    \end{aligned}
    \end{equation}
\end{linenomath*}
where  $\mathbf F^{\mathrm{ext}}$ represents external forces such as drag and gravity, and $\boldsymbol \sigma$ refers to the overall thrust produced by the UAV's rotors, defined by:
\begin{linenomath}
    \begin{equation}\label{eq:thrust_by_rotors}
        \boldsymbol \sigma=\sum_i\mathbf  A_i \boldsymbol \sigma_i=\sum_ic_i\omega_i^2 \mathbf A_i\frac{\boldsymbol \sigma_i}{\Vert\boldsymbol \sigma_i\Vert}
    \end{equation}
\end{linenomath}
where $\boldsymbol \sigma_i$  is the normal thrust generated by each rotor, $\omega_i$ is the rotational speed of each rotor, and $c_i$ is the lift coefficient for each rotor. The torque $\boldsymbol \tau$ generated by the thrust produced by the rotors on the vehicle's body can be expressed as:
\begin{linenomath}
    \begin{equation}
        \boldsymbol \tau=\sum_i \tilde{\mathbf d}_i \boldsymbol \sigma_i+k_i\boldsymbol \sigma_i
    \end{equation}
\end{linenomath}
where $k_i$ represents the coefficient for the counter-torque generated by each rotor.

When using quaternions to describe the rotation of the vehicle's coordinate system, the quaternion product can be utilized to express that:
\begin{linenomath}
    \begin{equation}
        \boldsymbol {\dot\Lambda}=\frac12 \boldsymbol \Lambda\otimes\boldsymbol \omega
    \end{equation}
\end{linenomath}
hence, there can be:
\begin{linenomath}
    \begin{equation}
        \boldsymbol{\dot\omega=J^{-1}\left(\tau-\tilde{\omega}J\omega \right)}
    \end{equation}
\end{linenomath}

For situations that require the assistance of visual information, two cameras can be connected using Marker technology, as shown in Fig.~\ref{fig:camera_fixed} and Fig.~\ref{fig:camera_universal}, to simulate the monocular camera fixed on the UAV and the camera connected via a two-axis gimbal.

The overall control architecture of the UAV is shown in Fig.~\ref{fig:entire_controller_pd}, which can be divided into the mission decision layer and the base layer control layer.

The multirotor is a kind of underactuated UAV system with four input variables corresponding to the motor speeds. Therefore, the multirotor can only track four desired commands, which are typically represented by three directional components and the yaw angle. The remaining variables are determined by the desired commands.

Given the desired flight trajectory for the multirotor, and utilizing its differential flatness property, the control can be decomposed into two parts: position and yaw, denoted as $\mathbf r_d(t)$ and $\psi_d(t)$. To achieve the desired trajectory for the UAV, the following conditions must be satisfied as time progresses:
\begin{linenomath}
    \begin{equation}
        \left\Vert \mathbf x(t) -\mathbf x_d(t)\right\Vert\to0
    \end{equation}
\end{linenomath}
Alternatively, the following condition must also hold:
\begin{linenomath}
    \begin{equation}
        \left\Vert \mathbf x(t) -\mathbf x_d(t)\right\Vert\to\mathrm{U}(0,\delta)
    \end{equation}
\end{linenomath}
Where $\mathbf x=\left[\mathbf r^{\prime} ,\psi \right]^{\prime}$, $\mathbf x_d=\left[\mathbf r_d^{\prime} ,\psi_d \right]^{\prime}$, and $\mathrm{U}(0,\delta)$ represents a neighborhood centered at the origin with radius.

With the total thrust provided by the rotors to the UAV body given by Eq.~\eqref{eq:thrust_by_rotors},  the torque generated by the rotors can be described as:
\begin{linenomath*}
    \begin{equation}
        \begin{aligned}
  \tau_{\theta}&=\frac{\sqrt{2}}{2}l_{i}c_{i}(\omega_{1}^{2}-\omega_{2}^{2}-\omega_{3}^{2}+\omega_{4}^{2}) \\
  \tau_{\phi}&=\frac{\sqrt{2}}{2}l_{i}c_{i}(\omega_{1}^{2}+\omega_{2}^{2}-\omega_{3}^{2}-\omega_{4}^{2}) \\
  \tau_{\psi}&=k_{i}(\omega_{1}^{2}-\omega_{2}^{2}+\omega_{3}^{2}-\omega_{4}^{2})
        \end{aligned}
    \end{equation}
\end{linenomath*}

Thus, the following can be obtained:
\begin{linenomath*}
    \begin{equation}\mathbf M=
\begin{pmatrix}
\Vert\boldsymbol \sigma \Vert \\
\tau_\theta \\
\tau_\phi \\
\tau_\psi
\end{pmatrix}
\begin{pmatrix}
\omega_1^2 \\
\omega_2^2 \\
\omega_3^2 \\
\omega_4^2
\end{pmatrix}^{-1}\end{equation}
\end{linenomath*}
where $\mathbf M$ represents the control allocation matrix, which allows the separation of upper-level control from lower-level control.

For tasks where the path cannot be explicitly specified, reinforcement learning-based algorithms can be employed. The control problem is decomposed into decision-making problems at each time step, resulting in the trajectory shown in Fig.~\ref{fig:ppo_demonstration}.

The MAPPO \citep{yu2022surprising} algorithm is one of the most widely used multi-agent reinforcement learning algorithms. The algorithm's framework is illustrated in Fig.~\ref{fig:ippo}. The network architecture for each UAV is shown in Fig.~\ref{fig:mappo_LNN}

\subsection{Pose Estimation and Non-cooperative Target Detection Based on Visual Information}\label{sec:perception}

During the simulation process, at each time step, the positions and orientations of the UAVs, their cameras, and the non-cooperative target are computed and fed into a post-processing module for rendering, as outlined in Alg.~\ref{alg:blender_rendering}. The visual information captured by the cameras, returned from the post-processing environment, is illustrated in Fig.~\ref{fig:camera_fixed} and Fig.~\ref{fig:camera_universal}.

\begin{figure}[h]
    \centering
        \subfloat[\centering]{\includegraphics[width=5.5cm]{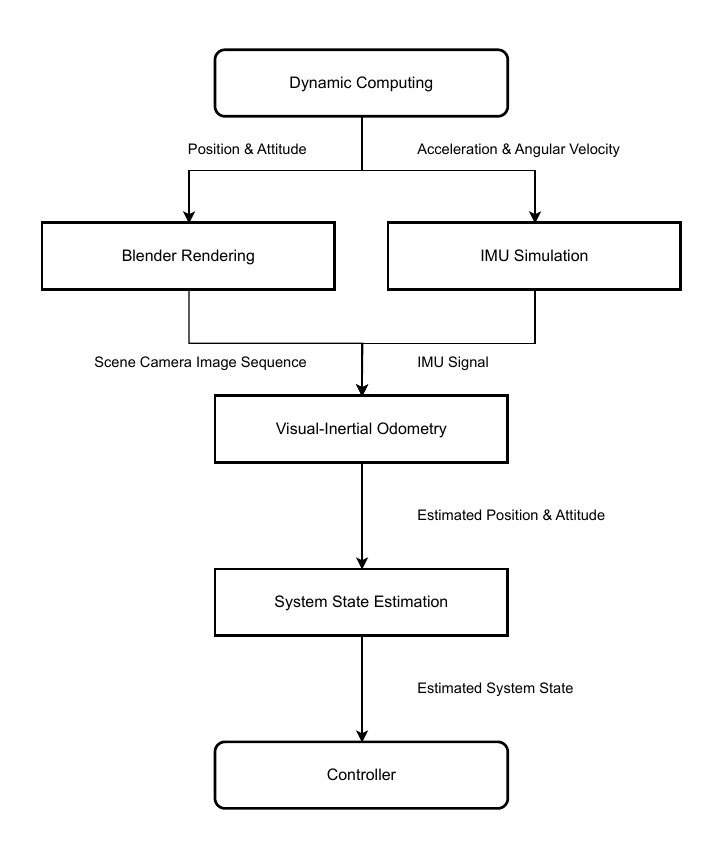}}
        \hfill
        \subfloat[\centering]{\includegraphics[width=7.7cm]{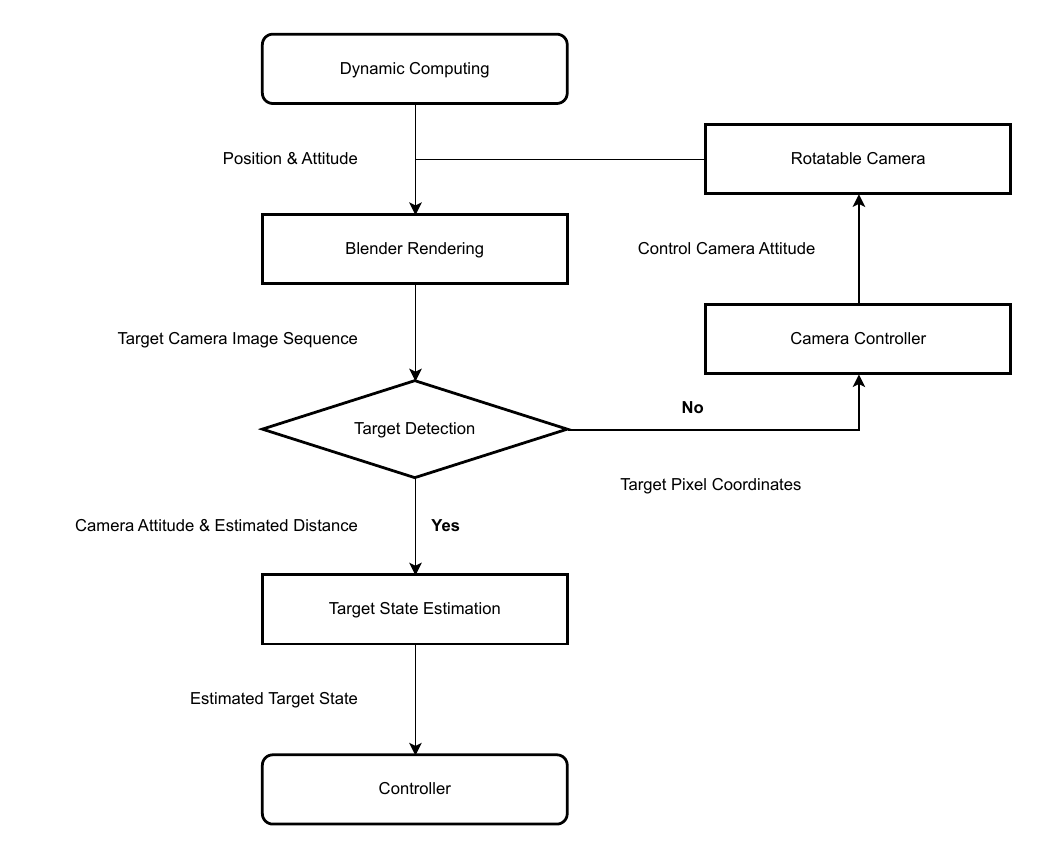}}\\
                
    \caption{Flowchart of state estimation,
(\textbf{a}) shows the estimation of UAV position and orientation by integrating visual information and simulated IMU signals,
(\textbf{b}) shows the estimation of the non-cooperative target’s position using visual detection models.}
    \label{fig:estimation & track}
\end{figure}

\begin{figure}[h]
    \centering
                \rotatebox{0}{\includegraphics[width=10cm]{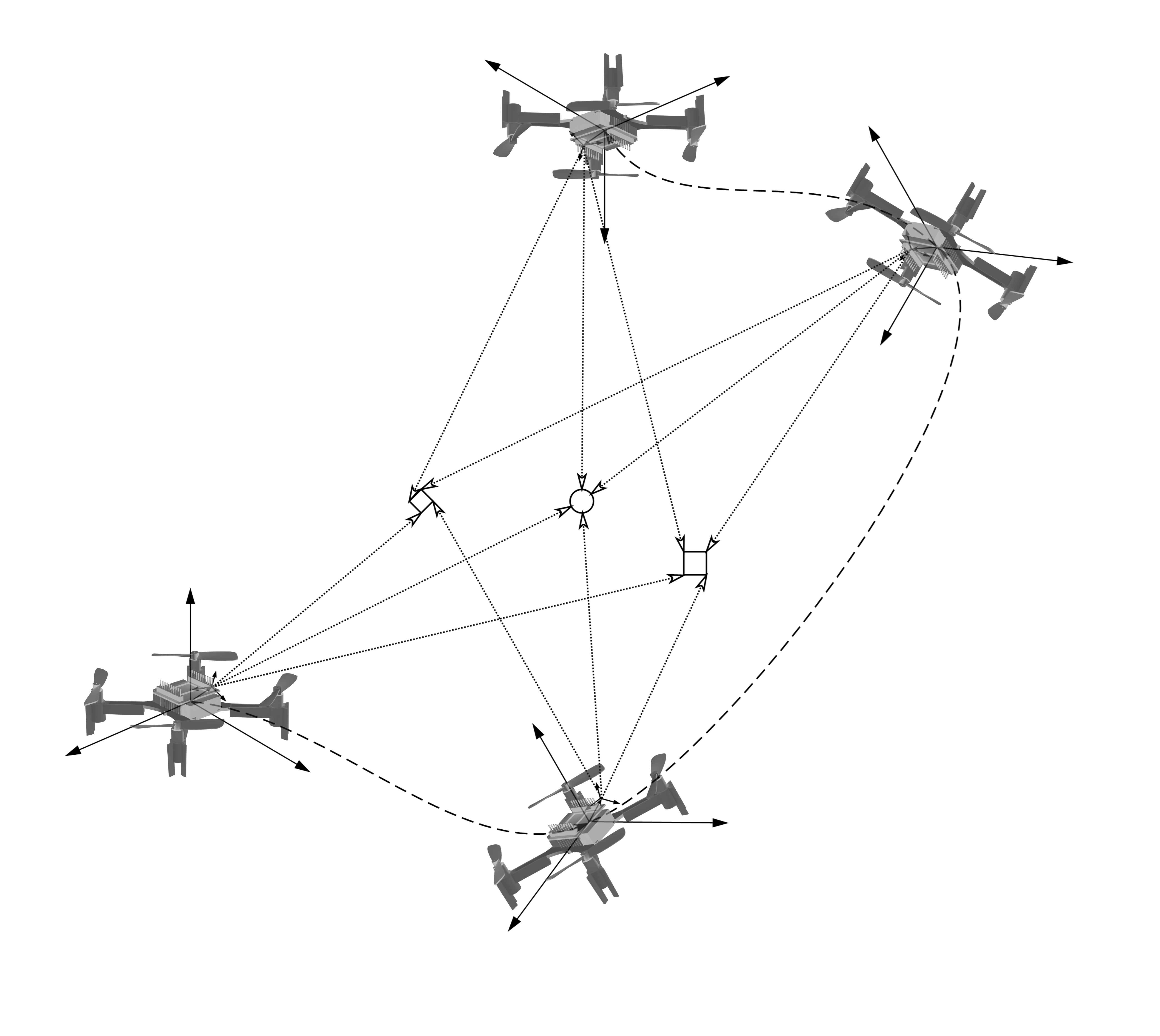}}
    \caption{Schematic diagram of UAV's estimated position and orientation by extracting features from visual information captured by the camera in Fig.~\ref{fig:camera_fixed} and combining them with simulated IMU signals from the simulation environment.}
    \label{fig:estimation_demonstration}
\end{figure}

\begin{figure}[h]
    \centering
                \includegraphics[width=8cm]{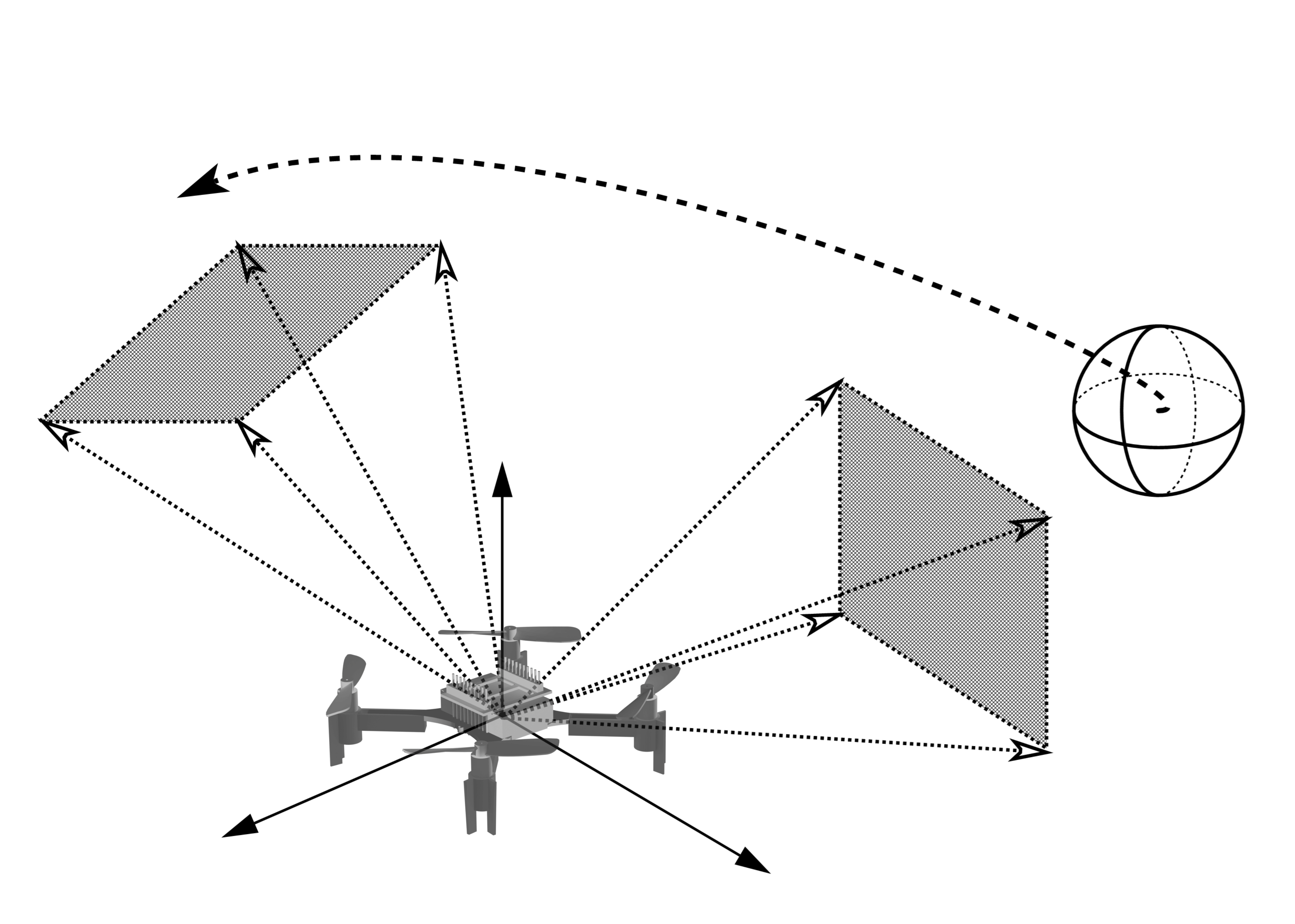}
    \caption{Schematic diagram of tracking process, the UAV captures visual information through the camera shown in Fig.~\ref{fig:camera_universal}. Based on the detected position of the non-cooperative target, the UAV updates the camera’s orientation to align with the target.}
    \label{fig:track_demonstration}
\end{figure}

With employing the vision-based estimation method shown in Fig.~\ref{fig:estimation & track}, the estimated states of the UAVs and the non-cooperative target at each time step are computed. The estimation of UAV position and orientation is implemented based on the VIO approach, shown in Fig.~\ref{fig:estimation_demonstration}, utilizing the widely used VINS-MONO \citep{qin2018vins} algorithm in engineering applications, as described in Alg.~\ref{alg:drone_state_estimation_process}. The detection and tracking of the non-cooperative target are achieved using DETR \citep{carion2020end} with a ResNet-101 \citep{he2016identity} backbone. The detected target from the visual feedback is used to control the two-axis gimbal shown in Fig.~\ref{fig:camera_universal}, aligning its orientation toward the target and achieving tracking, as demonstrated in Fig.~\ref{fig:track_demonstration}. This process is implemented in Alg.~\ref{alg:target_tracking_process}.

Based on the estimated UAV positions $\hat{\mathbf r}_i$, combined with the detected dimensions $[w_i,h_i]$ of the non-cooperative target in the images, define:
\begin{linenomath}
    \begin{equation}
        \alpha_i=\frac{w_i \cdot h_i}{W \cdot H}
    \end{equation}
\end{linenomath}
the radial distance from the non-cooperative target to the camera can then be approximated as:
\begin{linenomath}
    \begin{equation}
        \beta_i \propto \frac{1}{\alpha_i}
    \end{equation}
\end{linenomath}

Using three cameras from each set $i,j,k$, the estimated non-cooperative target state can be described as $\hat{\mathbf h}_t^{ijk}$, and by fusing $\binom{n}{3}$ sets of results caught by $n$ UAVs, the final estimated target state $\hat{\mathbf h}_t$ can be obtained.

The obtained state estimates serve as the required inputs for the algorithm depicted in Fig.~\ref{fig:mappo_LNN} at each time step, enabling the implementation of the control and decision-making tasks as outlined in Alg.~\ref{alg:controller_process_mappo}.

\section{Results and validation}\label{sec:main_res}

To demonstrate the feasibility of the proposed approach, we conducted a series of simulations that progressively evaluate and integrate the system's core modules. In Sect.~\ref{sec:rope_collision_uav}, the rope, collision, and UAV modules were individually validated using simplified scenarios, including a suspended rope net with payload, collision interactions, and UAV trajectory tracking through the controller. Building upon this, Sect.~\ref{sec:perception_control} focuses on validating the perception and control pipeline, where vision-inertial state estimation and collaborative target tracking were integrated to support accurate motion estimation. Finally, in Sect.~\ref{sec:multi_uav_capture}, all modules were integrated into a complete multi-UAV-tethered netted system, which was tested in representative capture scenarios involving both free-falling and maneuvering non-cooperative targets. These staged evaluations collectively verify the correctness and applicability of the proposed method under increasingly complex conditions.

\subsection{Validation of the Rope, Collision, and UAV Modules}\label{sec:rope_collision_uav}

\begin{figure}
    \centering
                \includegraphics[width=8cm]{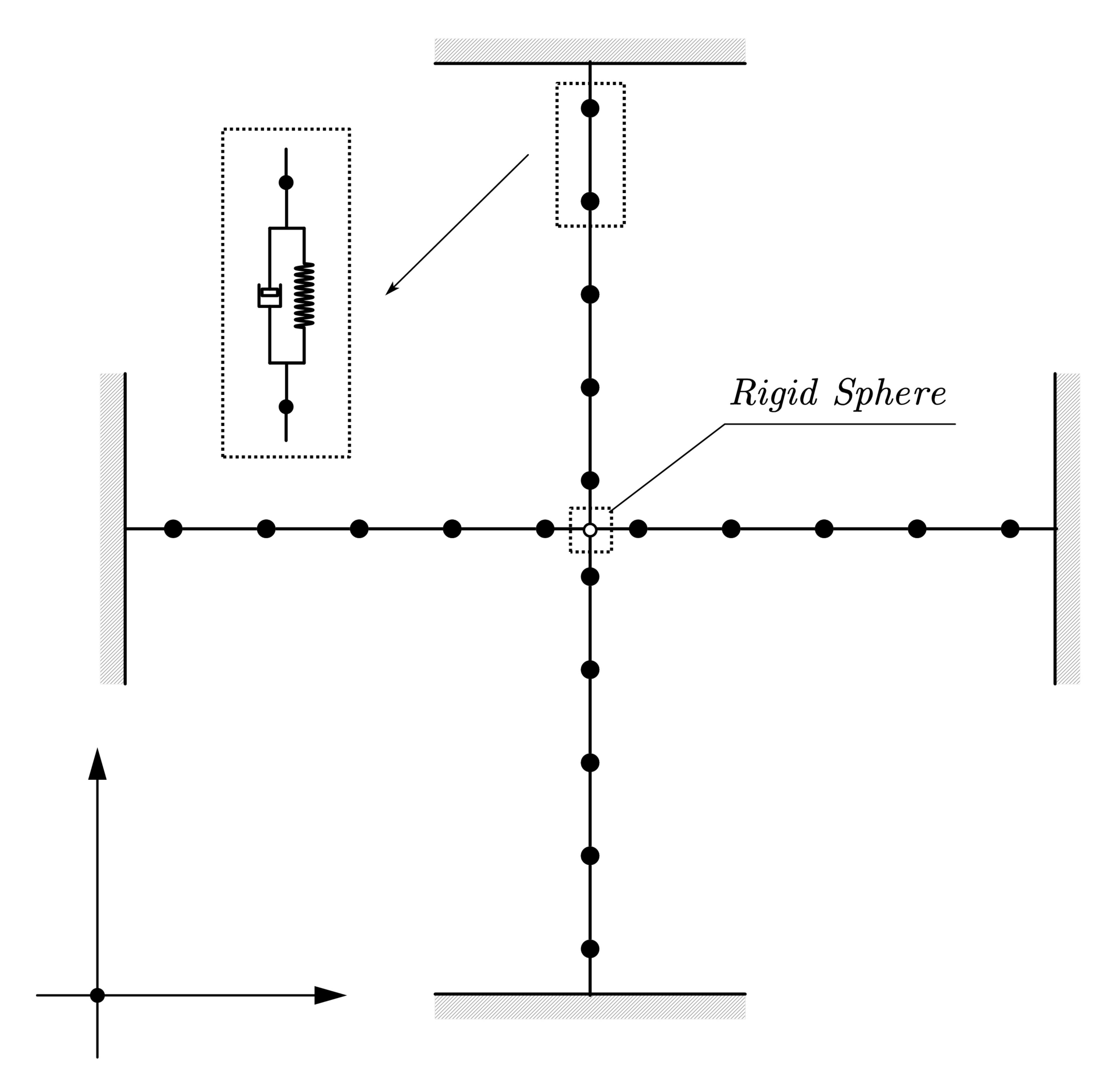}
    \caption{Schematic diagram of the simple rope net model, composed of rope modules from Fig.~\ref{fig:rope_model}, with fixed ends and a centrally attached payload.}
    \label{fig:rope_net_model_demonstration}
\end{figure}

To verify the feasibility of the rope module, a simple rope net model was constructed, as shown in Fig.~\ref{fig:rope_net_model_demonstration}, which illustrates a schematic diagram of the rope net composed of rope modules from Fig.~\ref{fig:rope_model}, where the net is fixed at all ends with a payload attached at the center, and its parameters were set according to Table.~\ref{table:rope_net_model}. The goal of this setup is to observe the net's dynamic response to the payload's motion.

\begin{table}
\caption{Parameters for the rope net model.\label{table:rope_net_model}}
\newcolumntype{Y}{>{\raggedright\arraybackslash}X}
\begin{tabularx}{\textwidth}{Y c c c}
\toprule
\textbf{Parameter} & \textbf{Symbol} & \textbf{Value} & \textbf{Unit} \\
\midrule
Axial stretching stiffness & \( EA \) & \( 1.00 \times 10^5 \) & \( N \) \\
Linear damping coefficient & \( d \) & \( 5.00 \times 10^1 \) & \( N \cdot s \cdot m^{-1} \) \\
Number of concentrated masses & \( N \) & \( 5.00 \times 10^0 \) & - \\
Initial rope length & \( L \) & \( 1.00 \times 10^0 \) & \( m \) \\
Point mass & \( m_i \) & \( 2.00 \times 10^{-1} \) & \( kg \) \\
Payload mass & \( m_{\text{payload}} \) & \( 2.00 \times 10^{-1} \) & \( kg \) \\
\bottomrule
\end{tabularx}
\end{table}

\begin{figure}
\centering
\begin{adjustwidth}{-\extralength}{0cm}
\subfloat[\centering]{\includegraphics[width=5.5cm]{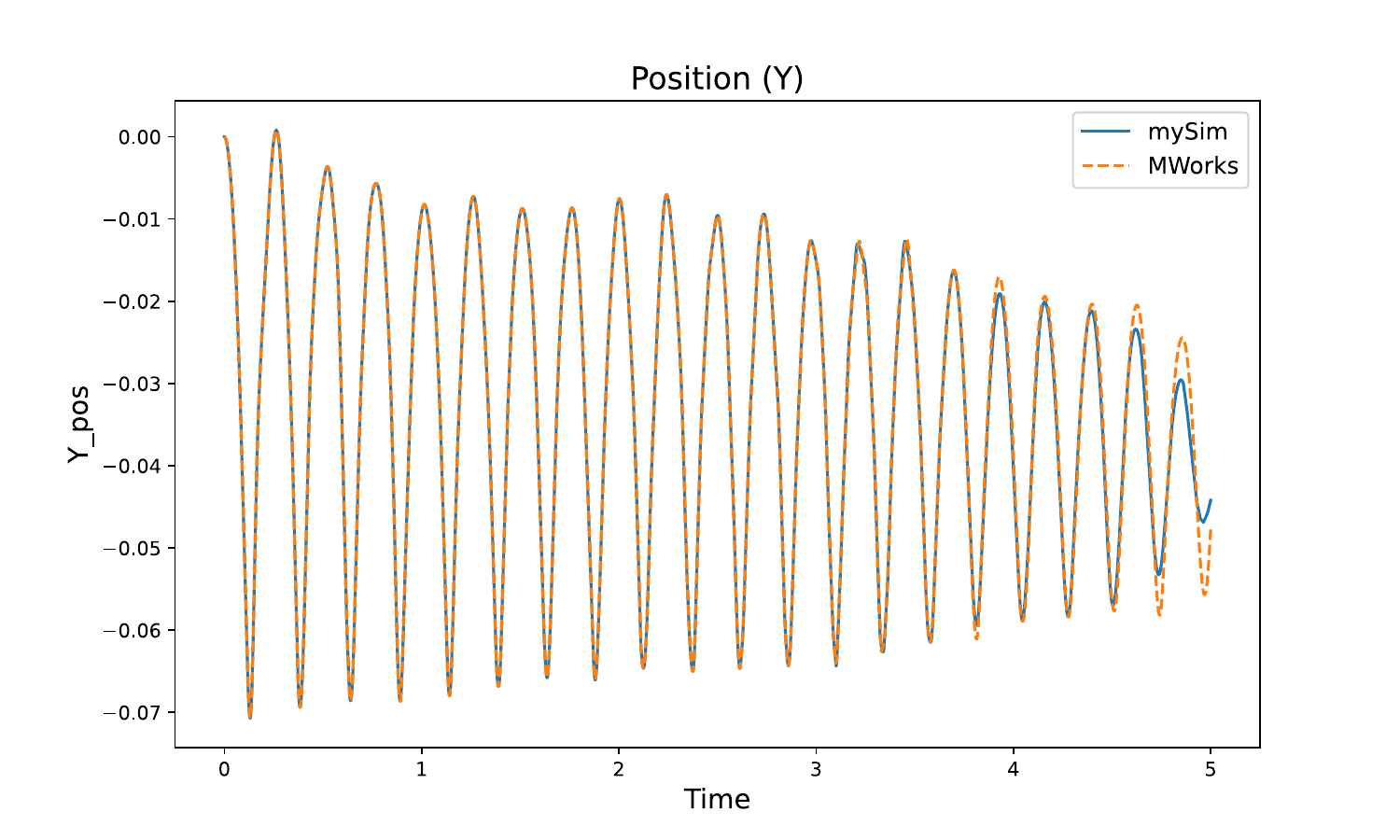}}
\hfill
\subfloat[\centering]{\includegraphics[width=5.5cm]{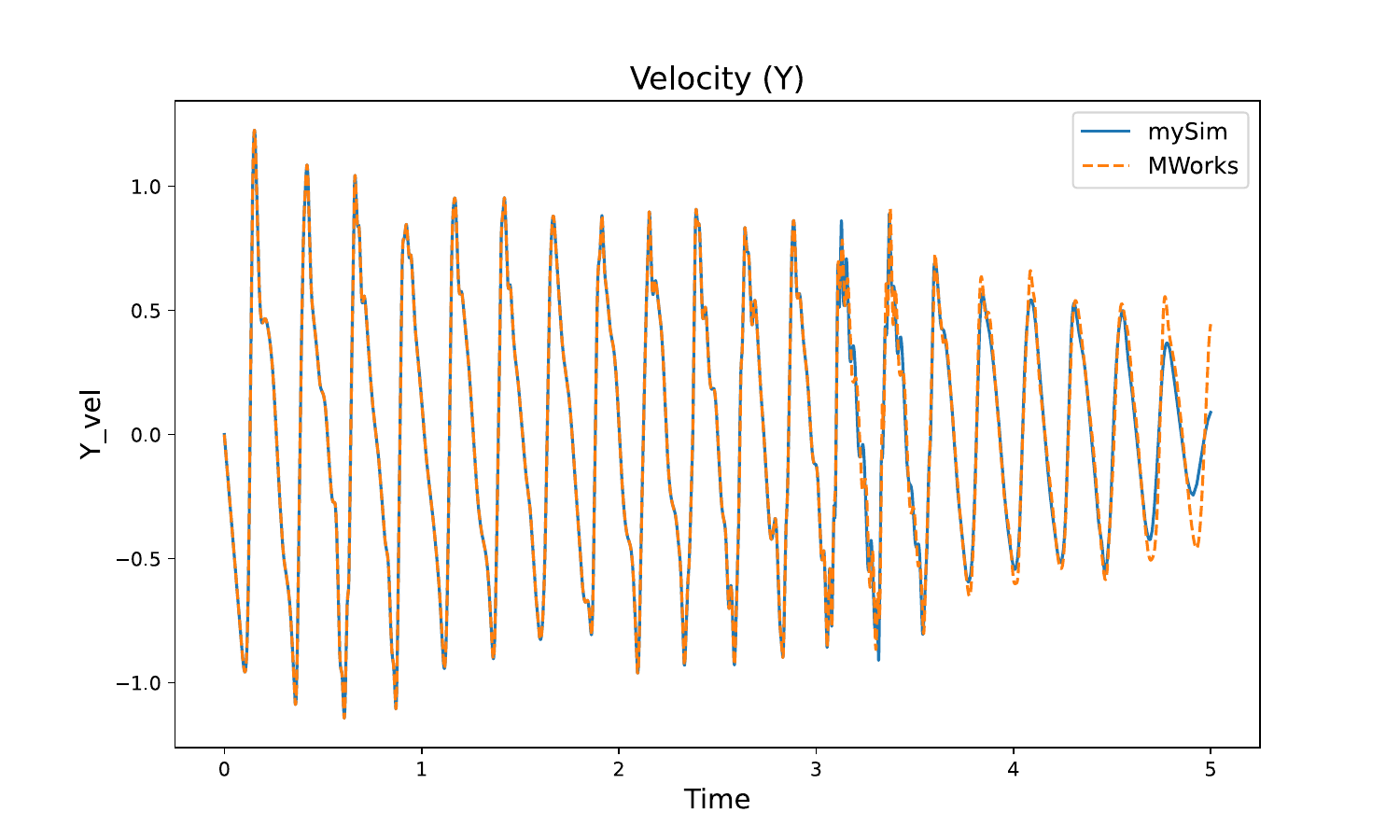}}
\hfill
\subfloat[\centering]{\includegraphics[width=5.5cm]{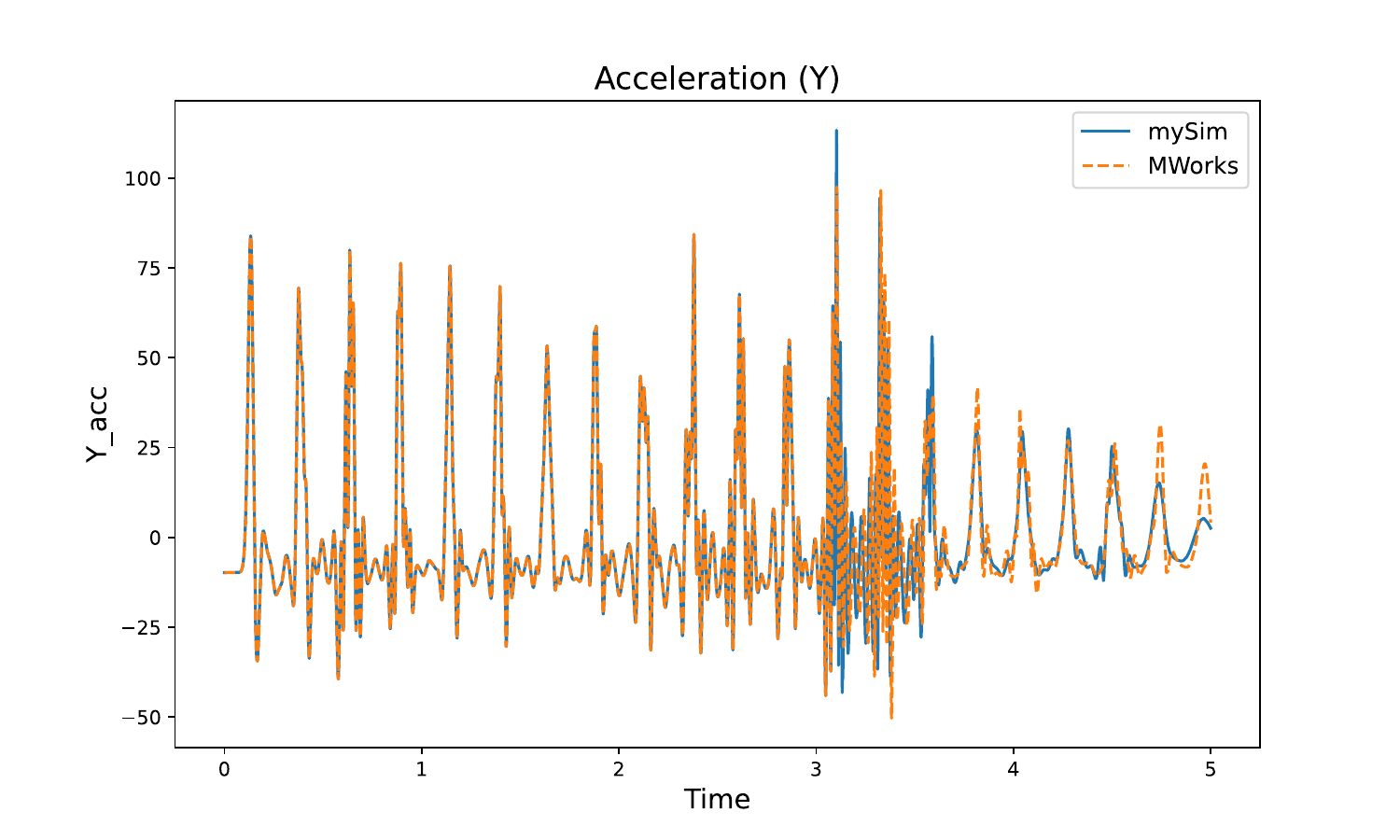}}

\end{adjustwidth}
\caption{Comparison of mySim and MWorks results for the simple rope net, where (\textbf{a}) shows the payload's position variation, (\textbf{b}) shows payload's velocity variation, and (\textbf{c}) shows payload's acceleration variation.
\label{fig:results_rope_net_model_sim}}
\end{figure}

Simulation results comparing mySim and MWorks for the rope net are provided in Fig.~\ref{fig:results_rope_net_model_sim}, which includes payload position, velocity, and acceleration variations. These results demonstrate a high degree of agreement between the two platforms. The similar trends and matching numerical behavior confirm that mySim can accurately simulate the dynamic characteristics of the rope module. This validates the correctness of the spring-damper model implementation.

\begin{figure}
\centering
% \begin{adjustwidth}{-\extralength}{0cm}
\subfloat[\centering]{\includegraphics[width=6.5cm]{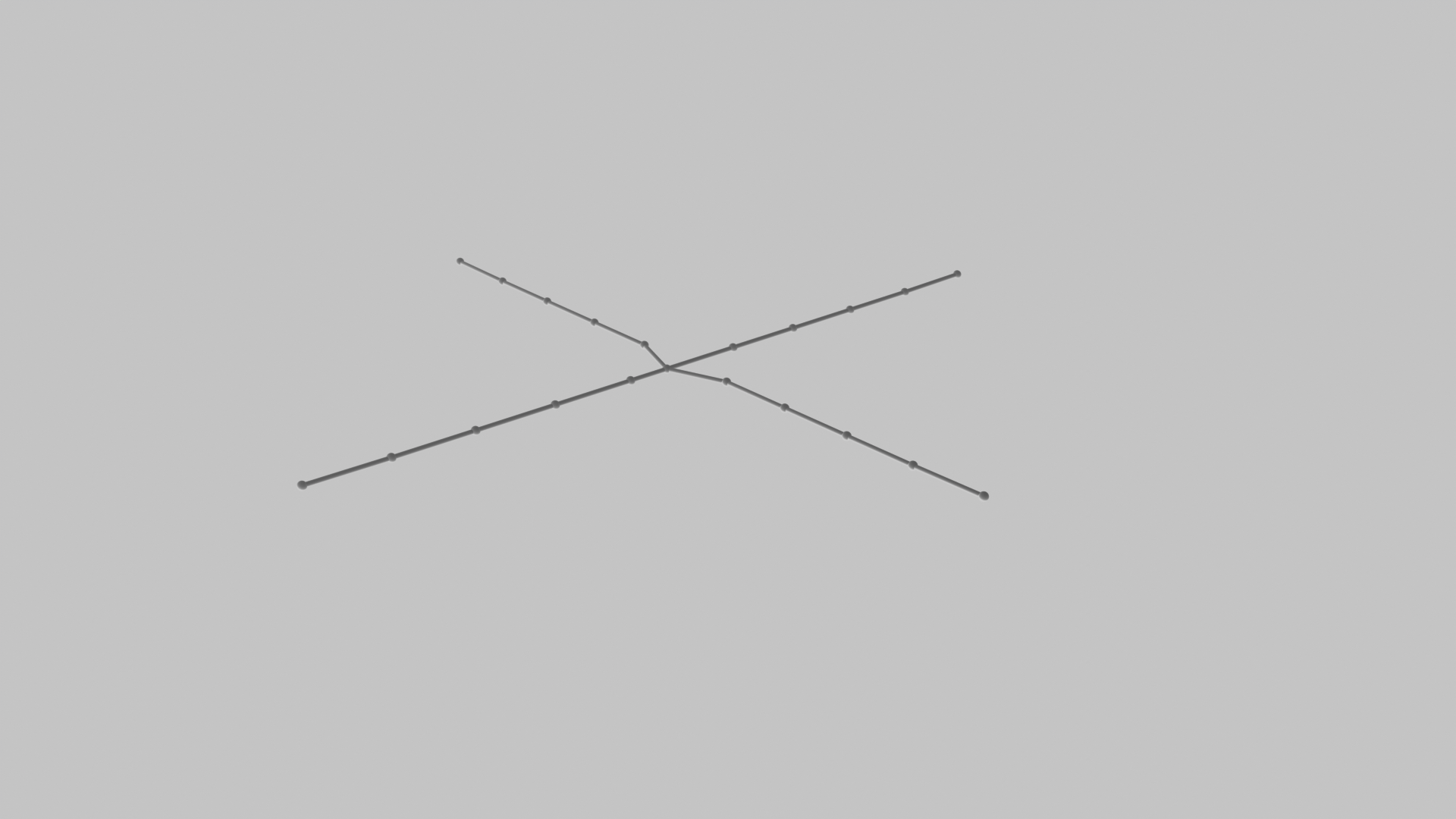}}
\hfill
\subfloat[\centering]{\includegraphics[width=6.5cm]{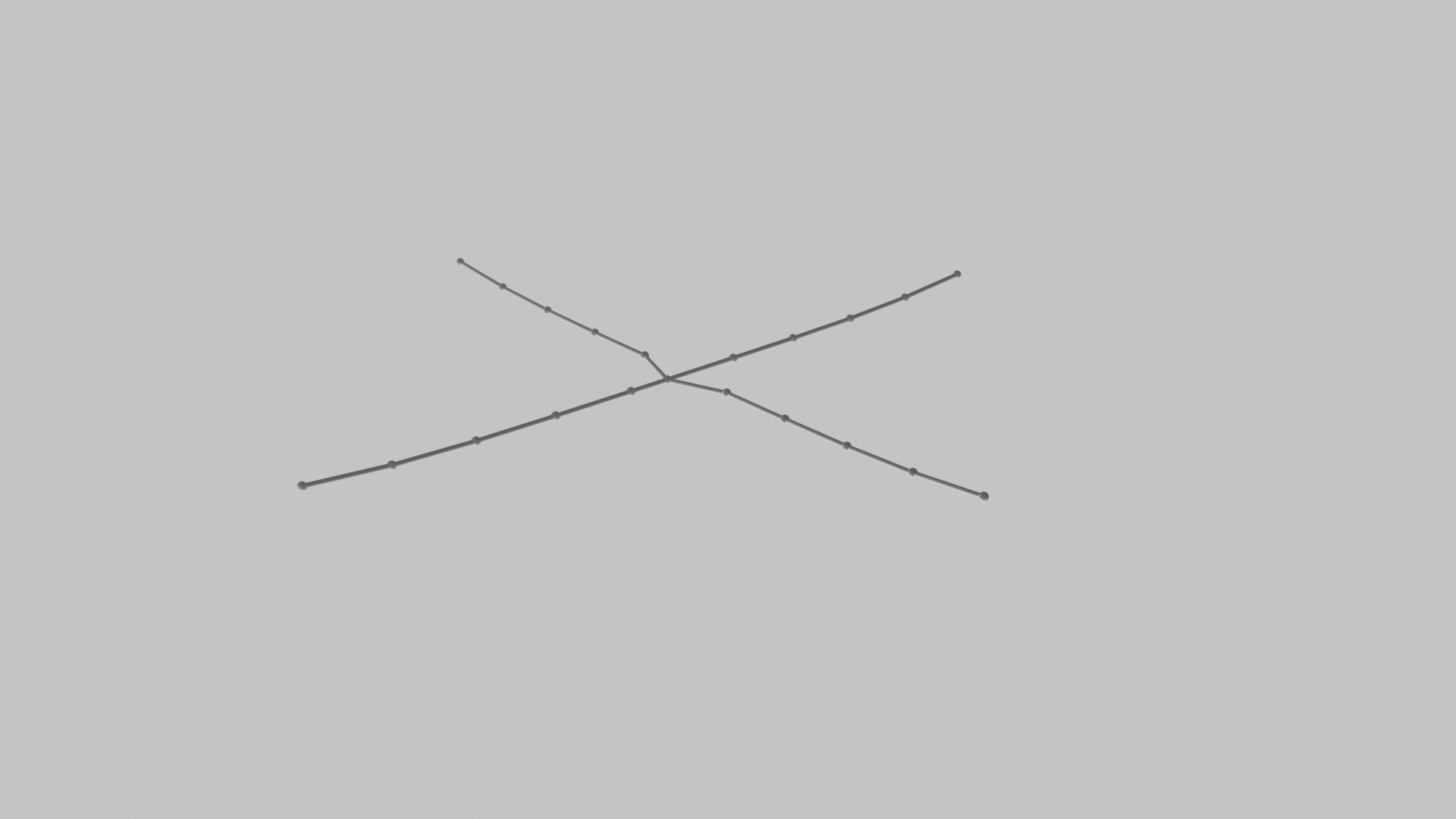}}\\
\subfloat[\centering]{\includegraphics[width=6.5cm]{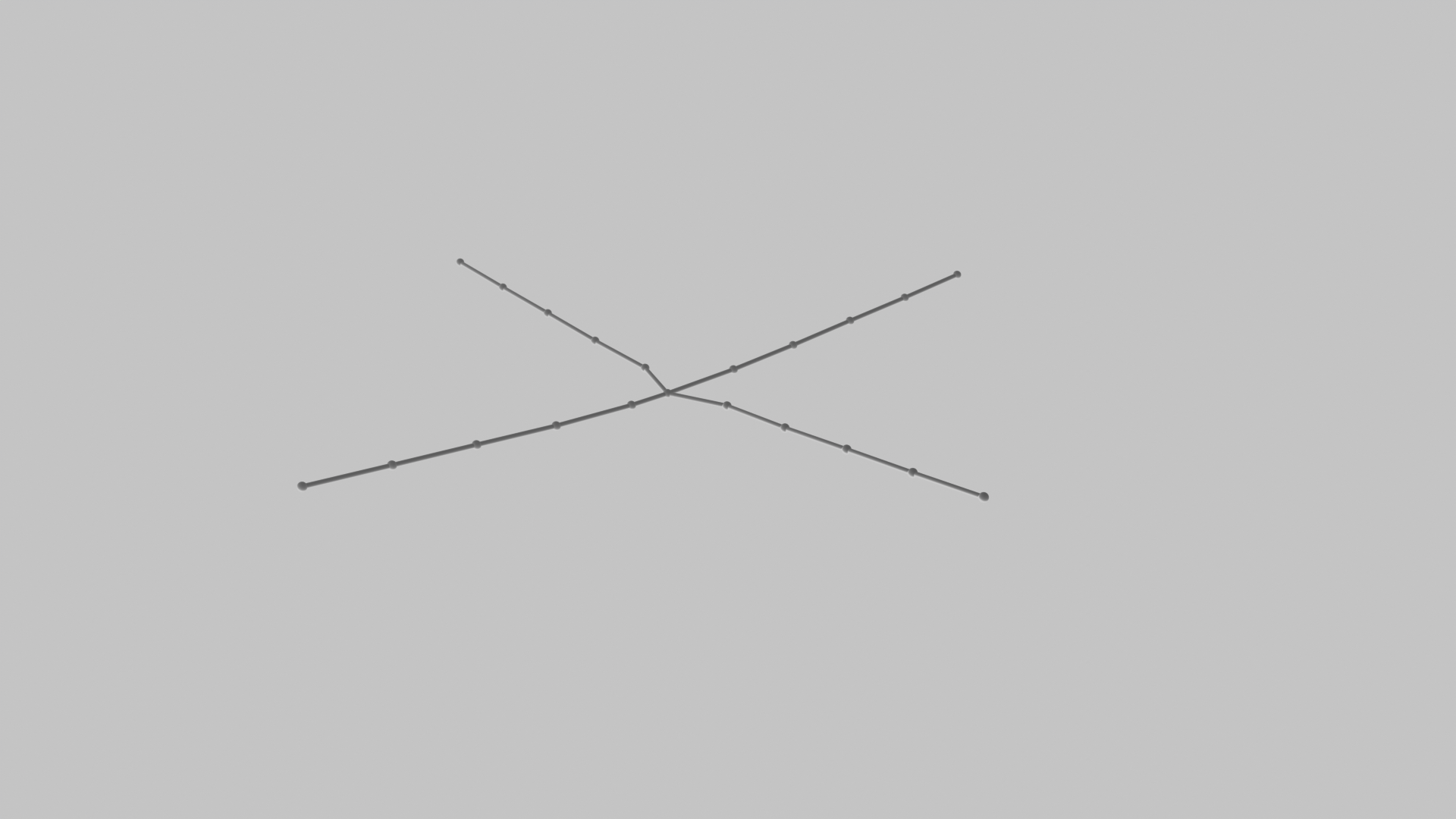}}
\hfill
\subfloat[\centering]{\includegraphics[width=6.5cm]{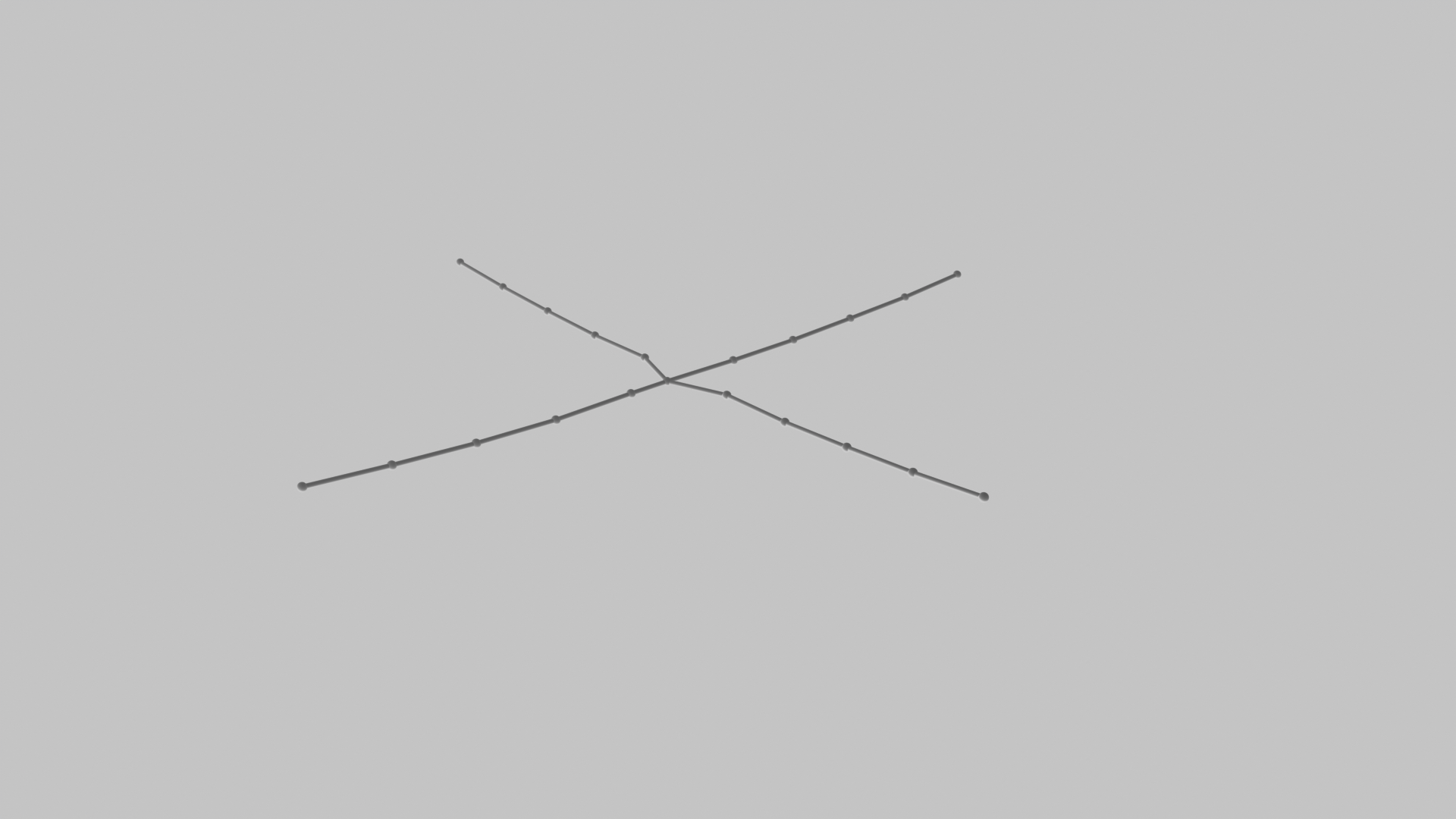}}\\
% \end{adjustwidth}
\caption{Visualization of the simple rope net model, where the centrally rendered sphere represents the payload as shown in Fig.~\ref{fig:rope_net_model_demonstration}, while other spheres indicate mass points within the net module. The sequence (\textbf{a})-(\textbf{d}) shows the position variations during motion.
\label{fig:results_rope_net_model_vis}}
\end{figure}

To further demonstrate the simulation accuracy, Fig.~\ref{fig:results_rope_net_model_vis} provides a visualized time sequence of the rope net's deformation during payload motion. This visualization clearly shows the dynamic behavior and deformation process, confirming the expected physical response of the model. These results verify the feasibility and correctness of the rope module implementation in mySim.

Following the validation of the rope module, the collision module can be verified based on the setup shown in Fig.~\ref{fig:collision_demonstration} and simulated the system according to Fig.~\ref{fig:collision}. The contact force models, as defined in Eq.~\eqref{eq:compact_force} and Eq.~\eqref{eq:collsion miu-v}, were configured with parameters from Table.~\ref{table:collision_module}, while the net model containing collision bodies was parameterized according to Table.~\ref{table:collision_example}.

\begin{figure}
\centering
\begin{adjustwidth}{-\extralength}{0cm}
\subfloat[\centering]{\includegraphics[width=5.5cm]{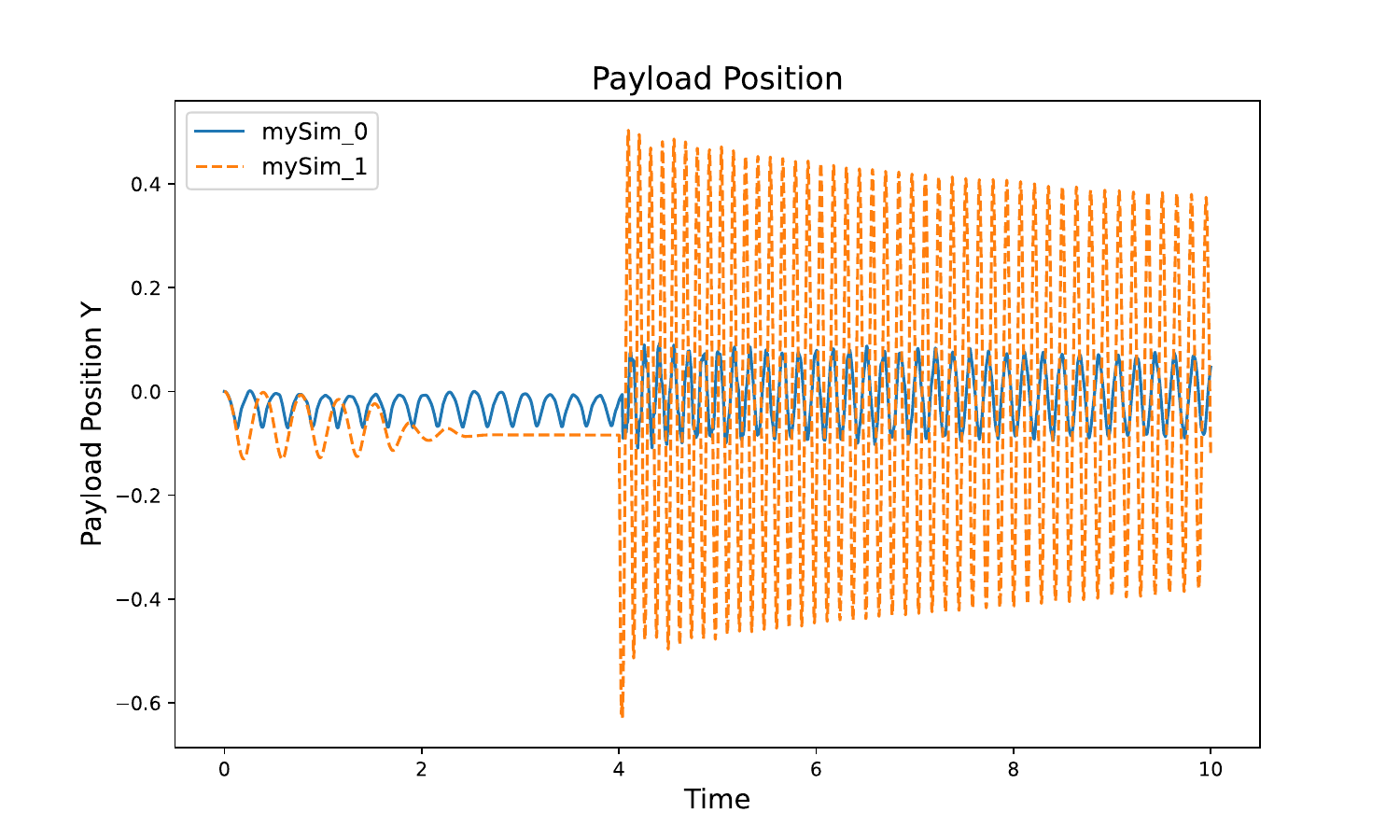}}
\hfill
\subfloat[\centering]{\includegraphics[width=5.5cm]{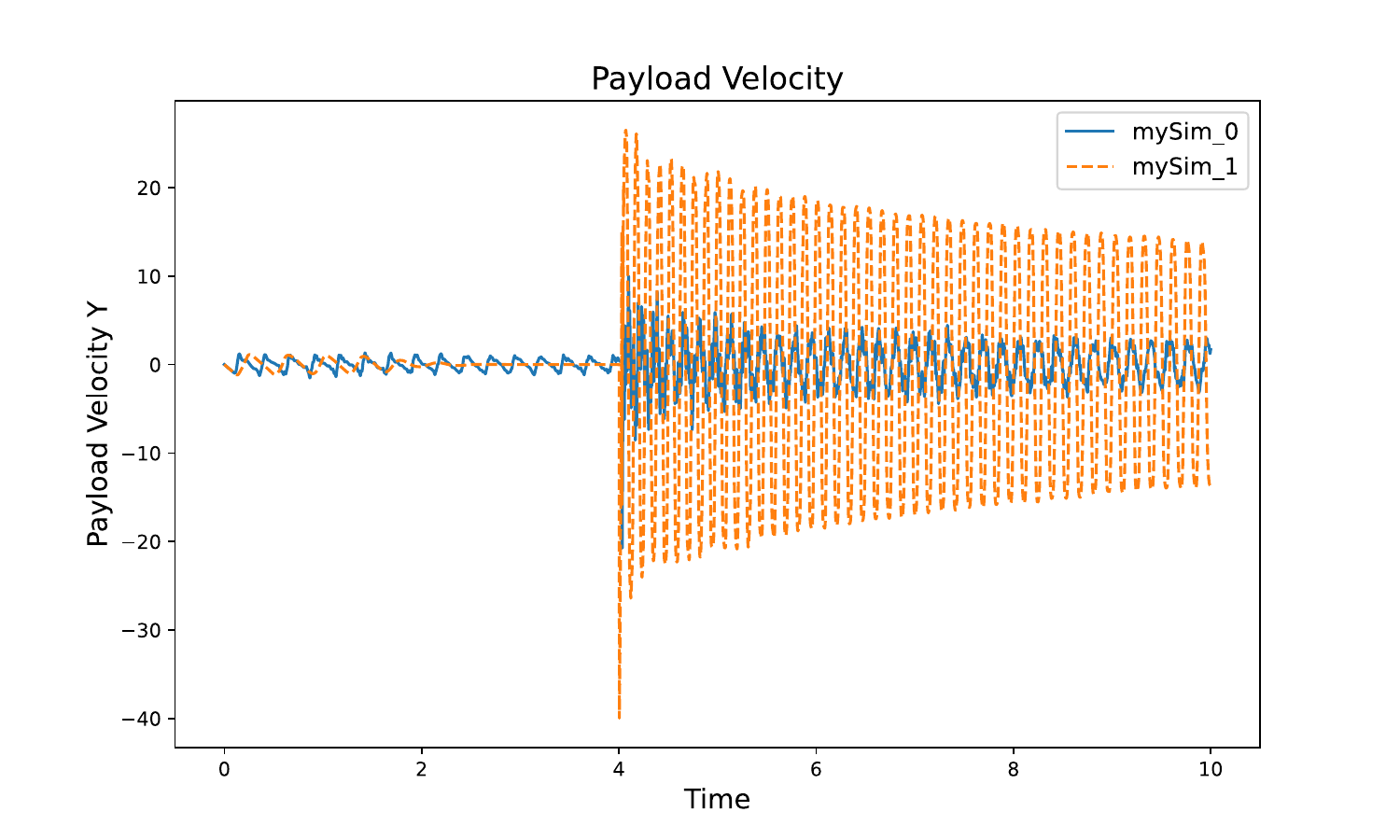}}
\hfill
\subfloat[\centering]{\includegraphics[width=5.5cm]{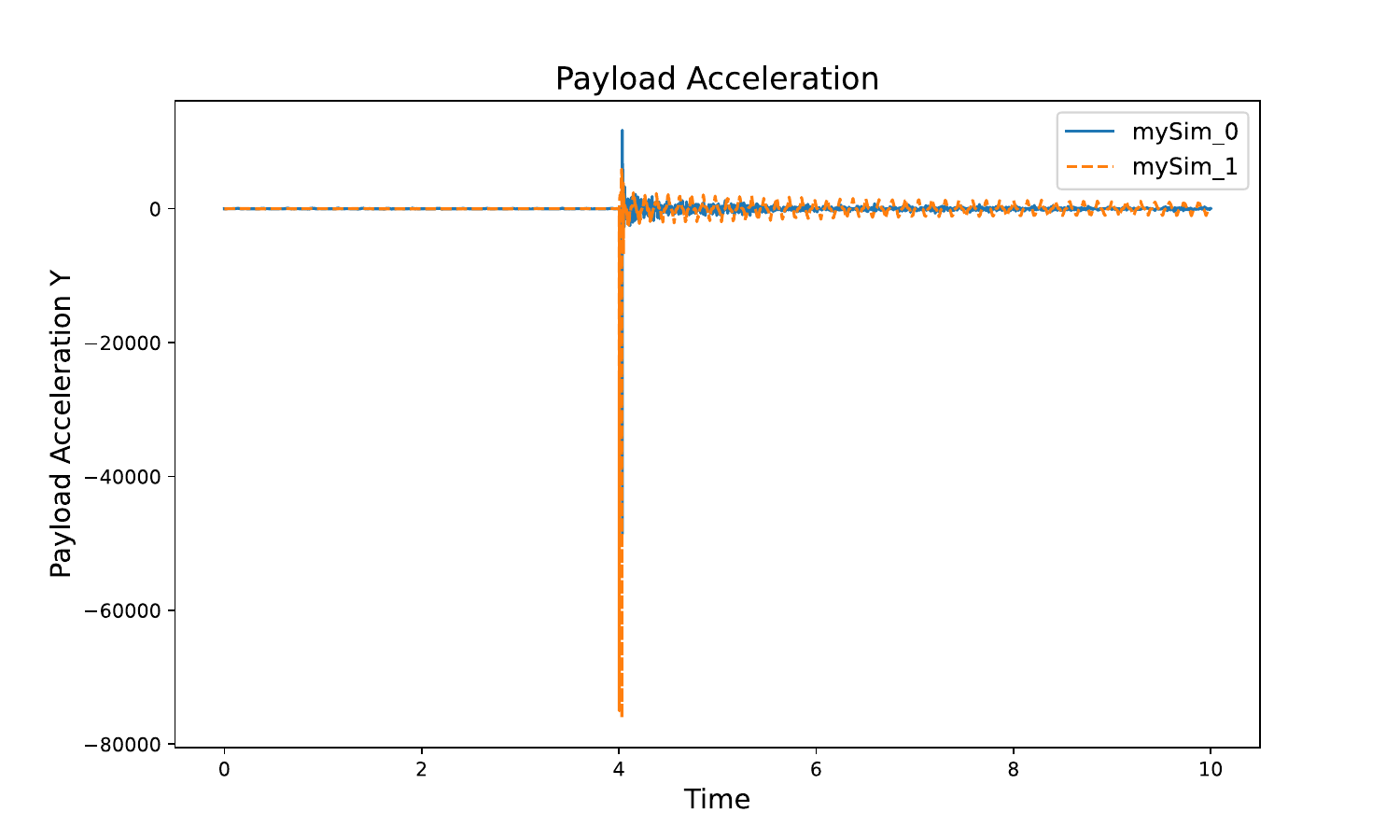}}\\
\subfloat[\centering]{\includegraphics[width=5.5cm]{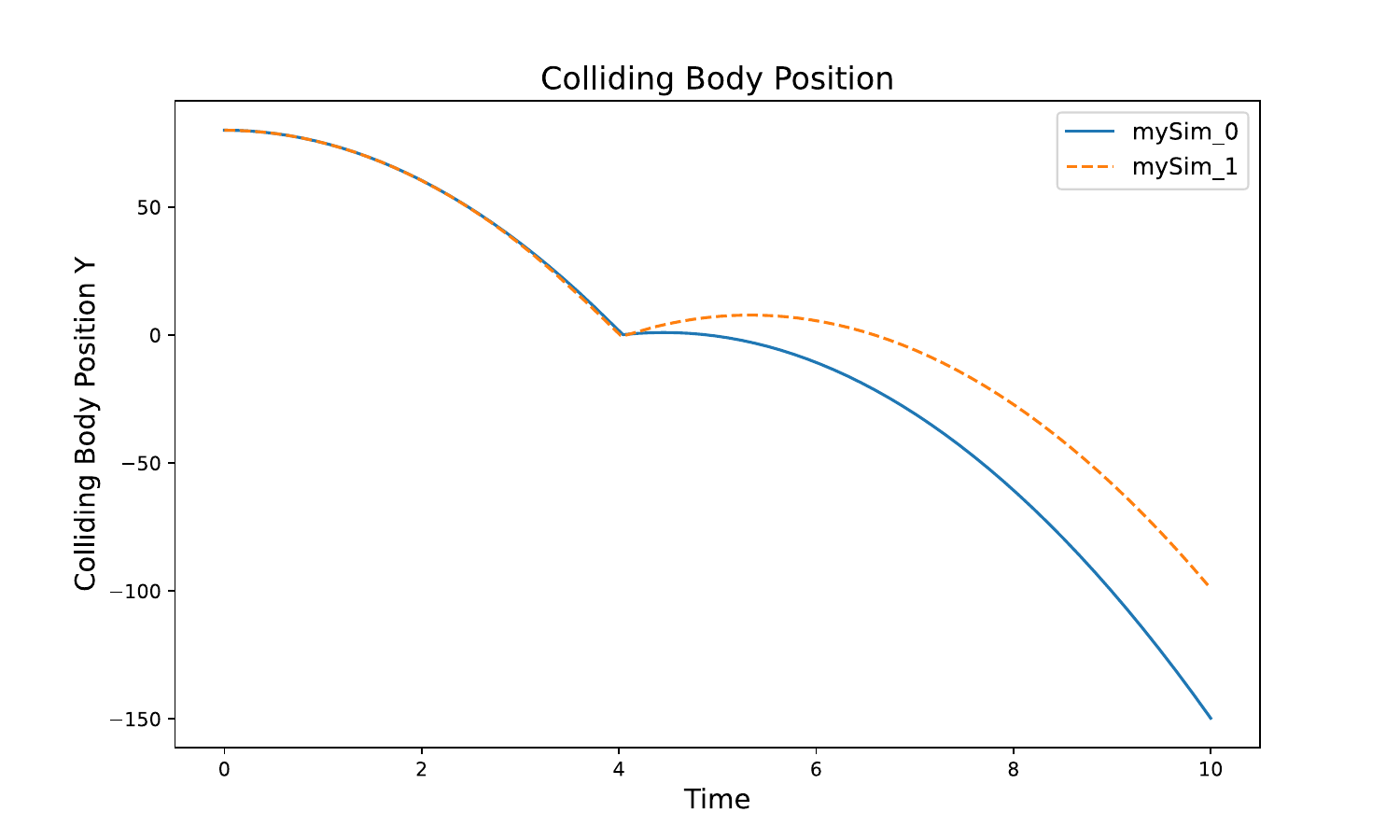}}
\hfill
\subfloat[\centering]{\includegraphics[width=5.5cm]{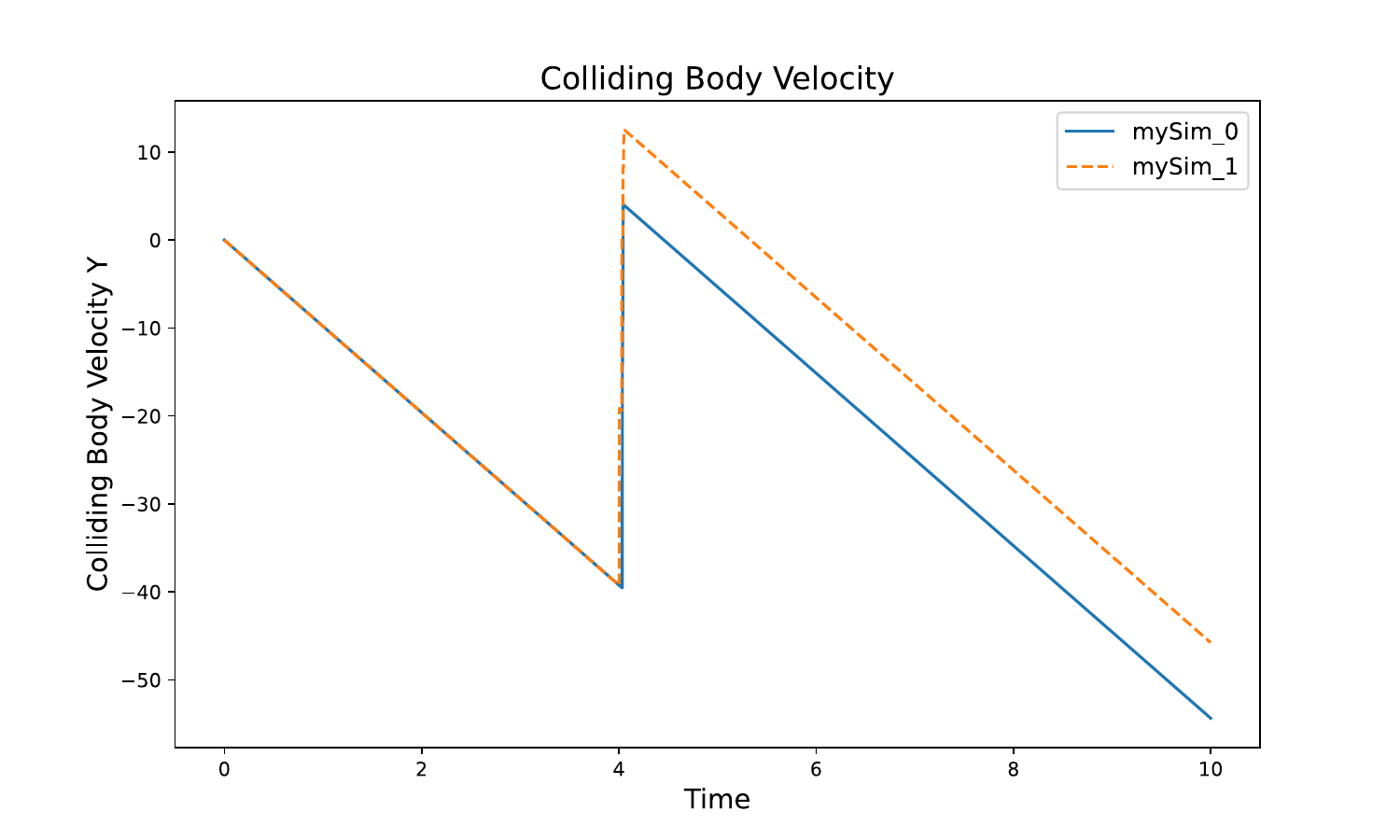}}
\hfill
\subfloat[\centering]{\includegraphics[width=5.5cm]{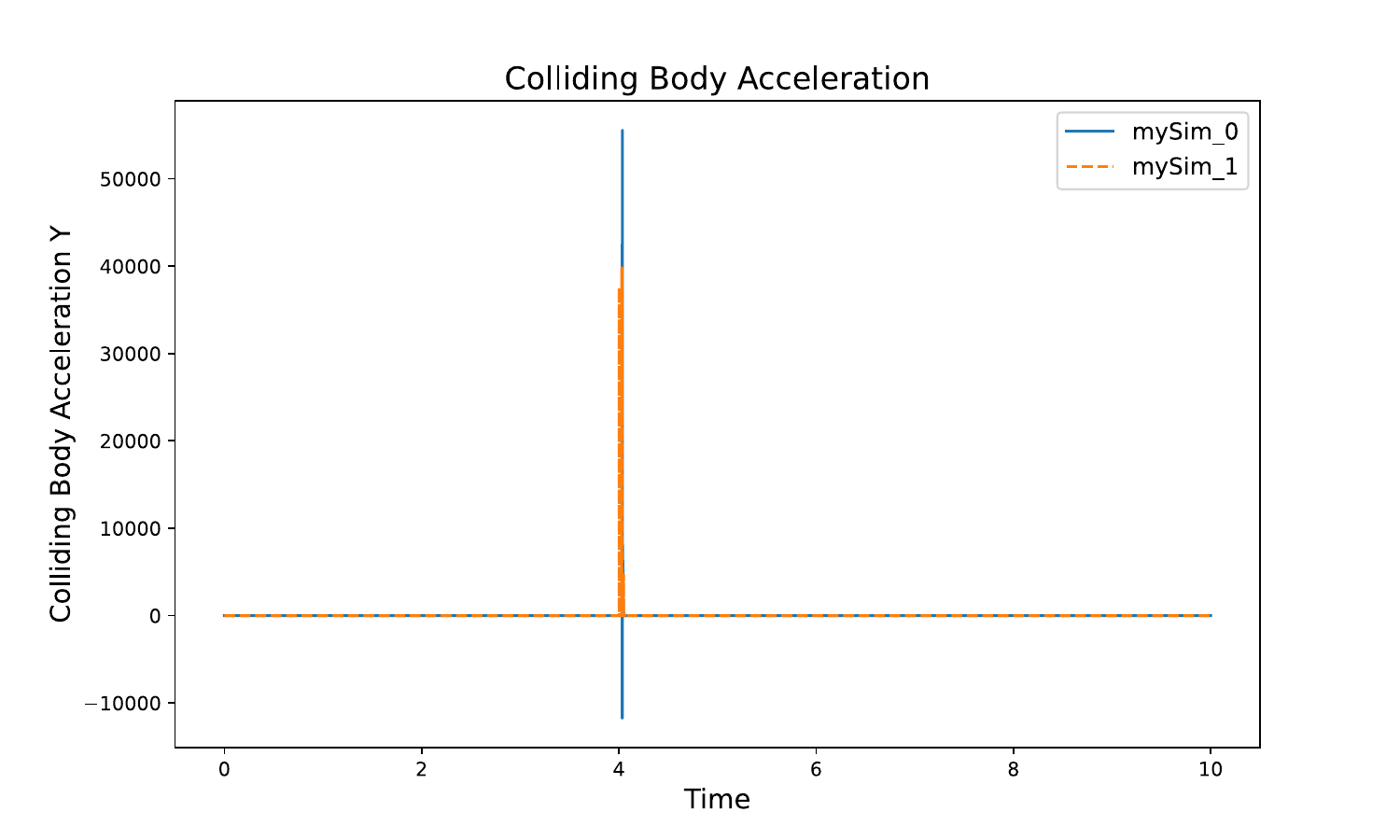}}

\end{adjustwidth}
\caption{
Comparison of two collision cases configured as in Table.~\ref{table:collision_example}, where (\textbf{a}) shows the payload's position variation, (\textbf{b}) shows payload's velocity variation, (\textbf{c}) shows payload's position variation, (\textbf{d}) shows the position variation of the collision object, (\textbf{e}) shows collision object's velocity variation, and (\textbf{f}) shows collision object's acceleration variation.
\label{fig:results_collision_sim}}
\end{figure}

The simulation results of two cases are presented in Fig.~\ref{fig:results_collision_sim}, where Fig.~\ref{fig:results_collision_sim}(\textbf{a})-(\textbf{c}) display the payload's position, velocity, and acceleration, and Fig.~\ref{fig:results_collision_sim}(\textbf{d})-(\textbf{f}) illustrate the same variables for the colliding body. The results indicate that in both collision scenarios, the dynamic responses exhibit expected oscillatory behaviors and appropriate damping, depending on the mass of the colliding object. This confirms that the collision model incorporating non-linear spring-damper and friction effects is functioning correctly.

\begin{table}
\caption{Parameters for the collision module.\label{table:collision_module}}
\newcolumntype{Y}{>{\raggedright\arraybackslash}X}
\begin{tabularx}{\textwidth}{Y c c c}
\toprule
\textbf{Parameter} & \textbf{Symbol} & \textbf{Value} & \textbf{Unit} \\
\midrule
Convex sphere radius & \( r_I \) & \( 3.00 \times 10^{-2} \) & \( m \) \\
Contact force stiffness coefficient & \( k \) & \( 1.00 \times 10^8 \) & \( N \cdot m^{-1} \) \\
Contact force damping coefficient & \( d \) & \( 1.00 \times 10^4 \) & \( N \cdot s \cdot m^{-1} \) \\
Stiffness index & \( n \) & \( 1.00 \times 10^0 \) & - \\
Maximum penetration depth & \( p \) & \( 1.00 \times 10^{-4} \) & \( m \) \\
Dynamic friction coefficient & \( \mu_d \) & \( 3.00 \times 10^{-2} \) & - \\
Static friction coefficient & \( \mu_s \) & \( 4.00 \times 10^{-2} \) & - \\
Static friction limit velocity & \( v_s \) & \( 1.00 \times 10^{-2} \) & \( m \cdot s^{-1} \) \\
Dynamic friction limit velocity & \( v_d \) & \( 1.00 \times 10^{-1} \) & \( m \cdot s^{-1} \) \\
\bottomrule
\end{tabularx}
\end{table}
\begin{table}
\caption{Parameters for the collision example.\label{table:collision_example}}
\newcolumntype{Y}{>{\raggedright\arraybackslash}X}
\begin{tabularx}{\textwidth}{Y c c c}
\toprule
\textbf{Parameter} & \textbf{Symbol} & \textbf{Value} & \textbf{Unit} \\
\midrule
Axial stretching stiffness & \( EA \) & \( 1.00 \times 10^5 \) & \( N \) \\
Linear damping coefficient & \( d \) & \( 5.00 \times 10^1 \) & \( N \cdot s \cdot m^{-1} \) \\
Number of concentrated masses & \( N \) & \( 5.00 \times 10^0 \) & - \\
Initial rope length & \( L \) & \( 1.00 \times 10^0 \) & \( m \) \\
Point mass & \( m_i \) & \( 2.00 \times 10^{-1} \) & \( kg \) \\
Payload mass & \( m_{\text{payload}} \) & \( <2.00 \times 10^{-1}, 1.00 \times 10^1> \) & \( kg \) \\
Collision object mass & \( m_{\text{contact}} \) & \( <2.00 \times 10^{-1}, 2.00 \times 10^1> \) & \( kg \) \\
\bottomrule
\end{tabularx}
\end{table}

\begin{figure}
\centering
\begin{adjustwidth}{-\extralength}{0cm}
\subfloat[\centering]{\includegraphics[width=4cm]{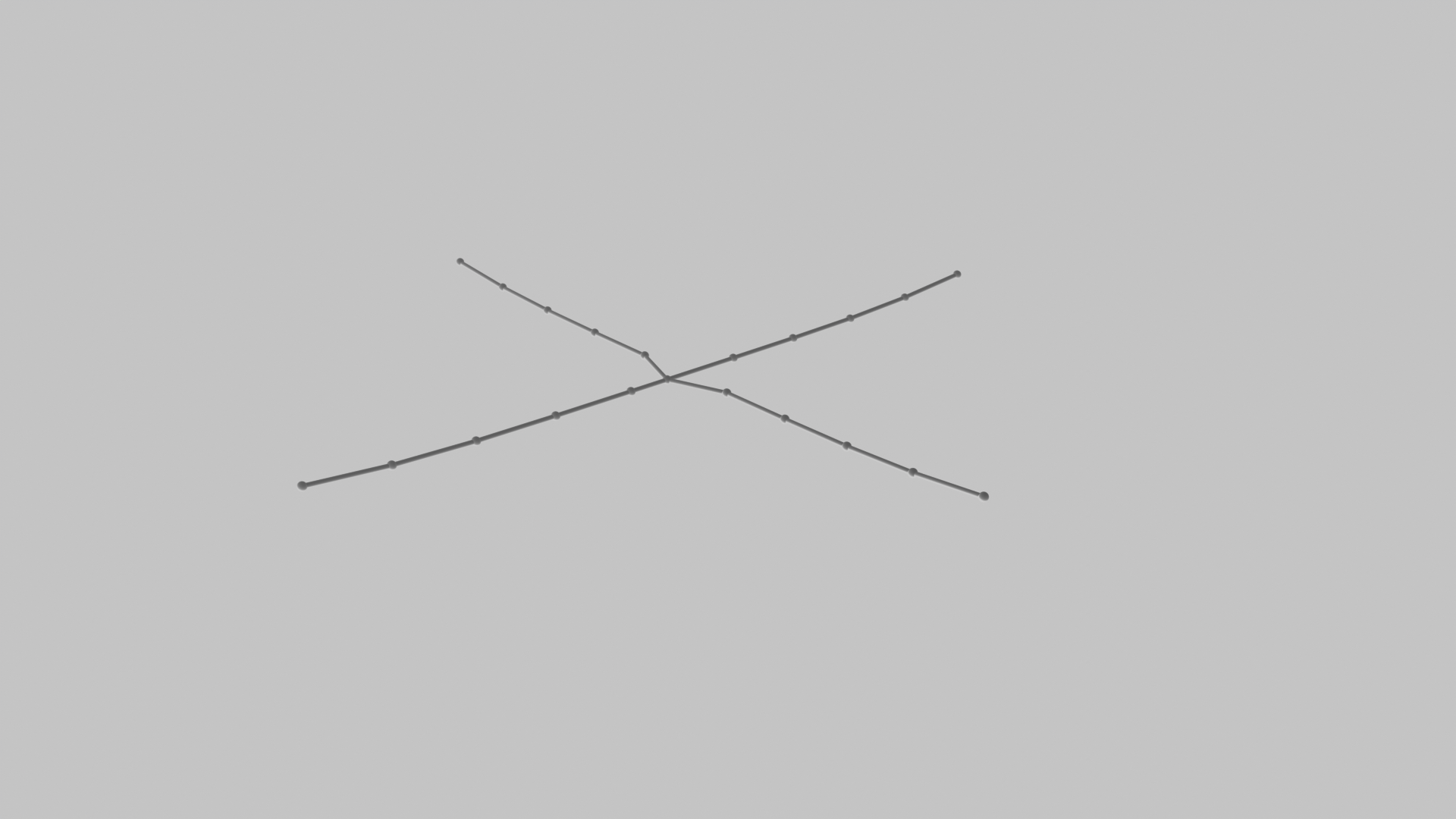}}
\hfill
\subfloat[\centering]{\includegraphics[width=4cm]{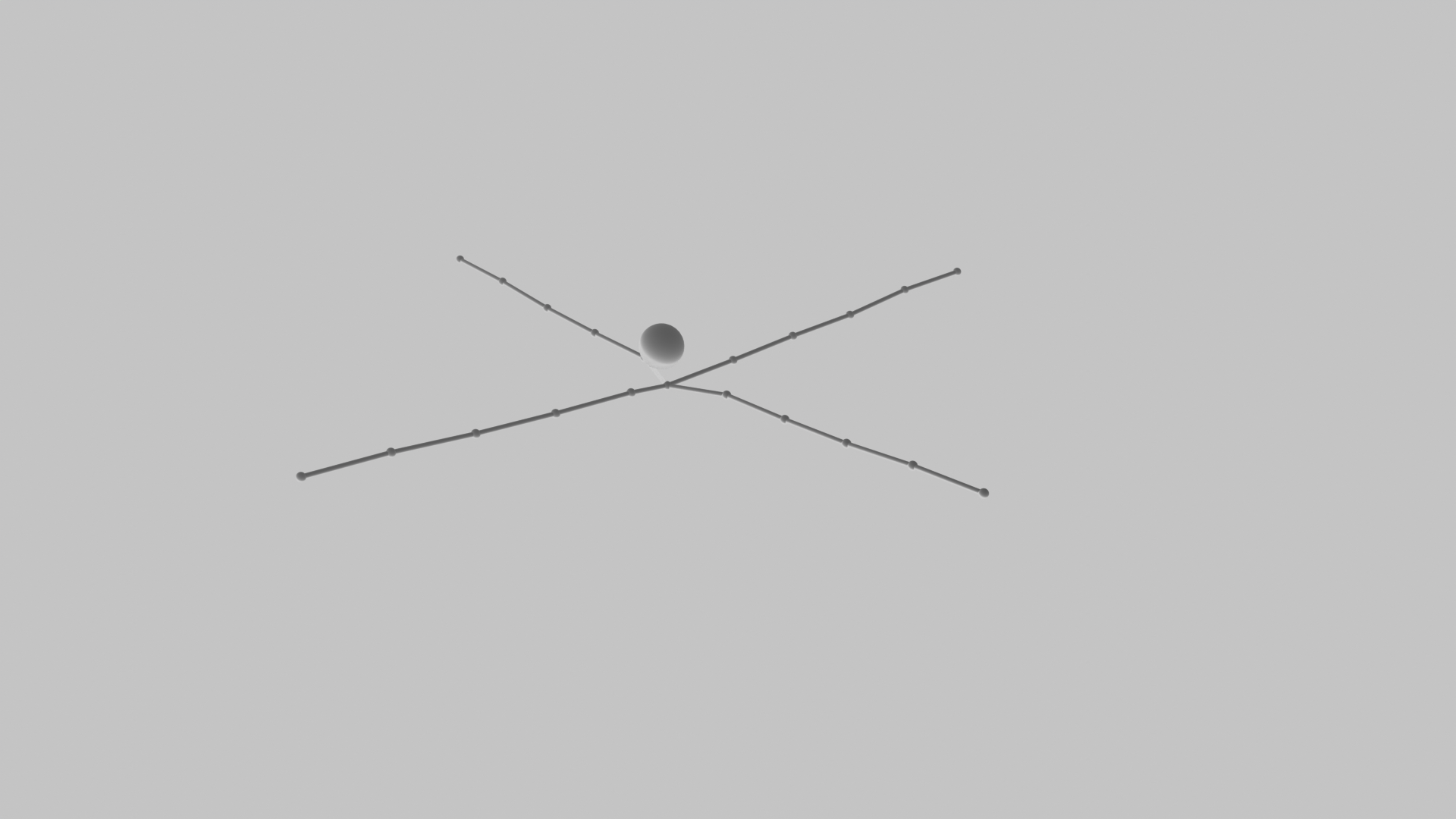}}
\hfill
\subfloat[\centering]{\includegraphics[width=4cm]{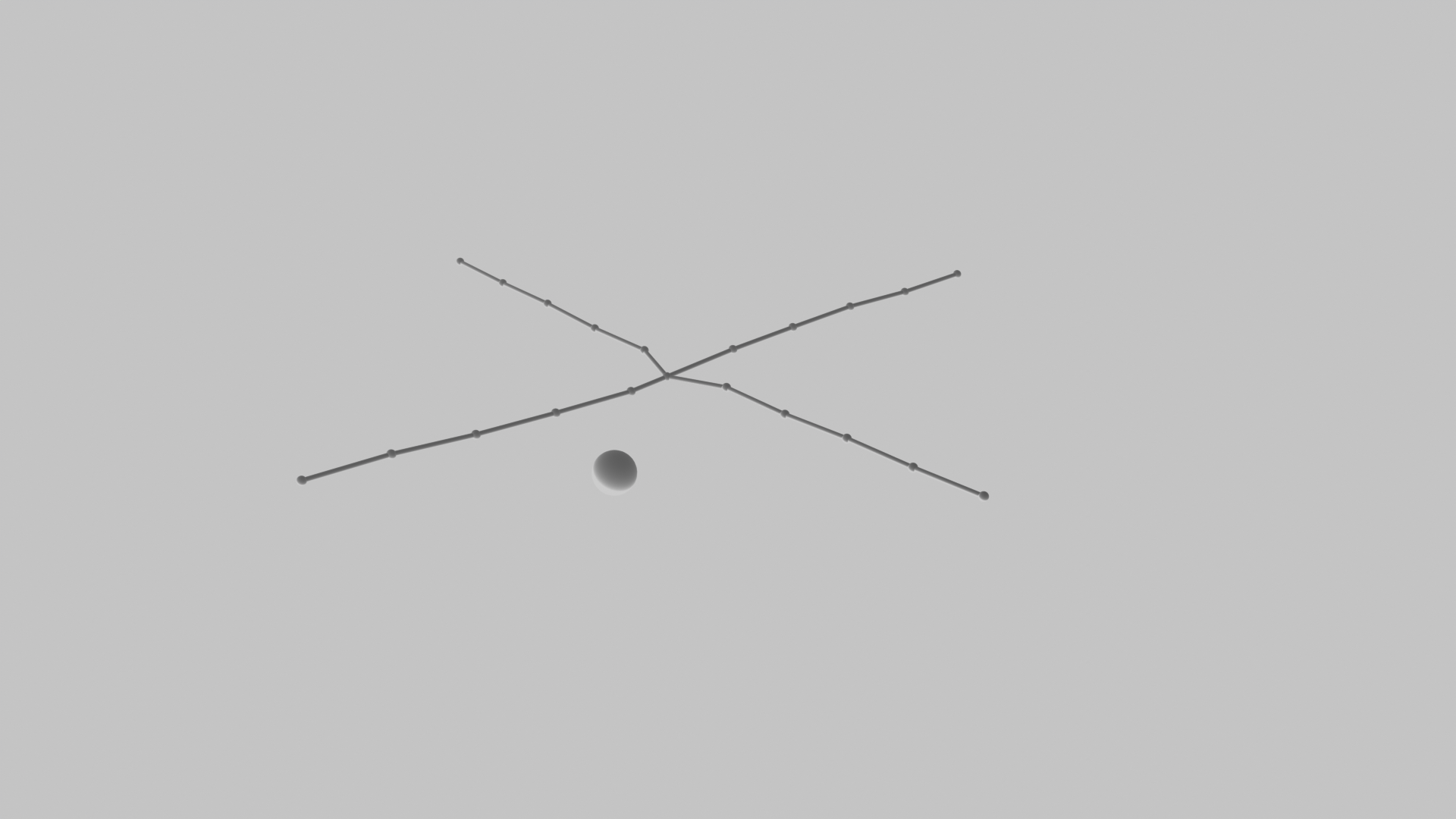}}
\hfill
\subfloat[\centering]{\includegraphics[width=4cm]{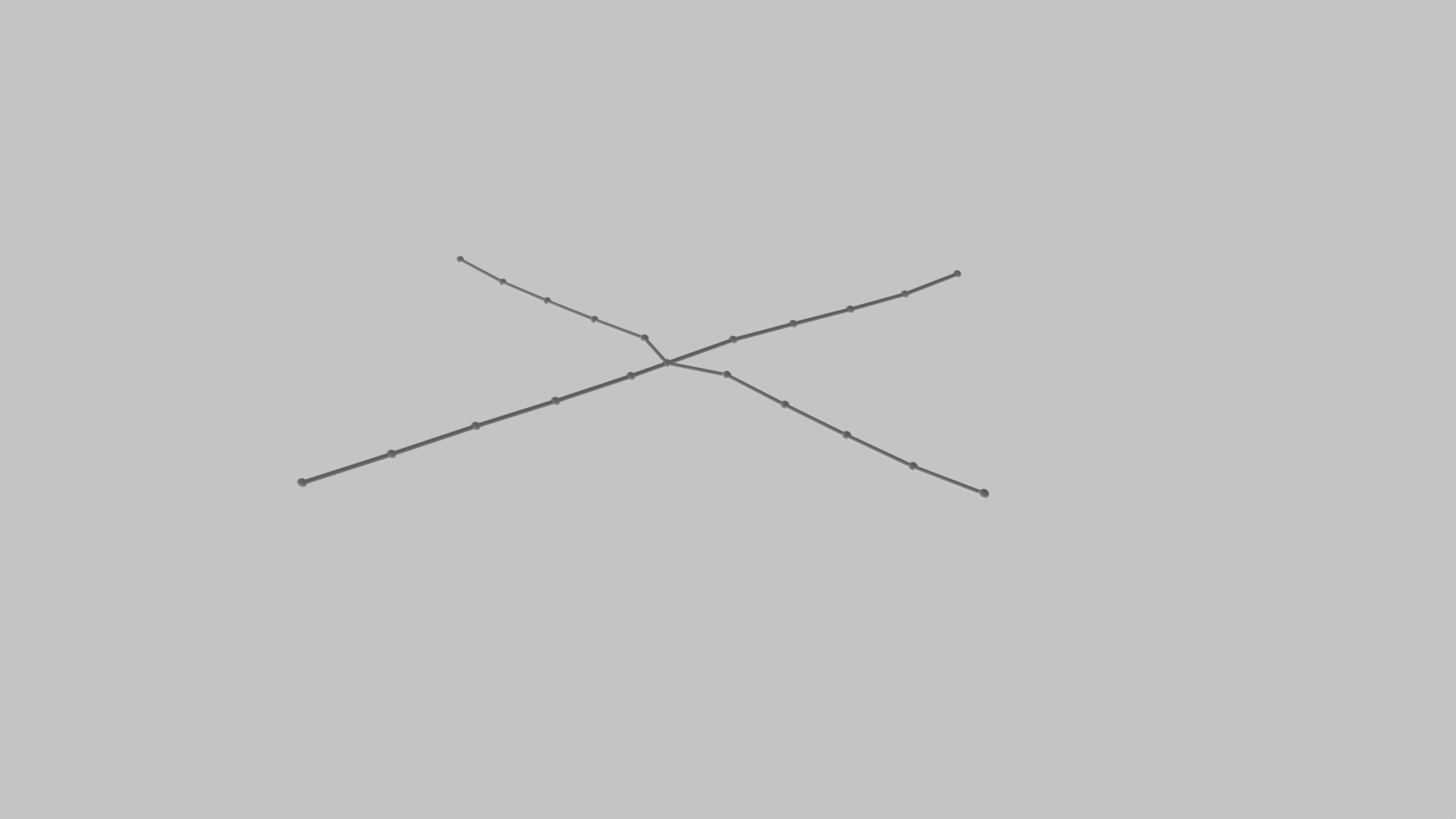}}\\
\subfloat[\centering]{\includegraphics[width=4cm]{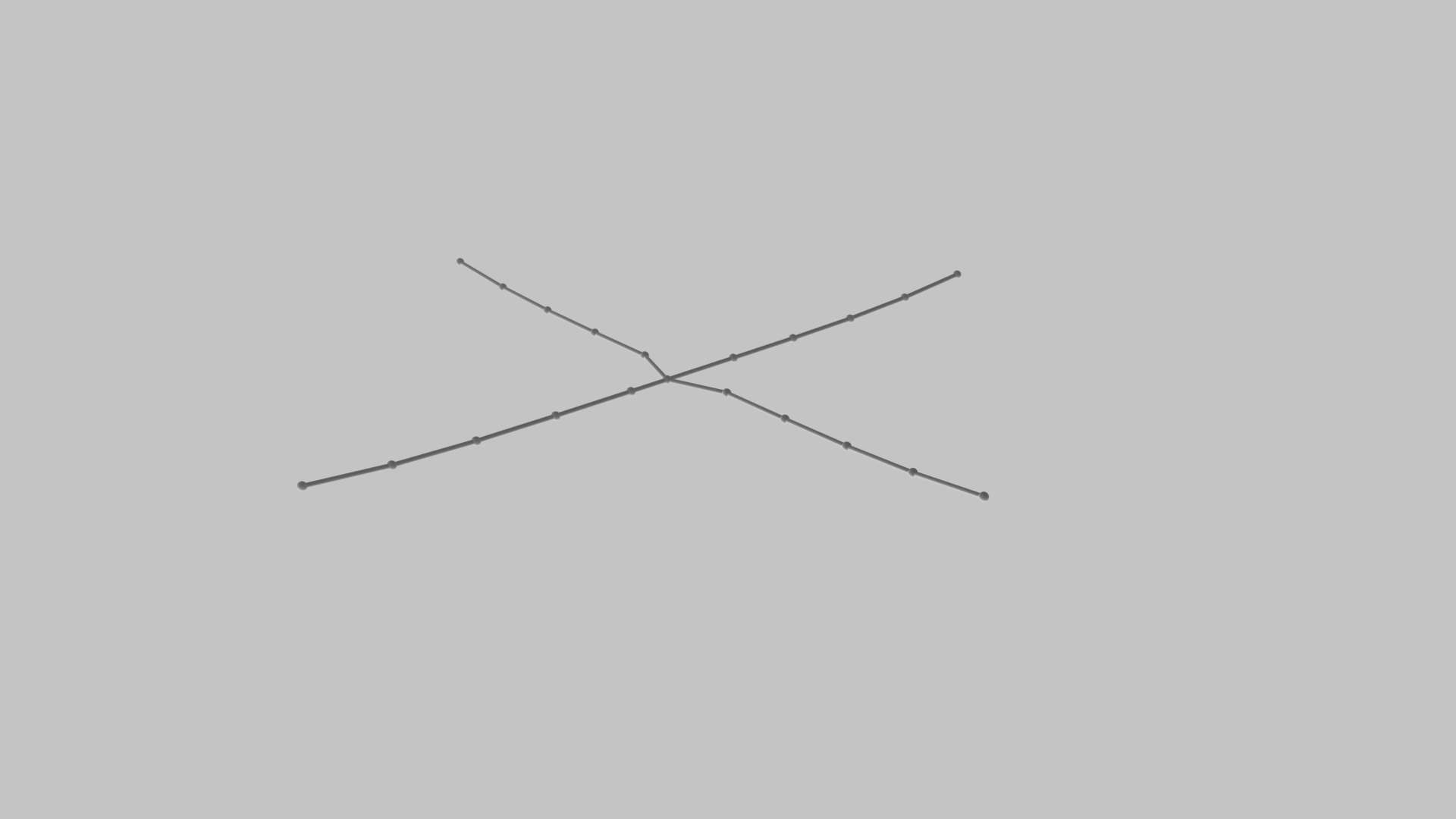}}
\hfill
\subfloat[\centering]{\includegraphics[width=4cm]{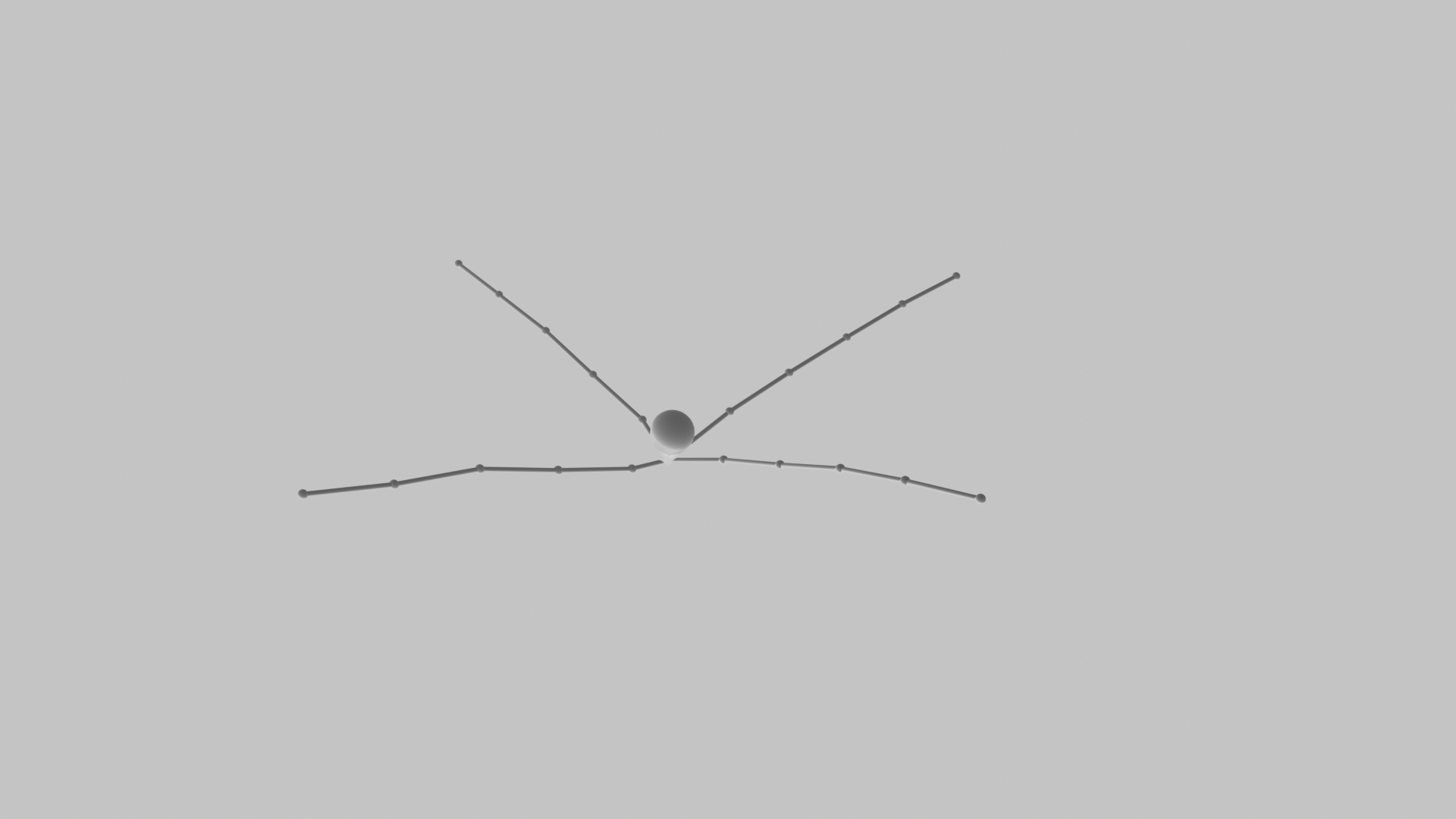}}
\hfill
\subfloat[\centering]{\includegraphics[width=4cm]{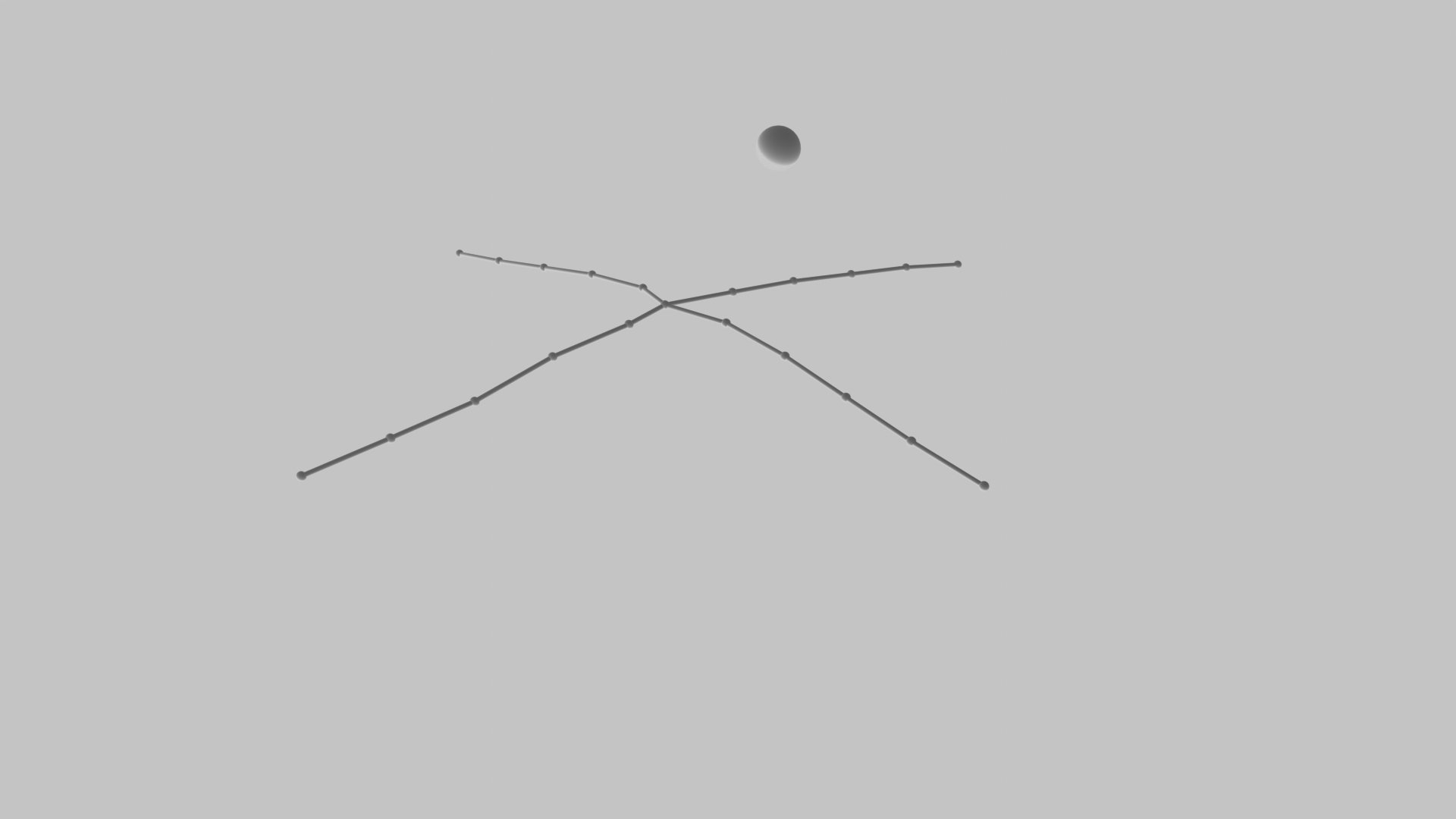}}
\hfill
\subfloat[\centering]{\includegraphics[width=4cm]{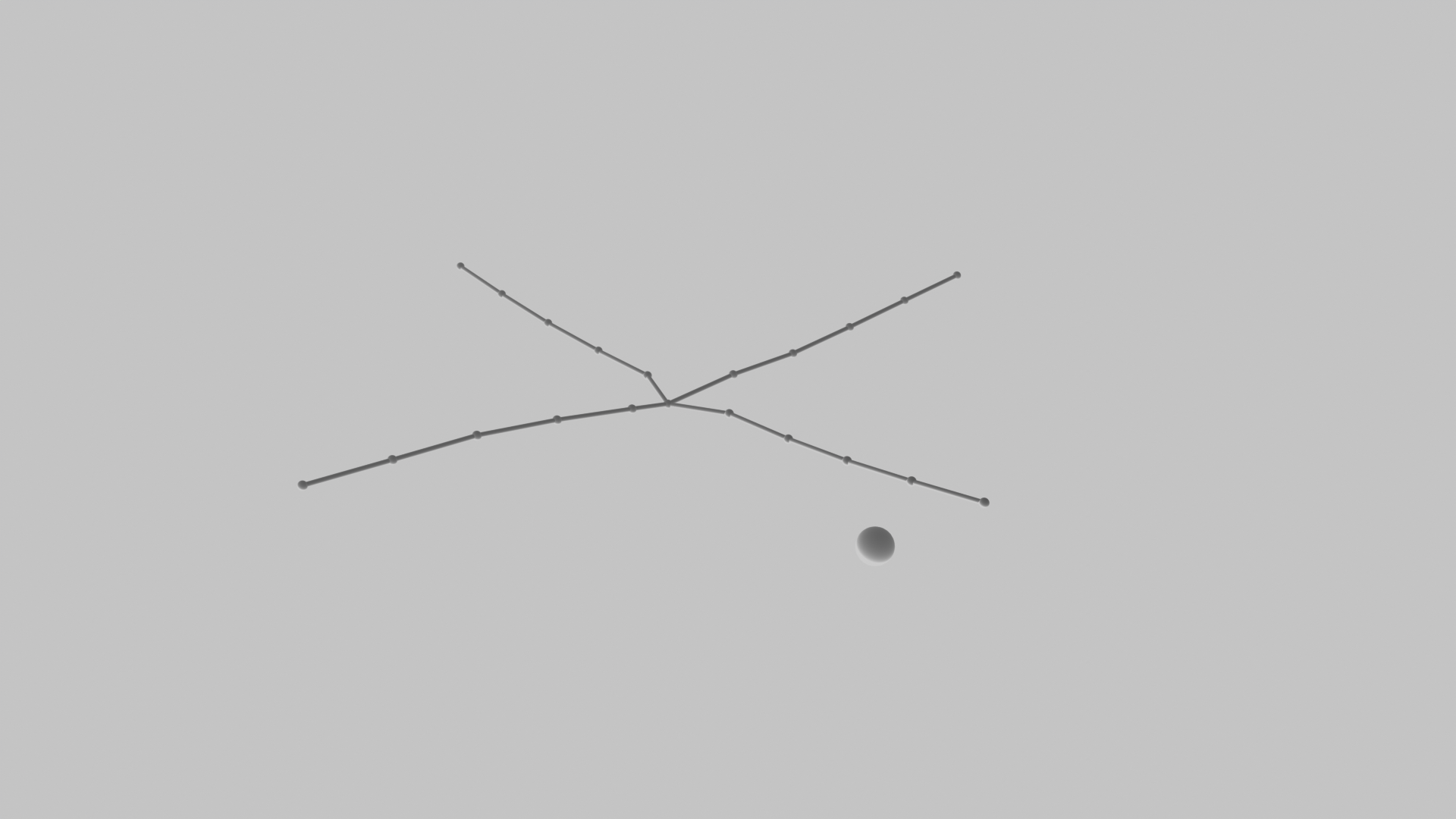}}

\end{adjustwidth}
\caption{Visualization of the rope and collision modules, where the central sphere represents the payload, and other spheres indicate mass points in the rope module. Small-mass groups show motion changes in (\textbf{a})-(\textbf{d}), while large-mass groups show motion sequences in (\textbf{e})-(\textbf{h}).
\label{fig:results_collision_vis}}
\end{figure}

Fig.~\ref{fig:results_collision_vis} provides a visual depiction of these collision scenarios, with sequential snapshots showing the net's response and the colliding body's interaction. The upper row corresponds to a small-mass colliding body, while the lower row represents a heavier object. The visual results are consistent with the motion curves in Fig.~\ref{fig:results_collision_sim} and validate the capability of mySim to simulate collision interactions accurately.

To verify the feasibility of UAV control, we implemented the controller in Fig.~\ref{fig:entire_controller_pd} following Alg.~\ref{alg:controller_process_pd}. The UAV’s dynamics parameters were set according to Table.~\ref{table:UAV_model}, while the controller parameters were configured as per Table.~\ref{table:PID_controller}.

\begin{figure}
\centering
\begin{adjustwidth}{-\extralength}{0cm}
\subfloat[\centering]{\includegraphics[width=8cm]{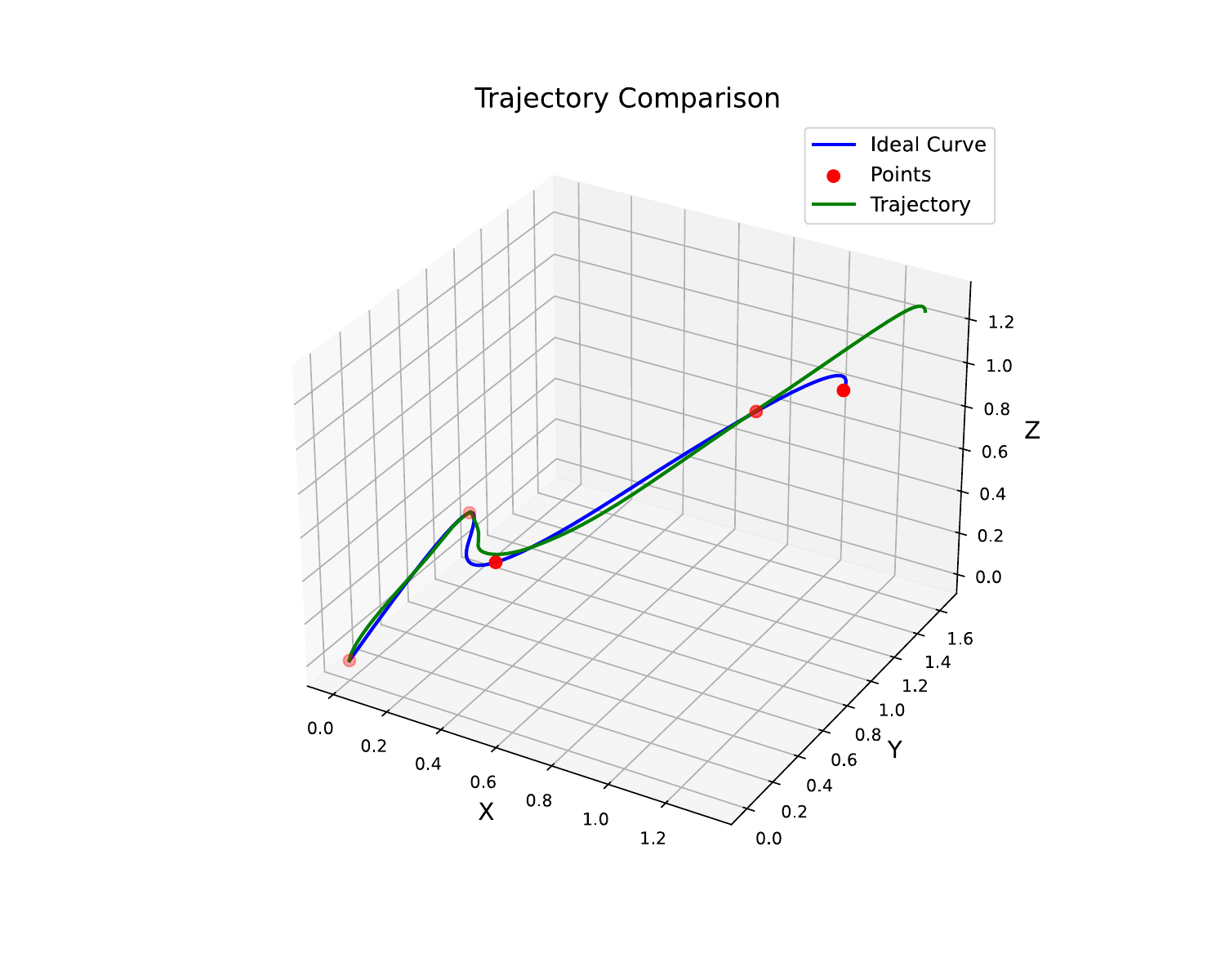}}
\hfill
\subfloat[\centering]{\includegraphics[width=8cm]{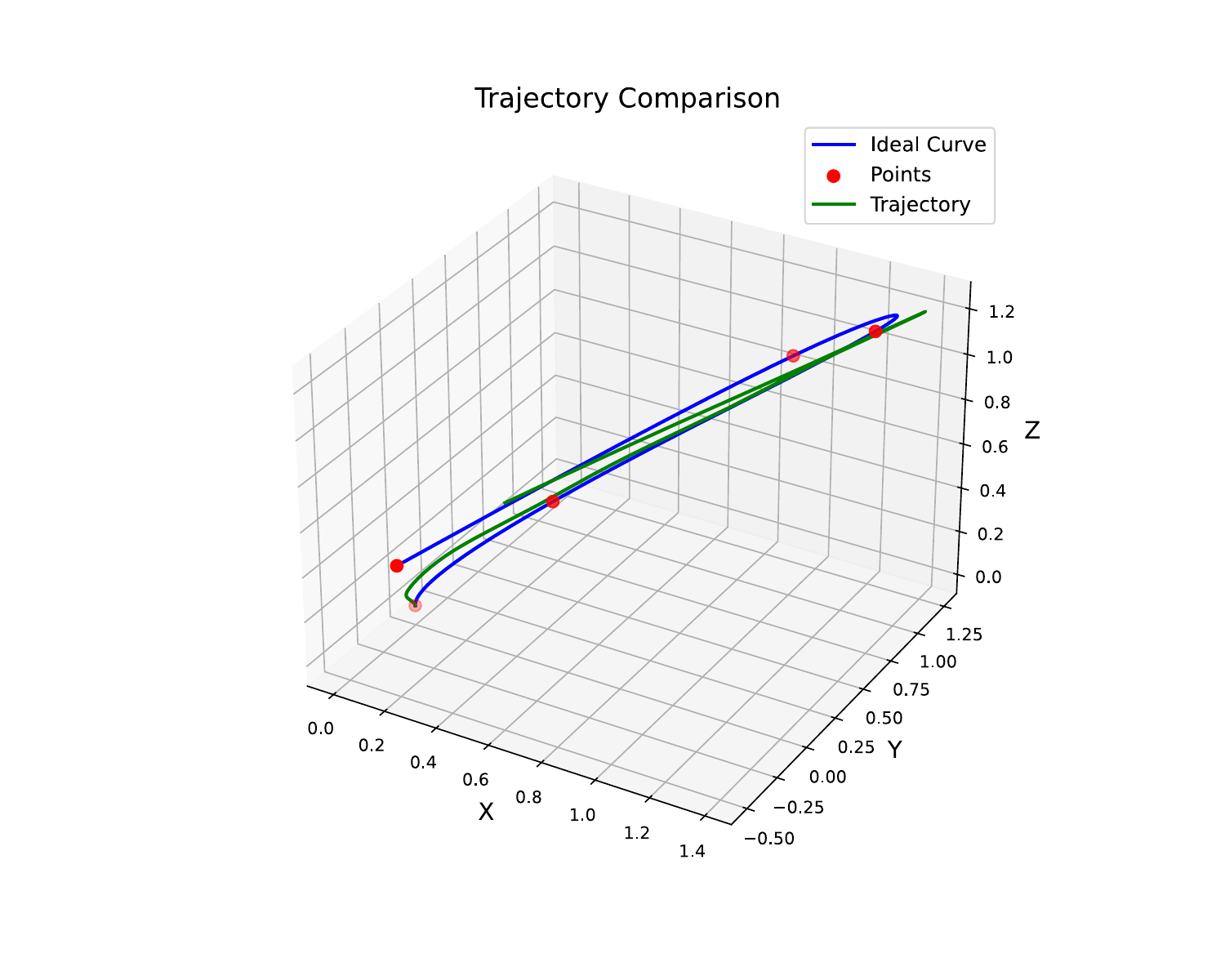}}\\

\subfloat[\centering]{\includegraphics[width=8cm]{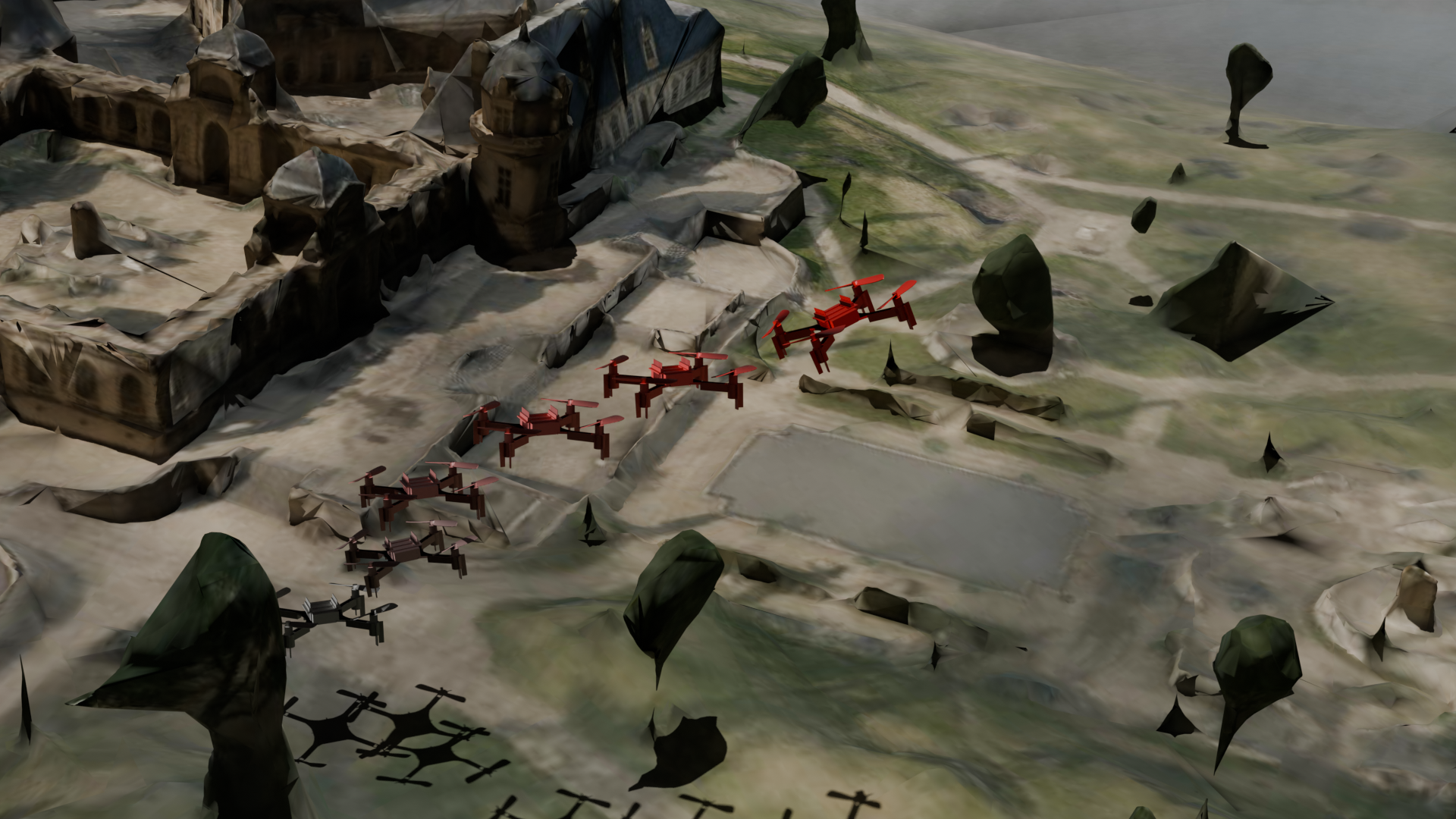}}
\hfill
\subfloat[\centering]{\includegraphics[width=8cm]{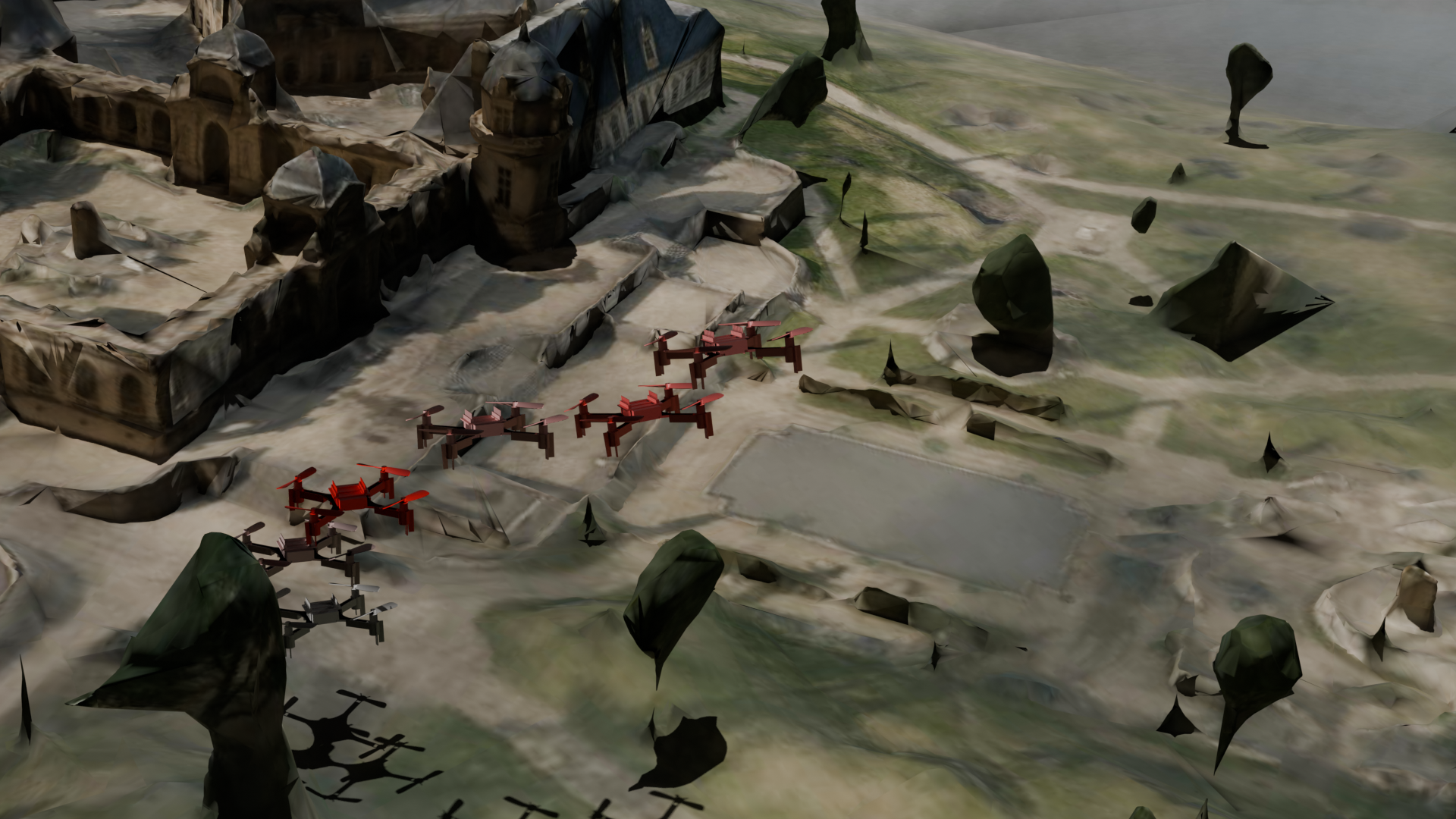}}

\end{adjustwidth}
\caption{Schematic diagram and visualization of controlled UAV motion with a given trajectory, where (\textbf{a}) compares a UAV's actual trajectory with a small-amplitude maneuver reference trajectory, (\textbf{b}) compares a large-amplitude maneuver with its reference trajectory, (\textbf{c}) visualizes the actual motion trajectory of (\textbf{a}), and (\textbf{d}) visualizes the actual motion trajectory of (\textbf{b}), with more vivid UAV positions indicating poses rendered at later time steps.
\label{fig:results_uav_model_control_sim_vis}}
\end{figure}

The performance is shown in Fig.~\ref{fig:results_uav_model_control_sim_vis}. where Fig.~\ref{fig:results_uav_model_control_sim_vis}(\textbf{a})-(\textbf{b}) compare the reference and ideal trajectories under small and large amplitude motions, respectively. These comparisons show that the UAVs can closely follow the desired paths with low tracking error.

Fig.~\ref{fig:results_uav_model_control_sim_vis}(\textbf{c})-(\textbf{d}) provide visualized renderings of the UAV trajectories corresponding to the two cases. The UAV positions and attitudes rendered at different time steps demonstrate the controller's ability to maintain stable and accurate flight. The results validate the dynamics model and control law design of the UAV module.

\begin{table}
\caption{Parameters for the UAV model.\label{table:UAV_model}}
\begin{adjustwidth}{-\extralength}{0cm} % Extend table width
\newcolumntype{Y}{>{\raggedright\arraybackslash}X}
\begin{tabularx}{\dimexpr\textwidth+\extralength\relax}{Y c c c}
\toprule
\textbf{Parameter} & \textbf{Symbol} & \textbf{Value} & \textbf{Unit} \\
\midrule
Rotational inertia of the UAV & \(\mathbf  J \) & \( \text{diag}\{1.40 \times 10^{-4}, 1.40 \times 10^{-4}, 2.17 \times 10^{-4}\} \) & \( kg\cdot m^2 \) \\
Mass of the UAV & \( m \) & \( 2.80 \times 10^{-1} \) & \( kg \) \\
Distance from rotor to UAV & \( l_i \) & \( [9.60 \times 10^{-2}, 9.60 \times 10^{-2}, 9.60 \times 10^{-2}, 9.60 \times 10^{-2}]' \) & \( m \) \\
Offset angle of rotors & \( \theta_i \) & \( [7.85 \times 10^{-1}, 2.36 \times 10^0, 3.93 \times 10^0, 5.50 \times 10^0]' \) & - \\
Lift coefficient of rotors & \( c_i \) & \( [2.88 \times 10^{-7}, 2.88 \times 10^{-7}, 2.88 \times 10^{-7}, 2.88 \times 10^{-7}]' \) & \( N\cdot rad^{-2}\cdot s^2 \) \\
Torque coefficient of rotors & \( k_i \) & \( [7.24 \times 10^{-9}, 7.24 \times 10^{-9}, 7.24 \times 10^{-9}, 7.24 \times 10^{-9}]' \) & \( N\cdot m \cdot rad^{-2}\cdot s^2 \) \\
\bottomrule
\end{tabularx}
\end{adjustwidth}
\end{table}
\begin{table}
\caption{Parameters for the PID controller.\label{table:PID_controller}}
\begin{adjustwidth}{-\extralength}{0cm} % Extend table width
\newcolumntype{Y}{>{\raggedright\arraybackslash}X}
\begin{tabularx}{\dimexpr\textwidth+\extralength\relax}{Y c c c}
\toprule
\textbf{Parameter} & \textbf{Symbol} & \textbf{Value} & \textbf{Unit} \\
\midrule
Proportional gain for position control & \( \mathbf k_p^p \) & \( [4.00 \times 10^{-1}, 4.00 \times 10^{-1}, 1.25 \times 10^0]' \) & \( N\cdot m^{-1} \) \\
Integral gain for position control & \(\mathbf k_i^p \) & \( [5.00 \times 10^{-2}, 2.50 \times 10^{-2}, 5.00 \times 10^{-2}]' \) & \( N\cdot m^{-1}\cdot s^{-1} \) \\
Derivative gain for position control & \(\mathbf k_d^p \) & \( [2.00 \times 10^{-1}, 1.20 \times 10^{-1}, 5.00 \times 10^{-1}]' \) & \( N\cdot m^{-1}\cdot s \) \\
Proportional gain for attitude control & \(\mathbf k_p^a \) & \( [7.00 \times 10^4, 7.00 \times 10^4, 6.00 \times 10^4]' \) & \( N\cdot m\cdot rad^{-1} \) \\
Integral gain for attitude control & \(\mathbf k_i^a \) & \( [0.00, 0.00, 5.00 \times 10^2]' \) & \( N\cdot m\cdot rad^{-1}\cdot s^{-1} \) \\
Derivative gain for attitude control & \(\mathbf k_d^a \) & \( [2.00 \times 10^4, 2.00 \times 10^4, 1.20 \times 10^4]' \) & \( N\cdot m\cdot rad^{-1}\cdot s \) \\
\bottomrule
\end{tabularx}
\end{adjustwidth}
\end{table}

\subsection{Validation of the Perception and Control Modules}\label{sec:perception_control}

To validate the perception module, the UAV state estimation was implemented according to Alg.~\ref{alg:drone_state_estimation_process}, using the camera configuration illustrated in Fig.~\ref{fig:camera_fixed}. The camera and IMU simulation parameters were set as shown in Table.~\ref{table:camera} and Table.~\ref{table:IMU_simulator}, respectively. UAVs followed the control scheme described in Fig.~\ref{fig:entire_controller_pd}.

\begin{table}
\caption{Parameters for the camera.\label{table:camera}}
\begin{adjustwidth}{-\extralength}{0cm} % Extend table width
\newcolumntype{Y}{>{\raggedright\arraybackslash}X}
\begin{tabularx}{\dimexpr\textwidth+\extralength\relax}{Y c c c}
\toprule
\textbf{Parameter} & \textbf{Symbol} & \textbf{Value} & \textbf{Unit} \\
\midrule
Camera center position relative to UAV & \(\mathbf {d}_c \) & \( [3.00 \times 10^{-2}, 0, -3.00 \times 10^{-2}]' \) & \( m \) \\
Camera rotational inertia & \(\mathbf {J}_c \) & \( \text{diag}\{2.17 \times 10^{-6}, 4.83 \times 10^{-6}, 5.67 \times 10^{-6}\} \) & \( kg \cdot m^2 \) \\
Camera mass & \( m_c \) & \( 2.00 \times 10^{-2} \) & \( kg \) \\
Camera focal length & \( f \) & \( 50 \) & \( mm \) \\
Camera sensor size & \( [s_x, s_y] \) & \( [36, 24] \) & \( mm \) \\
Image resolution & \( [W, H] \) & \( [1920, 1080] \) &  -  \\
Image principal point coordinates & \( [c_x, c_y] \) & \( [960, 540] \) &  -  \\
Image distortion coefficients & \( [k_1, k_2, p_1, p_2] \) & \( [0, 0, 0, 0] \) &  -  \\
\bottomrule
\end{tabularx}
\end{adjustwidth}
\end{table}
\begin{table}
\caption{Parameters for the IMU simulator.\label{table:IMU_simulator}}
\newcolumntype{Y}{>{\raggedright\arraybackslash}X}
\begin{tabularx}{\textwidth}{Y c c c}
\toprule
\textbf{Parameter} & \textbf{Symbol} & \textbf{Value} & \textbf{Unit} \\
\midrule
Accelerometer noise density & \( \sigma_a \) & \( 2.00 \times 10^{-3} \) & \( m \cdot s^{-2} \cdot Hz^{-0.5} \) \\
Gyroscope noise density & \( \sigma_g \) & \( 1.00 \times 10^{-4} \) & \( rad \cdot s^{-1} \cdot Hz^{-0.5} \) \\
Accelerometer bias drift & \( b_a \) & \( 1.00 \times 10^{-4} \) & \( m \cdot s^{-3} \) \\
Gyroscope bias drift & \( b_g \) & \( 1.00 \times 10^{-6} \) & \( rad \cdot s^{-2} \) \\
\bottomrule
\end{tabularx}
\end{table}

\begin{figure}
\centering
\begin{adjustwidth}{-\extralength}{0cm}
\subfloat[\centering]{\includegraphics[width=4cm]{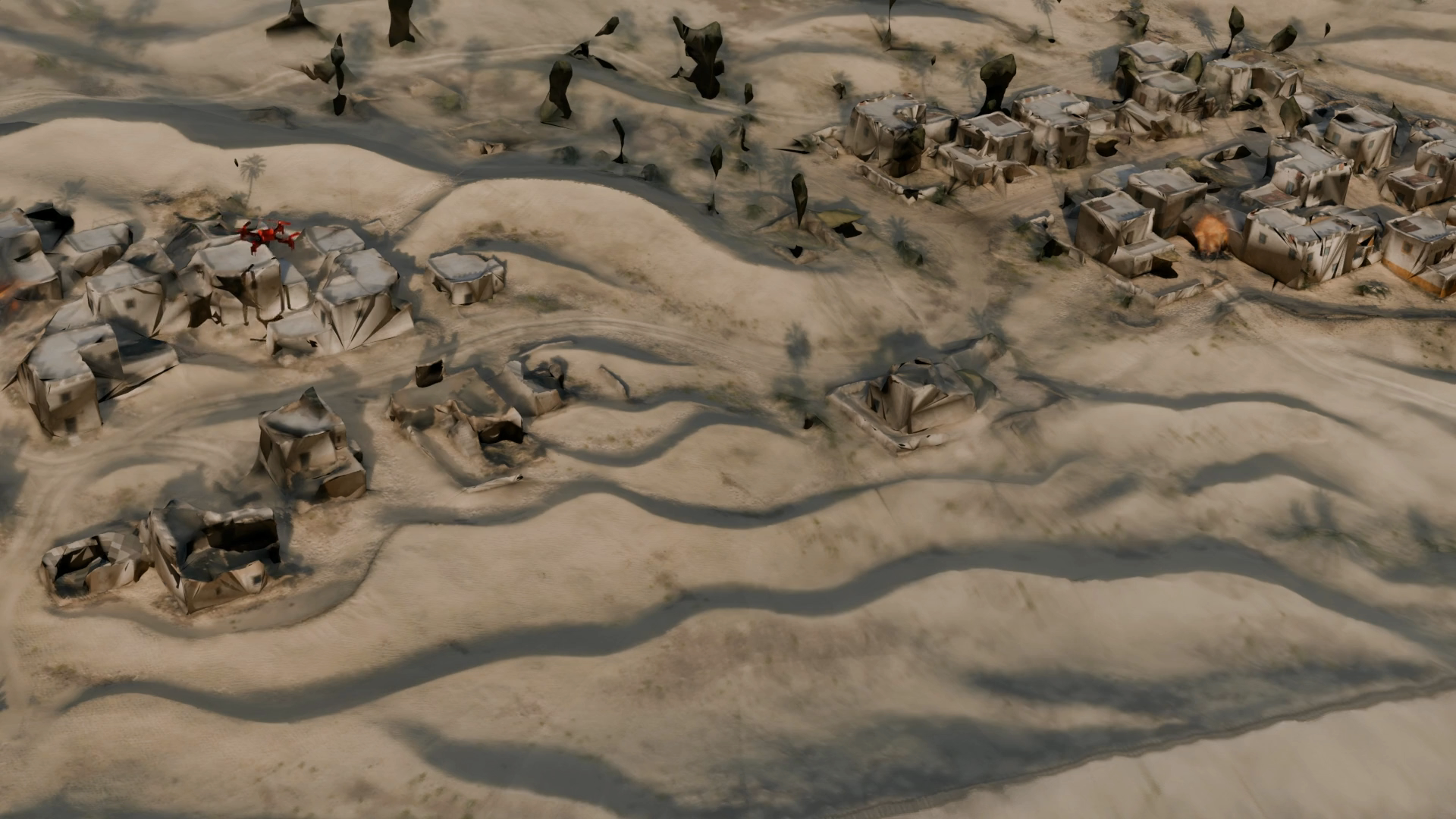}}
\hfill
\subfloat[\centering]{\includegraphics[width=4cm]{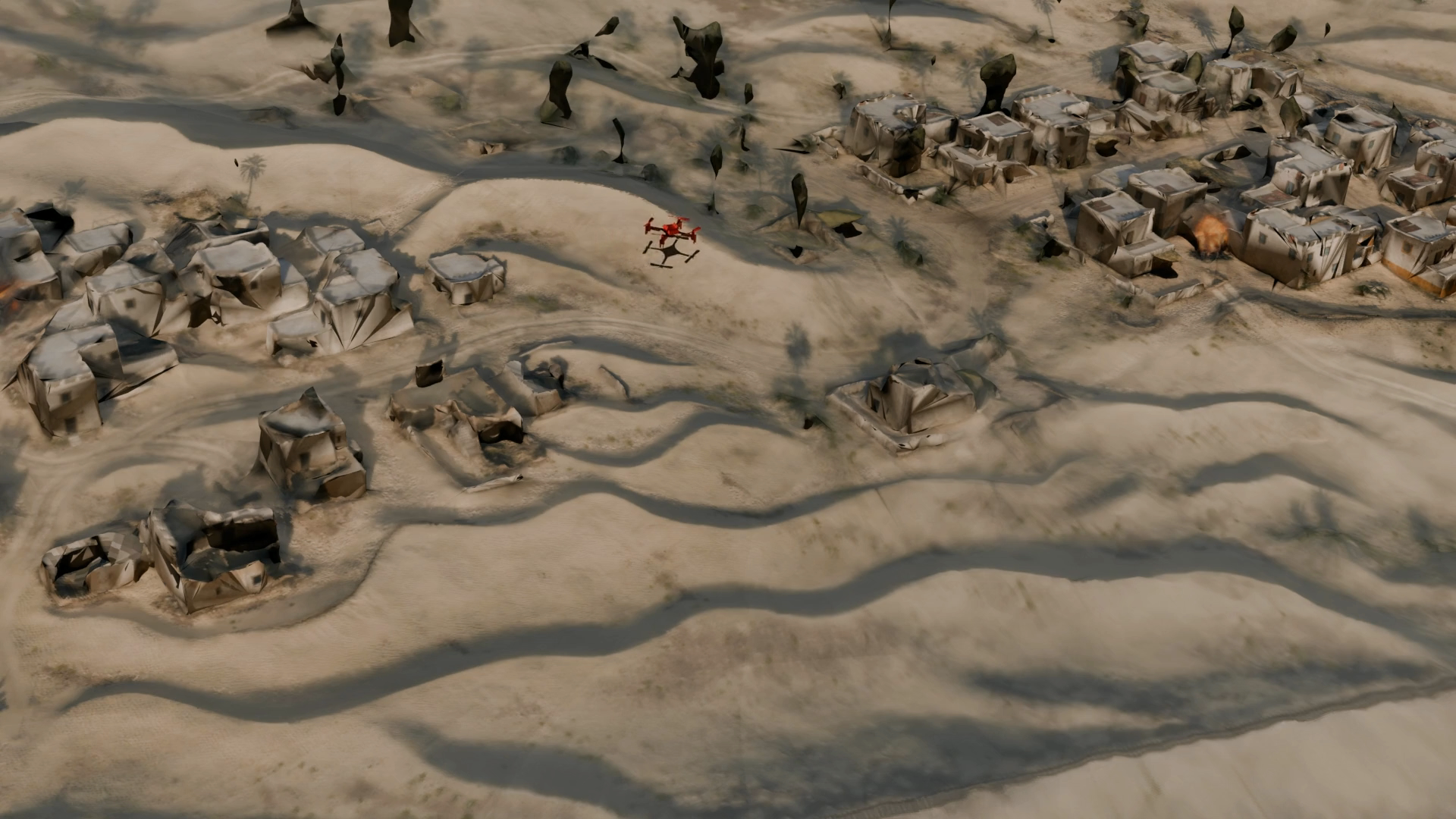}}
\hfill
\subfloat[\centering]{\includegraphics[width=4cm]{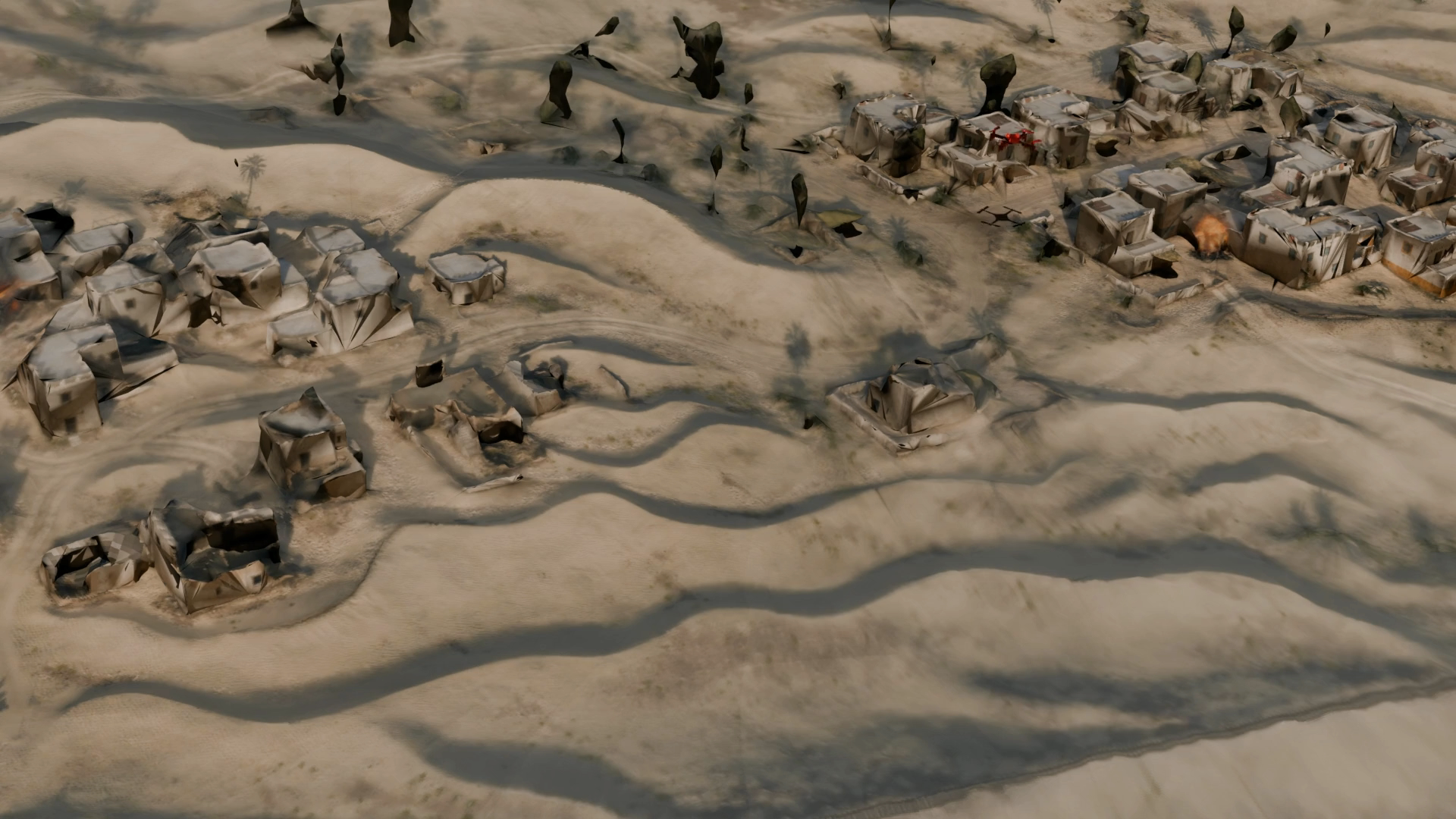}}
\hfill
\subfloat[\centering]{\includegraphics[width=4cm]{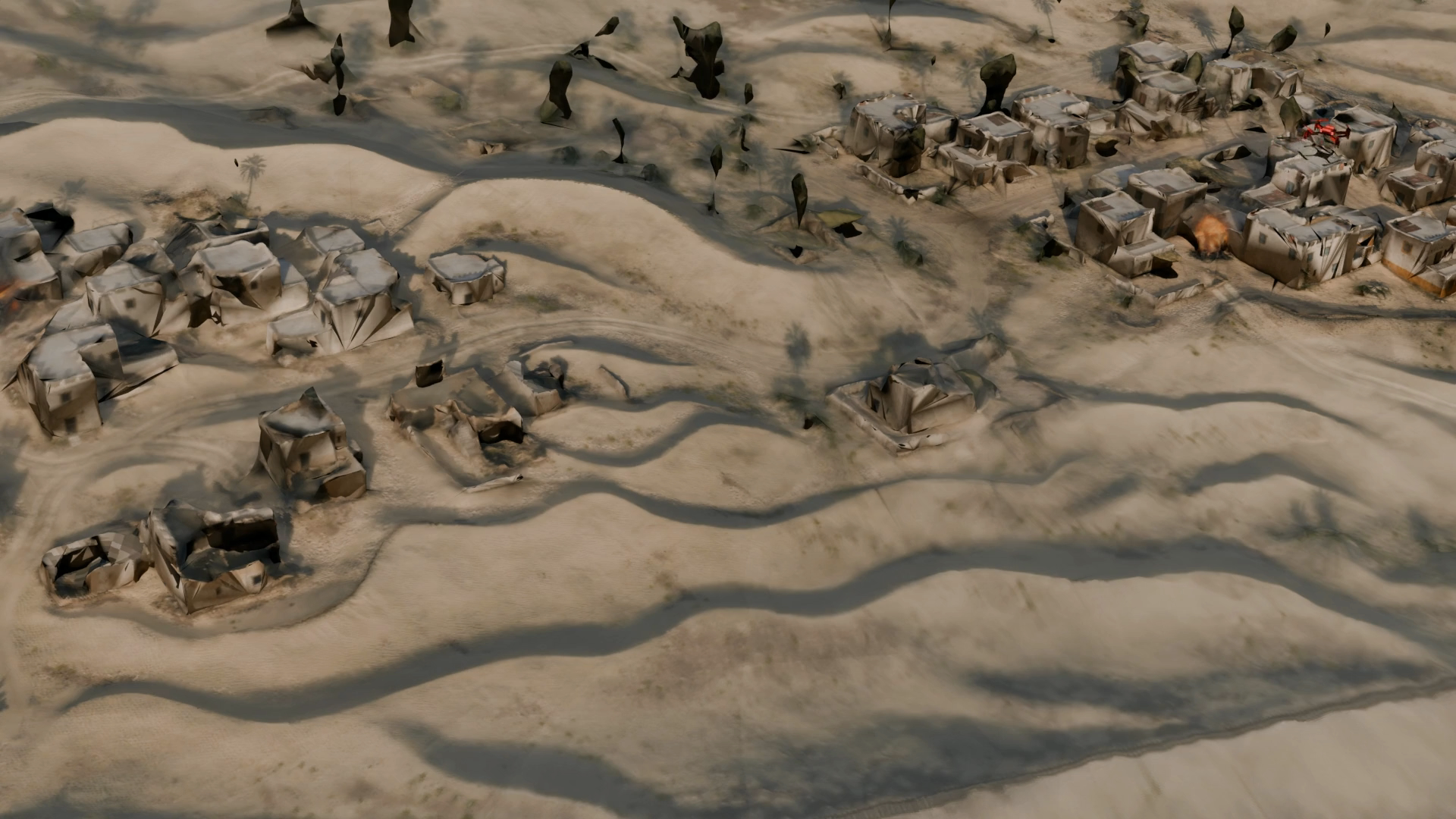}}\\
\subfloat[\centering]{\includegraphics[width=4cm]{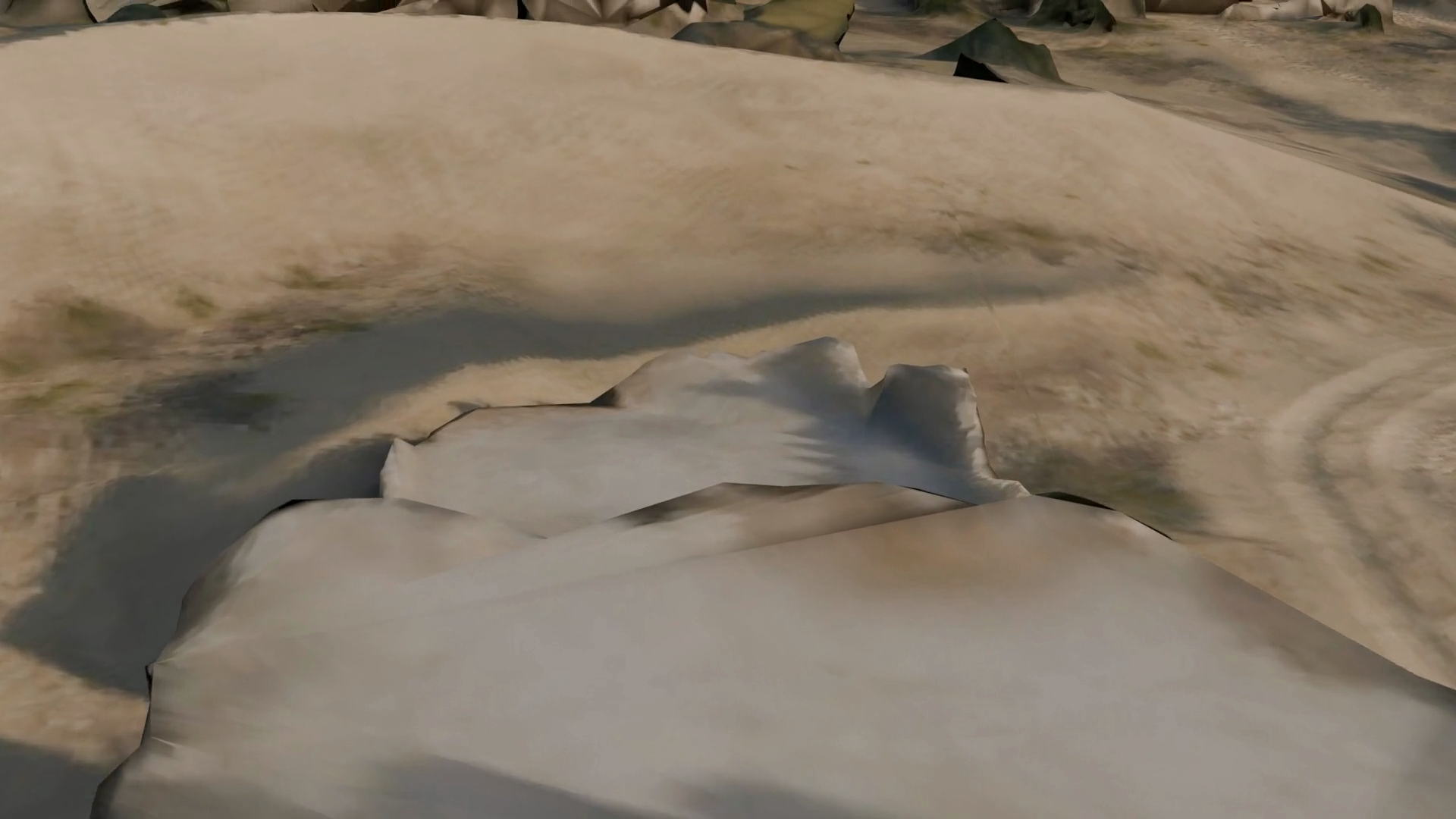}}
\hfill
\subfloat[\centering]{\includegraphics[width=4cm]{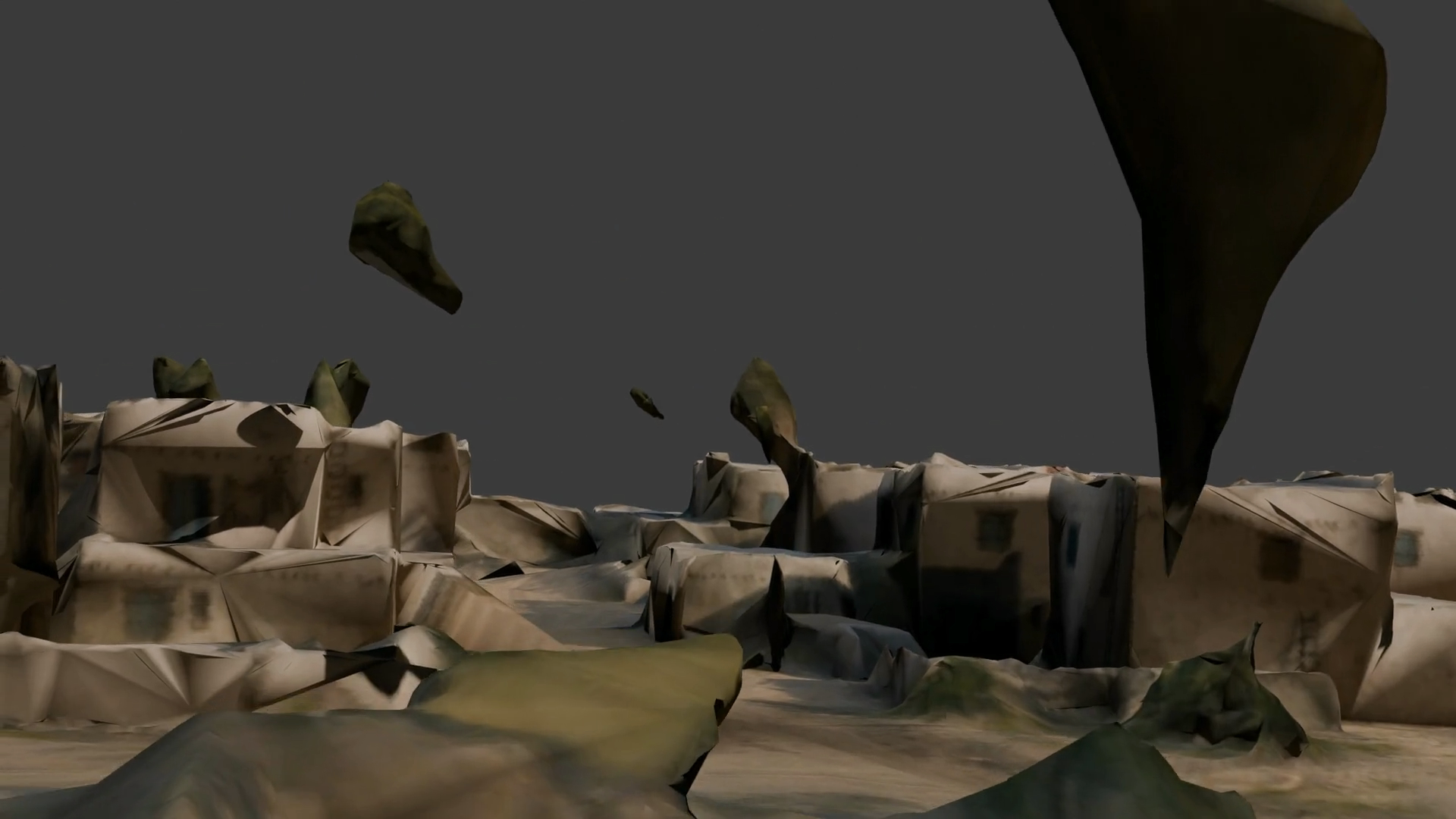}}
\hfill
\subfloat[\centering]{\includegraphics[width=4cm]{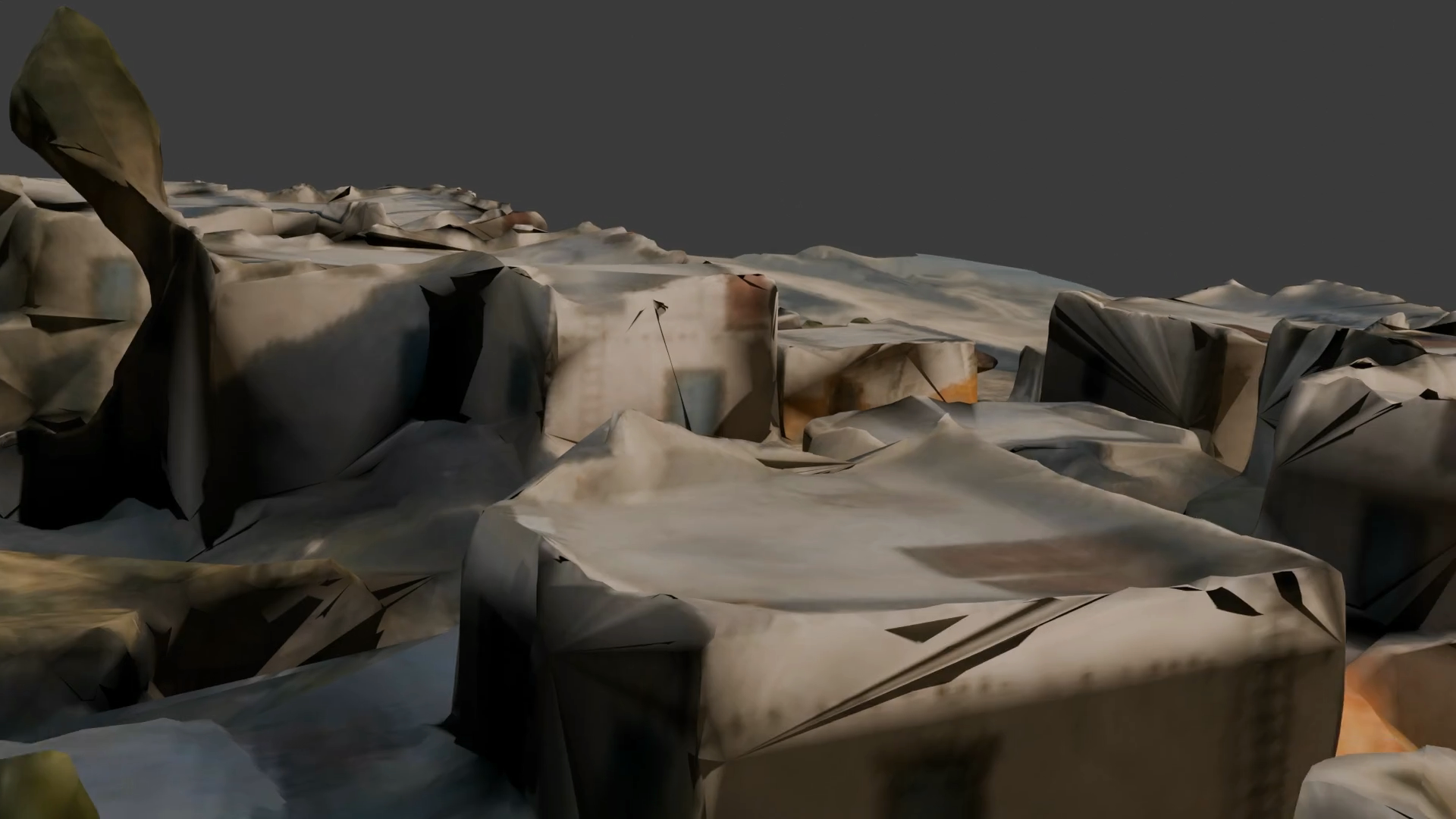}}
\hfill
\subfloat[\centering]{\includegraphics[width=4cm]{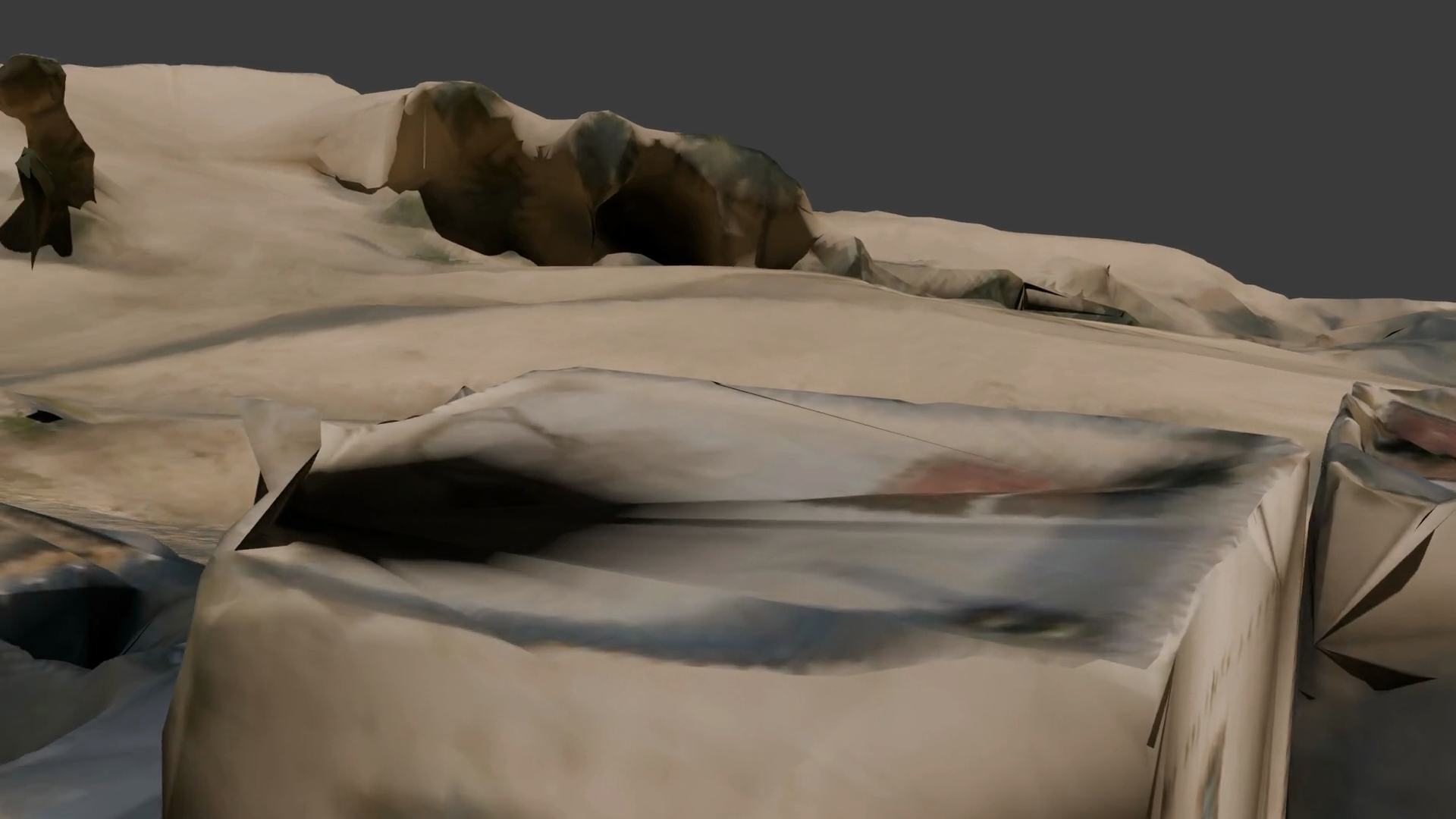}}

\end{adjustwidth}
\caption{Visualization of UAV self-estimated trajectory, where (\textbf{e})-(\textbf{h}) show sequential visual information captured by the camera setup in Fig.~\ref{fig:camera_fixed}, while (\textbf{a})-(\textbf{d}) show the corresponding UAV positions and orientations from another viewpoint.
\label{fig:results_estimation_vis}}
\end{figure}

\begin{figure}
\centering
% \begin{adjustwidth}{-\extralength}{0cm}
\subfloat[\centering]{\includegraphics[width=8cm]{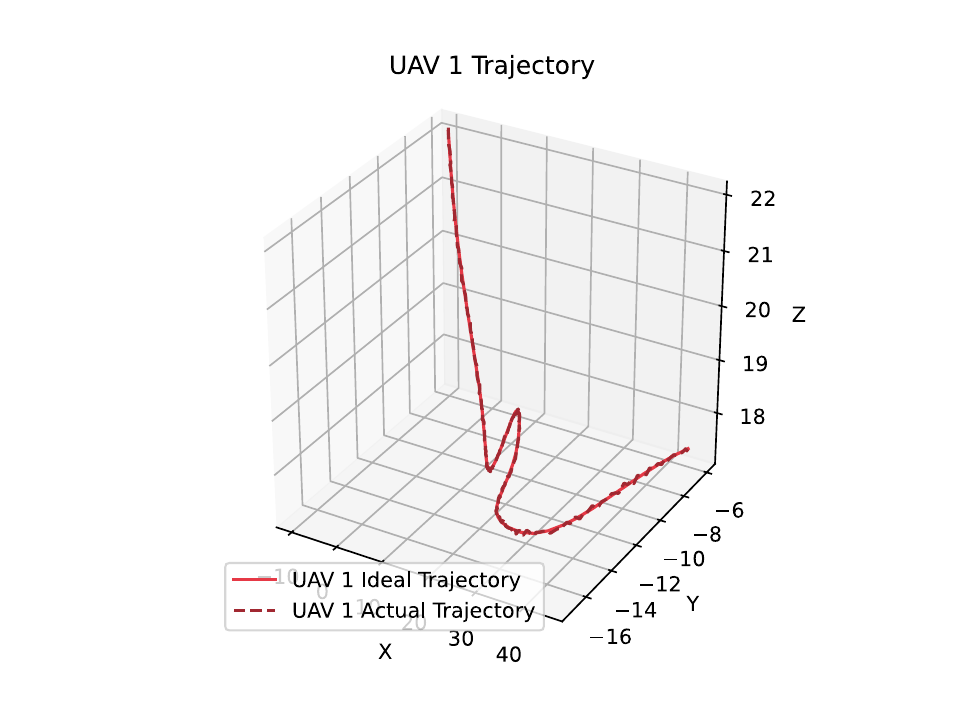}}
\hfill
\subfloat[\centering]{\includegraphics[width=8cm]{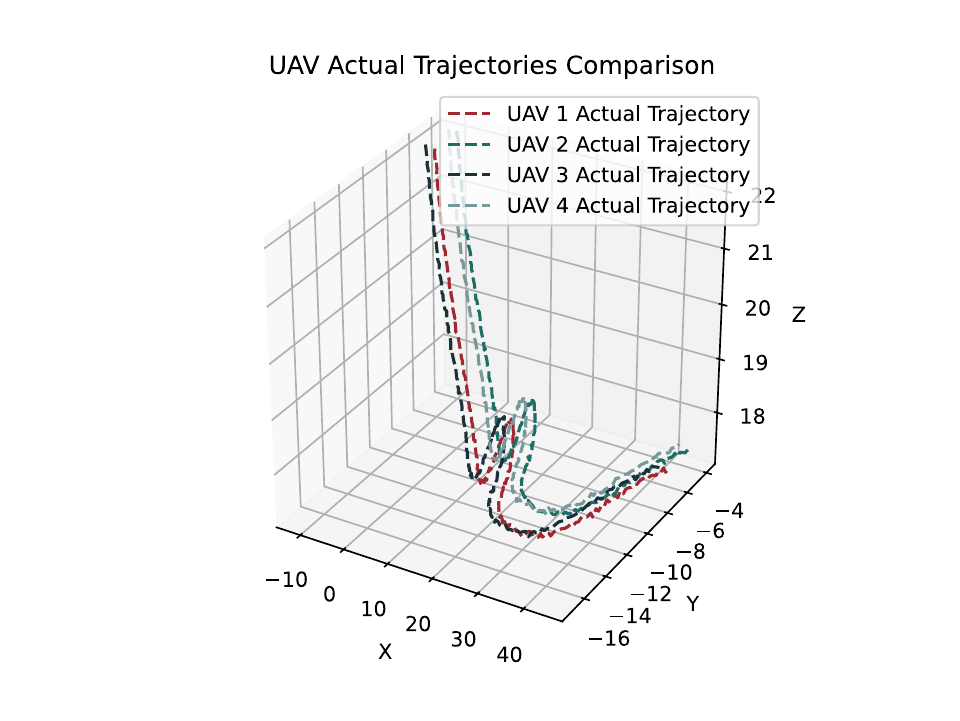}}

% \end{adjustwidth}
\caption{Comparison of trajectories, where (\textbf{a}) compares estimated trajectory of UAV 1 with its reference trajectory, utilizing vision like Fig.~\ref{fig:results_estimation_vis}(\textbf{e})-(\textbf{h}) and simulated IMU signals. (\textbf{b}) shows all the estimated trajectories of UAVs.
\label{fig:results_estimation_sim}}
\end{figure}

Fig.~\ref{fig:results_estimation_vis} shows the visual perception results during UAV flight. Specifically, Fig.~\ref{fig:results_estimation_vis}(\textbf{a})–(\textbf{d}) present the UAV’s estimated position and attitude from a third-person view, while Fig.~\ref{fig:results_estimation_vis}(\textbf{e})–(\textbf{h}) display the sequential images captured by the onboard camera. These image frames correspond to key moments during the UAV’s motion and demonstrate that the camera system can reliably capture environmental information and contribute to positon and attitude estimation. The combination of third-person images and camera images confirms the proper integration of vision and motion in the state estimation pipeline.

Quantitative comparison of estimated versus ideal UAV trajectories is shown in Fig.~\ref{fig:results_estimation_sim}. Fig.~\ref{fig:results_estimation_sim}(\textbf{a}) displays the estimated trajectory of UAV 1 compared to its reference trajectory. Fig.~\ref{fig:results_estimation_sim}(\textbf{b}) shows the estimated trajectories of all UAVs. The close alignment between estimated and ideal paths indicates that the proposed vision-inertial state estimation approach can effectively support real-time UAV estimation. This validates the robustness of the perception pipeline under motion and sensor noise.

\begin{figure}
\centering
\begin{adjustwidth}{-\extralength}{0cm}
\subfloat[\centering]{\includegraphics[width=5.5cm]{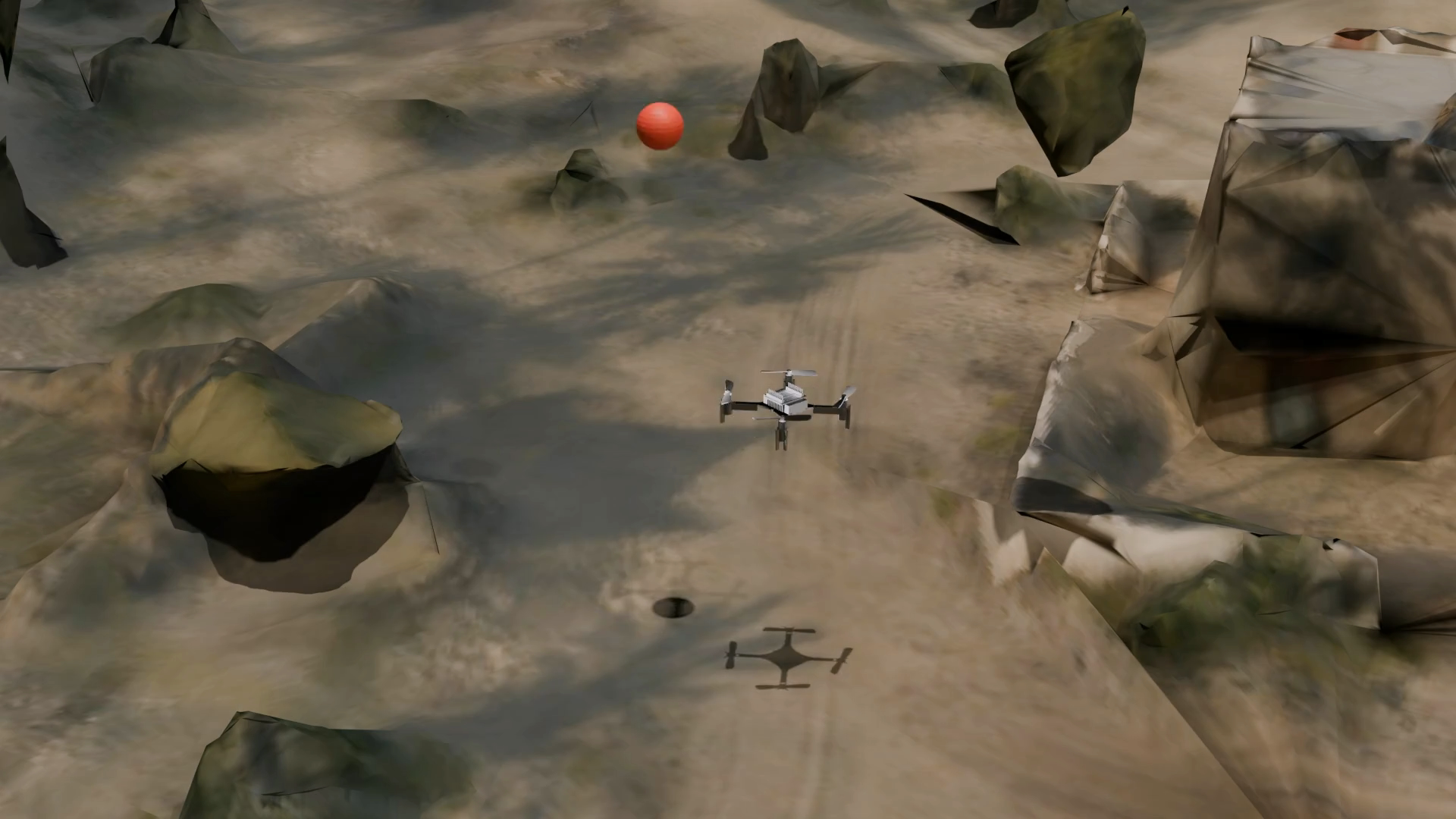}}
\hfill
\subfloat[\centering]{\includegraphics[width=5.5cm]{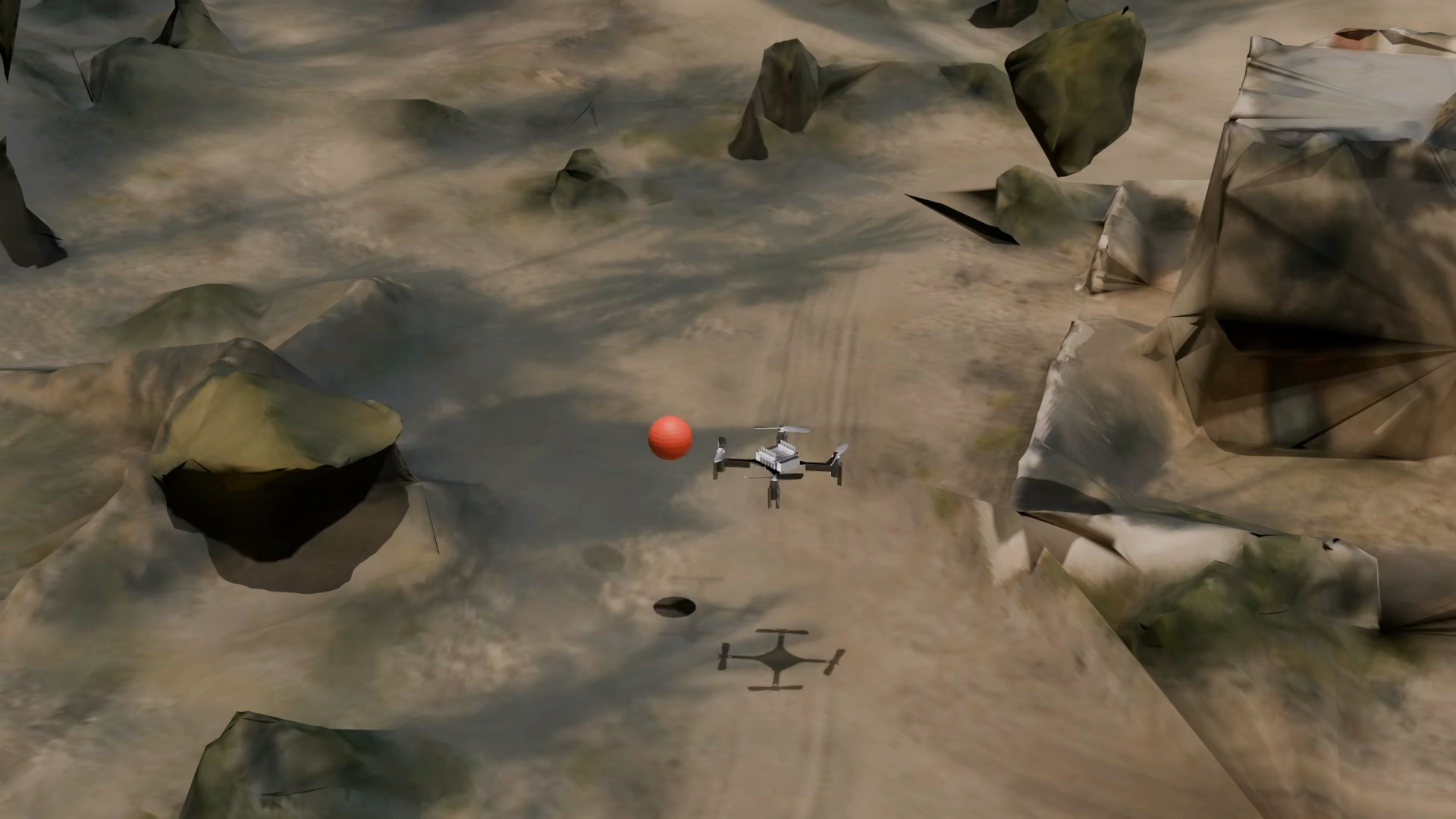}}
\hfill
\subfloat[\centering]{\includegraphics[width=5.5cm]{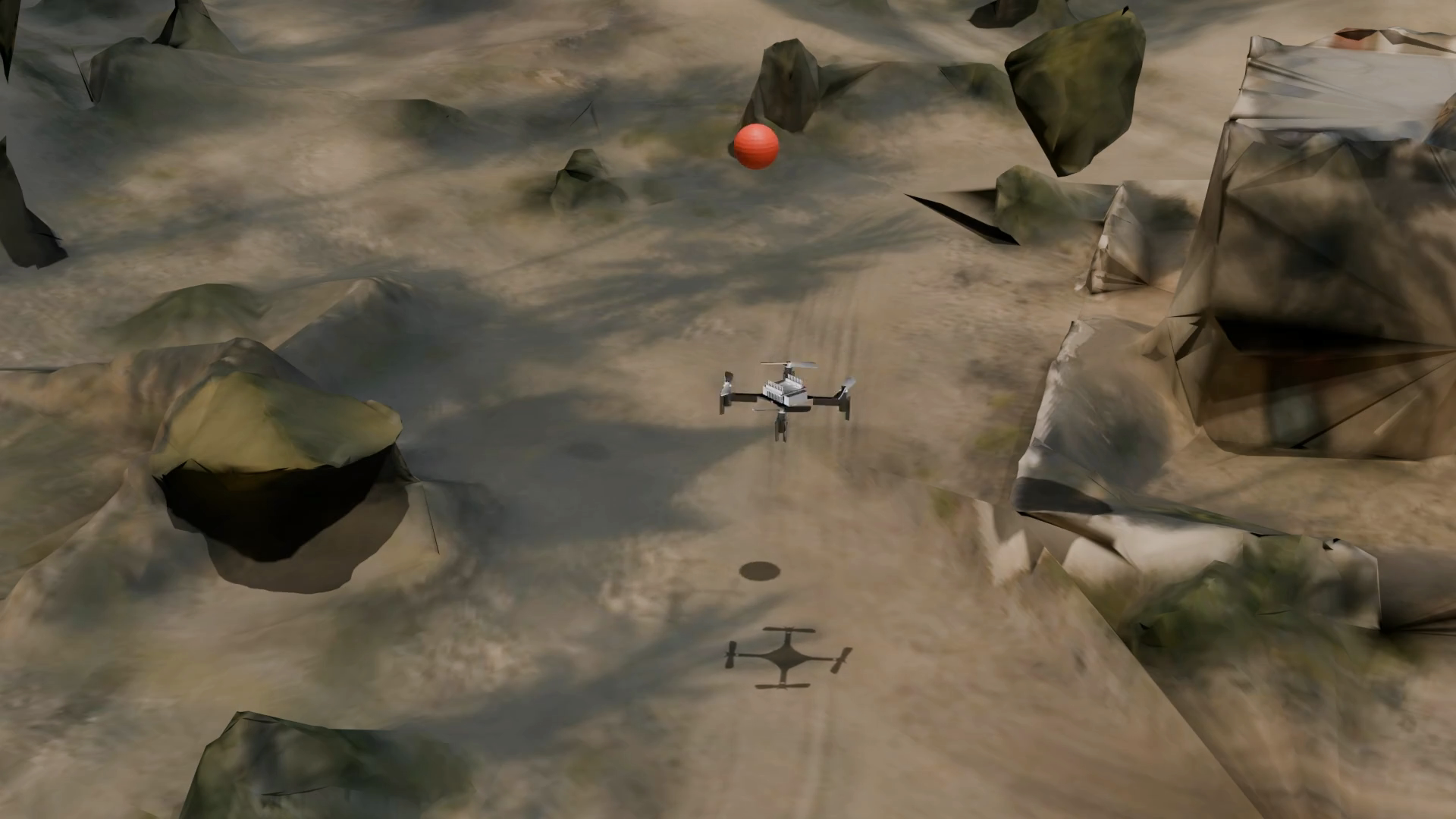}}\\
\subfloat[\centering]{\includegraphics[width=5.5cm]{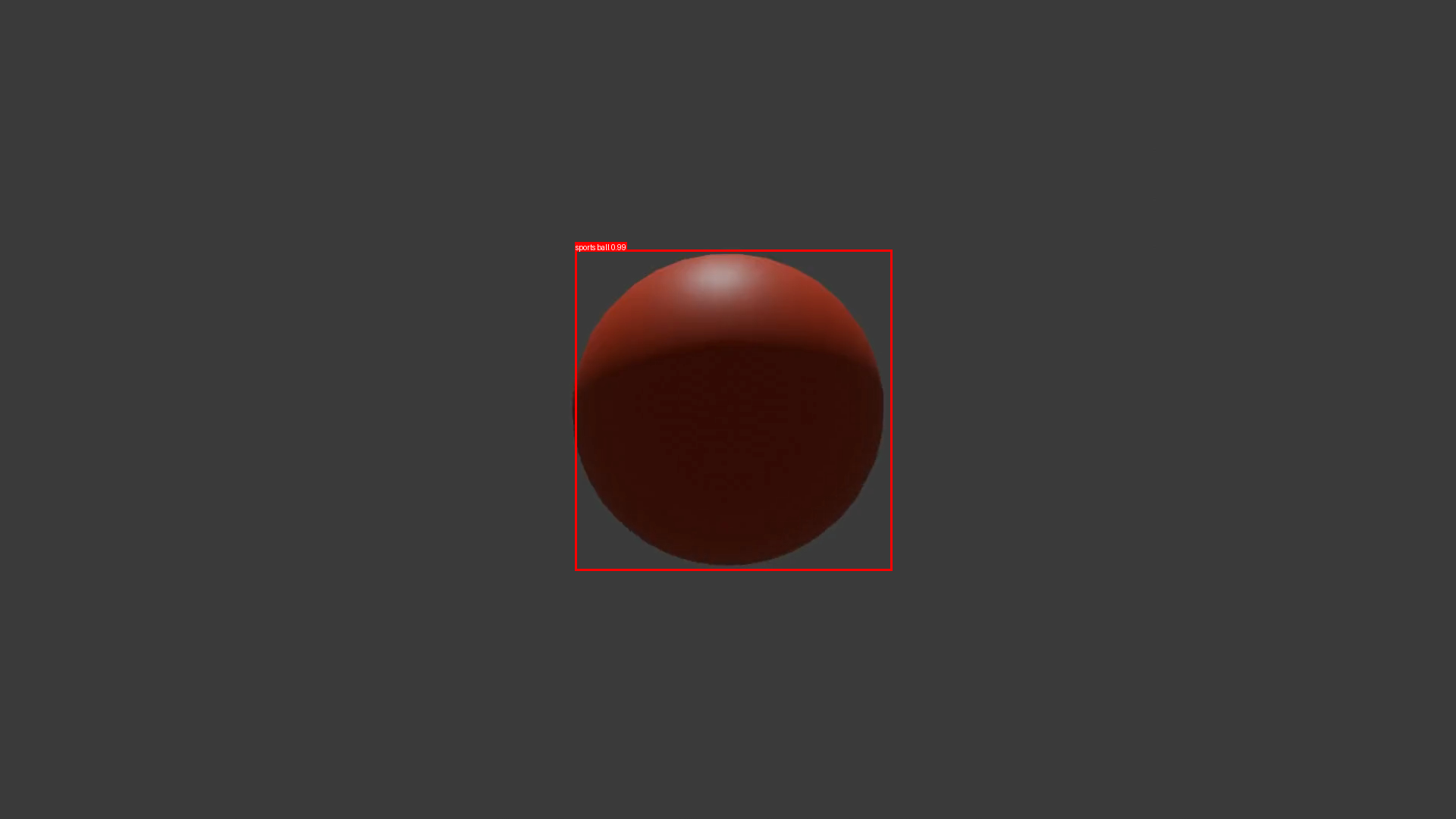}}
\hfill
\subfloat[\centering]{\includegraphics[width=5.5cm]{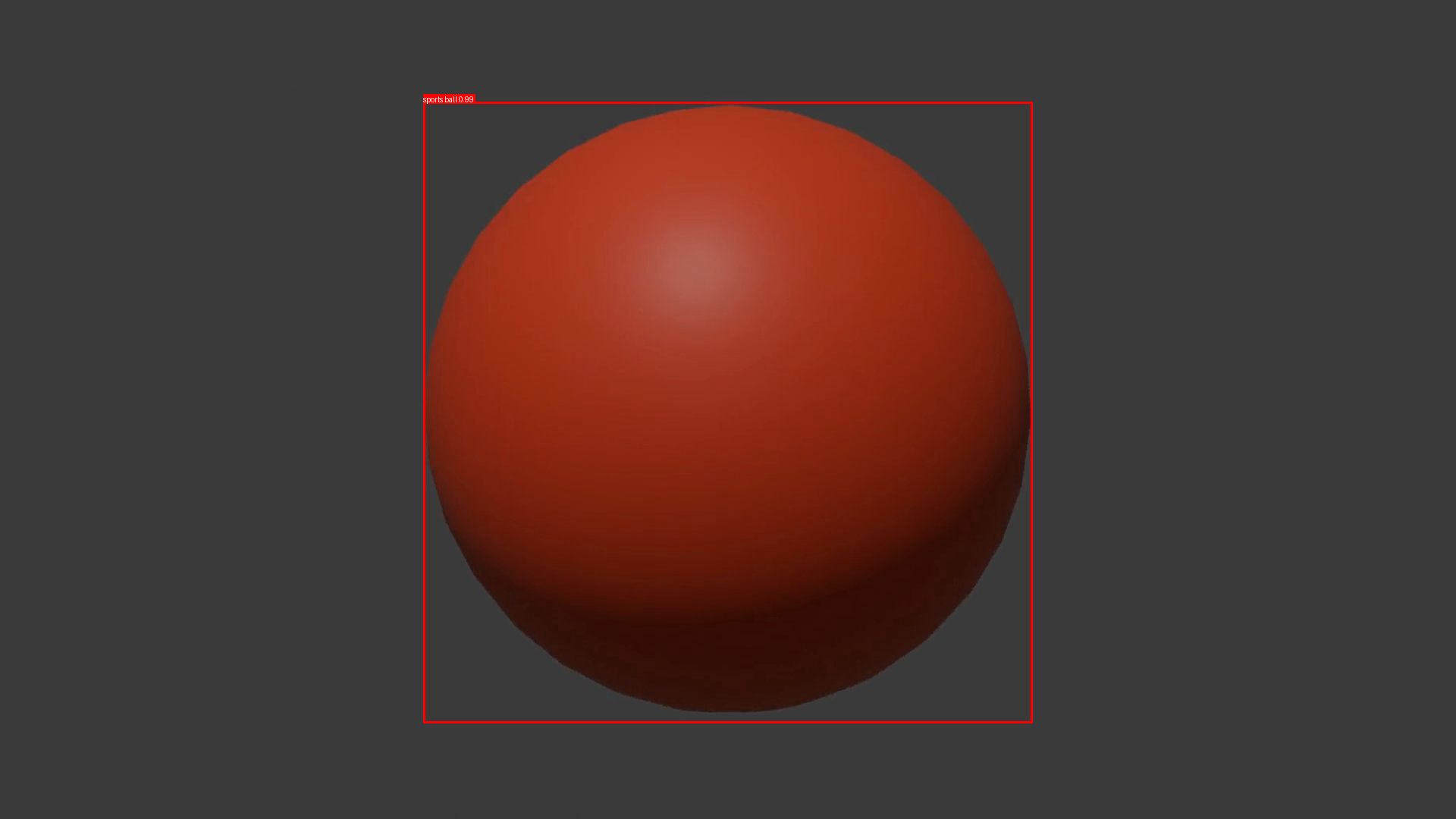}}
\hfill
\subfloat[\centering]{\includegraphics[width=5.5cm]{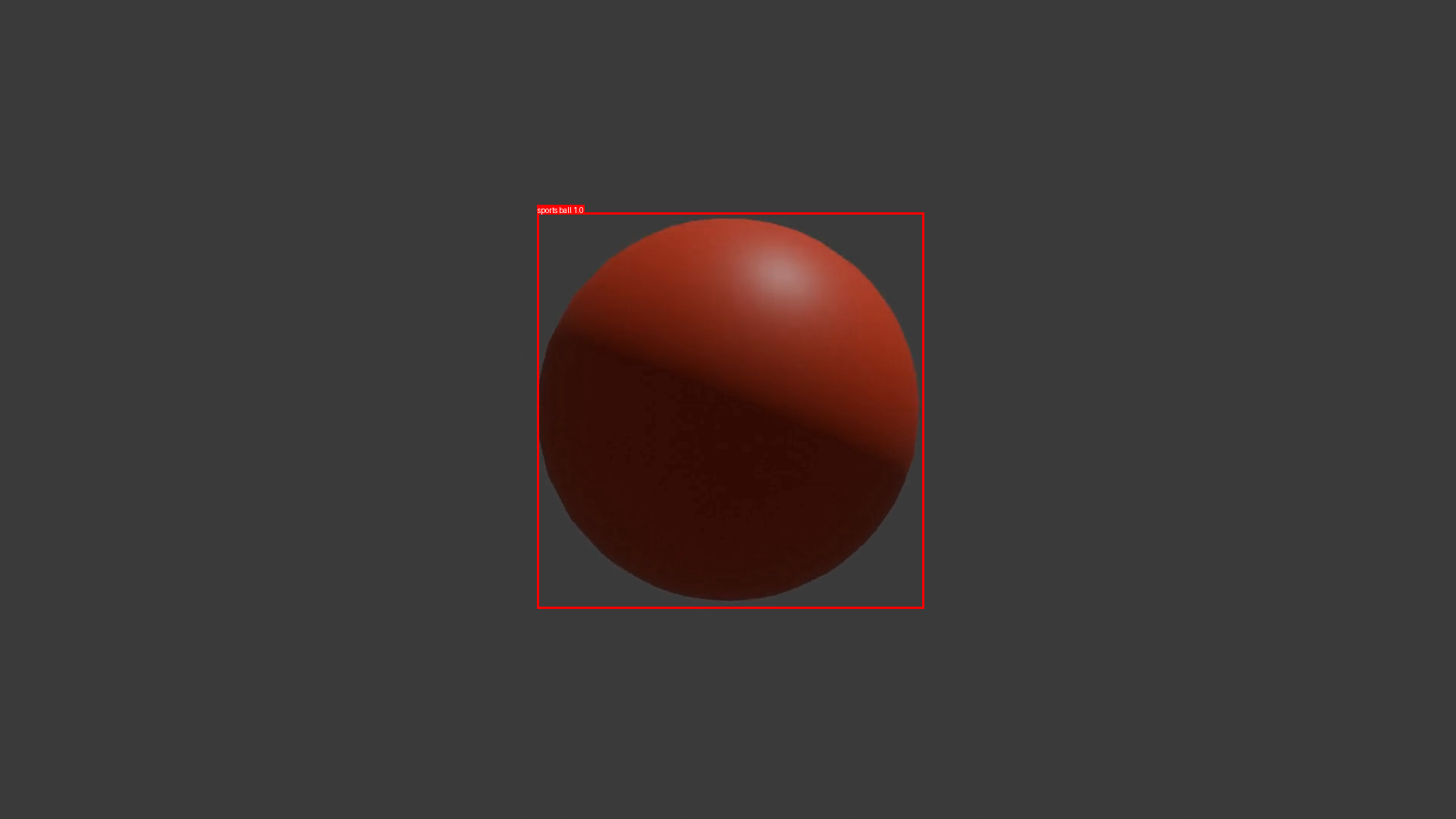}}

\end{adjustwidth}
\caption{Visualization of tracking and detection results from UAV 1's gimbal camera, where (\textbf{a})-(\textbf{c}) illustrate the location of UAV 1 and target, and (\textbf{d})-(\textbf{f}) present the corresponding detection results.
\label{fig:results_track_vis}}
\end{figure}

\begin{figure}
    \centering
        \subfloat[\centering]{\includegraphics[width=6cm]{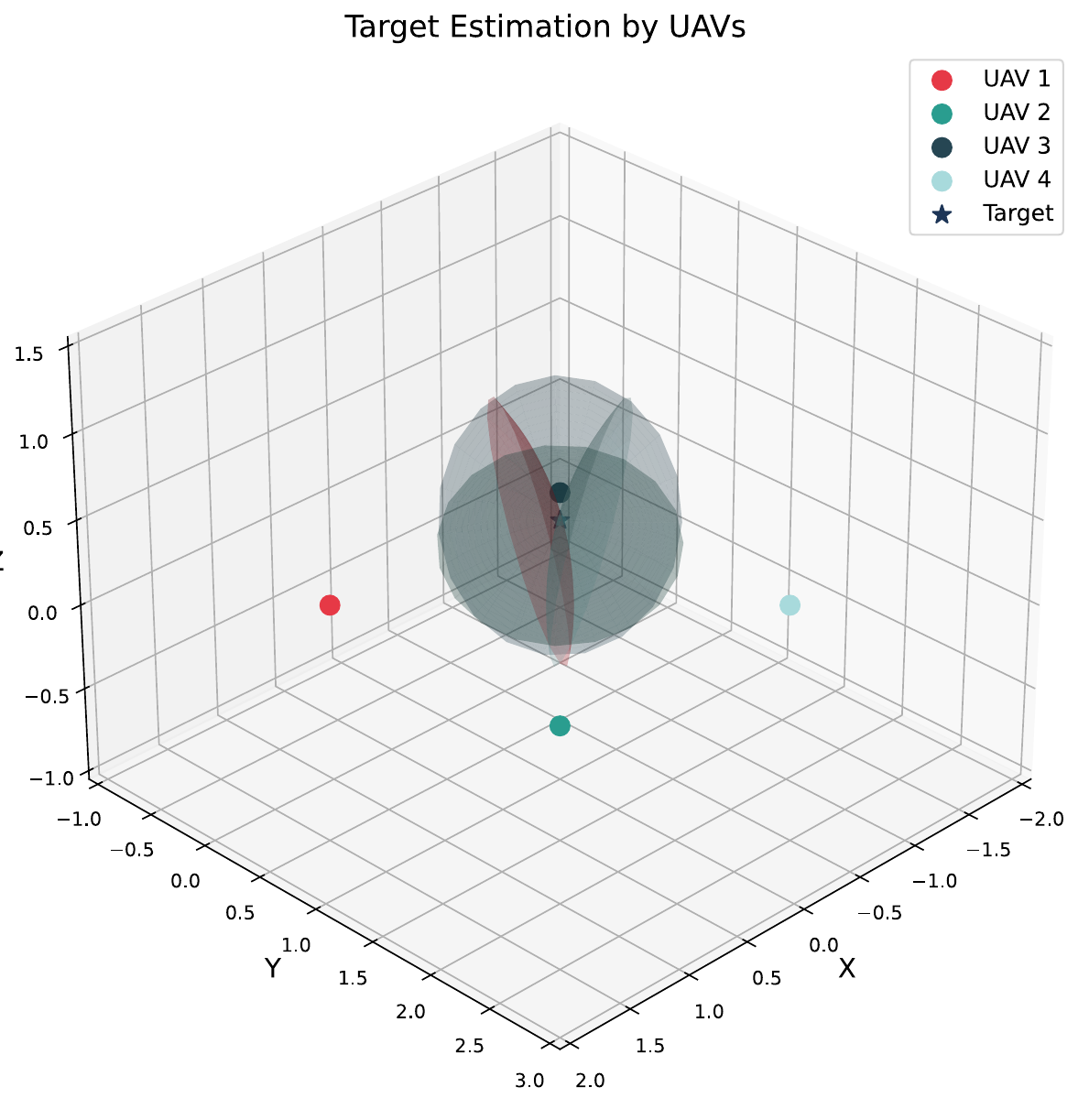}}
        \hfill
        \subfloat[\centering]{\includegraphics[width=6cm]{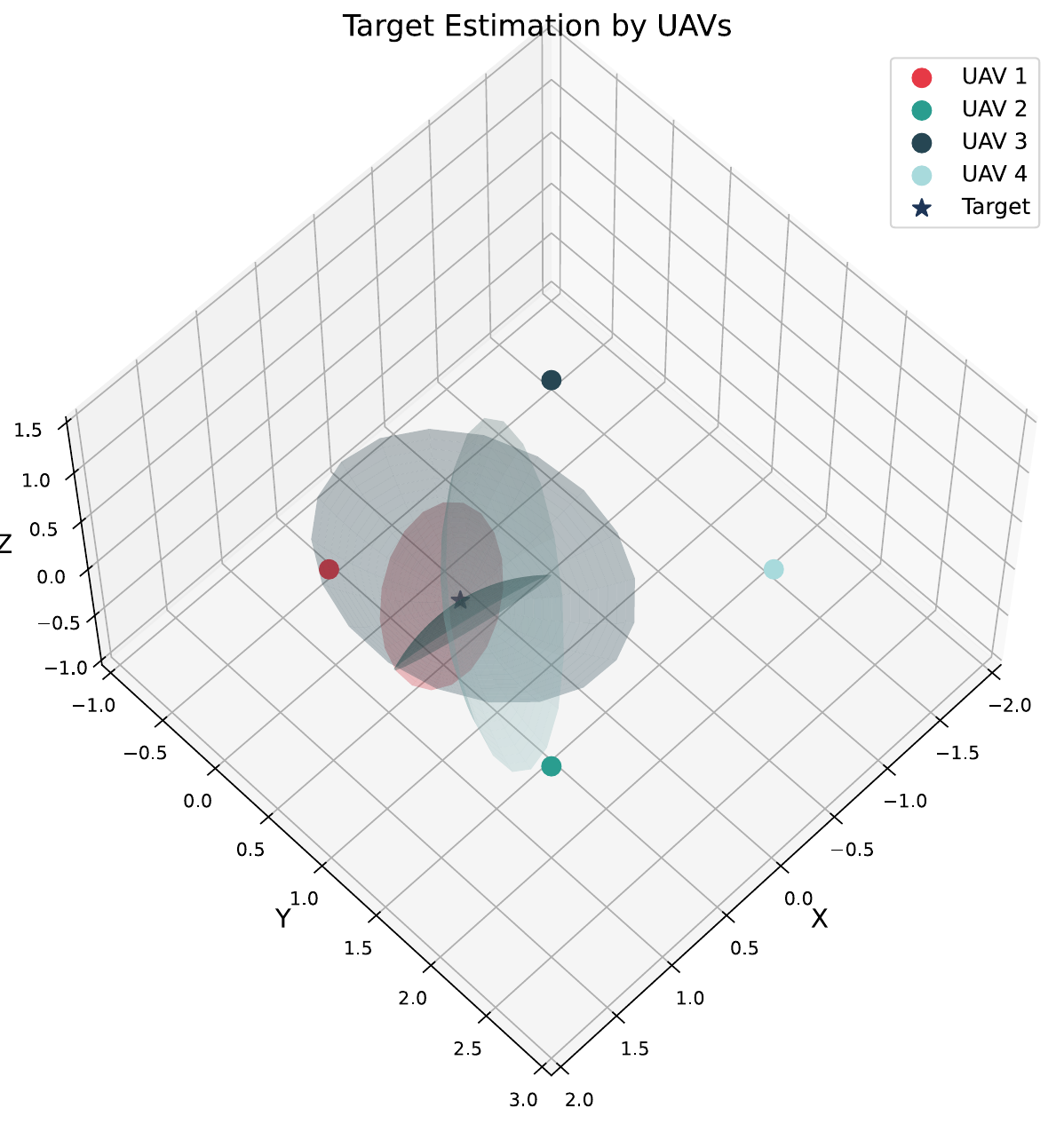}}\\
                
    \caption{Schematic diagram of UAVs-based non-cooperative target position estimation, where (\textbf{a}) and (\textbf{b}) show estimated target positions at different time steps based on UAVs' observations.
    \label{fig:results_track_sim}}
\end{figure}

The non-cooperative target tracking module was implemented according to Alg.~\ref{alg:target_tracking_process} using the camera configuration in Fig.~\ref{fig:camera_universal}. The tracking and detection process for UAV 1 is illustrated in Fig.~\ref{fig:results_track_vis}, where (\textbf{a})–(\textbf{c}) show the relative positions between UAV and target, and (\textbf{d})–(\textbf{f}) display the detection results from camera. These frames demonstrate the onboard tracking module’s ability to localize the target in varying spatial configurations and perspectives.

Based on the detection results of each UAV shown in Fig.~\ref{fig:results_track_vis}, the estimated distances between the non-cooperative target and each UAV can be determined. Using these distances along with the estimated positions of the UAVs, the target's location is subsequently estimated. The fusion-based position estimation of the target from multi-UAV observations is shown in Fig.~\ref{fig:results_track_sim}, where (\textbf{a}) and (\textbf{b}) display the estimated target position at different time steps, where the overlapping regions of the cones---colored similarly to their corresponding UAVs---align with the target's ideal position. Highlighting the benefit of collaborative perception across UAVs. These outcomes validate the feasibility of vision-based target estimation in multi-UAV scenarios. The results demonstrate a high degree of consistency, validating the accuracy of the estimation method.

\subsection{Implementation of Multi-UAV-Tethered Net Capture of Non-Cooperative Targets}\label{sec:multi_uav_capture}

To evaluate the end-to-end performance of the proposed multi-UAV-tethered netted system, we accomplish it in two representative scenarios: one involving a non-propelled target, and another with an actively maneuvering target governed by unknown internal propulsion. In both cases, estimated UAV and target states from the perception module serve as inputs to the learning-based control framework according to Alg.~\ref{alg:controller_process_mappo} with the network architecture defined in Fig.~\ref{fig:mappo_LNN}. The reward parameters in Fig.~\ref{fig:ippo} were set as per Table.~\ref{table:rewards}.

\begin{table}
\caption{Parameters for the rewards.\label{table:rewards}}
\centering
\newcolumntype{Y}{>{\raggedright\arraybackslash}X}
\begin{tabularx}{\textwidth}{Y c c c}
\toprule
\textbf{Parameter} & \textbf{Symbol} & \textbf{Value} & \textbf{Unit} \\
\midrule
Reward for distance between UAV and target & \( r_{\text{distance}} \) & \( 1.2 \) &  -  \\
Reward for alignment with target motion & \( r_{\text{alignment}} \) & \( 0.6 \) &  -  \\
Reward for UAV spin control & \( r_{\text{spin}} \) & \( 0.8 \) &  -  \\
Reward for low energy consumption & \( r_{\text{effort}} \) & \( 0.1 \) &  -  \\
Reward for UAV swing relative to target & \( r_{\text{swing}} \) & \( 0.8 \) &  -  \\
Reward for maintaining safe distance between UAVs & \( r_{\text{safe}} \) & \( 0.5 \) &  -  \\
Reward for collision between rope and target & \( r_{\text{collision}} \) & \( 2.8 \) &  -  \\
Reward for stability after UAV collides with target & \( r_{\text{stability}} \) & \( 4.5 \) &  -  \\
\bottomrule
\end{tabularx}
\end{table}

\begin{figure}
\centering
\subfloat[\centering]{\includegraphics[width=6cm]{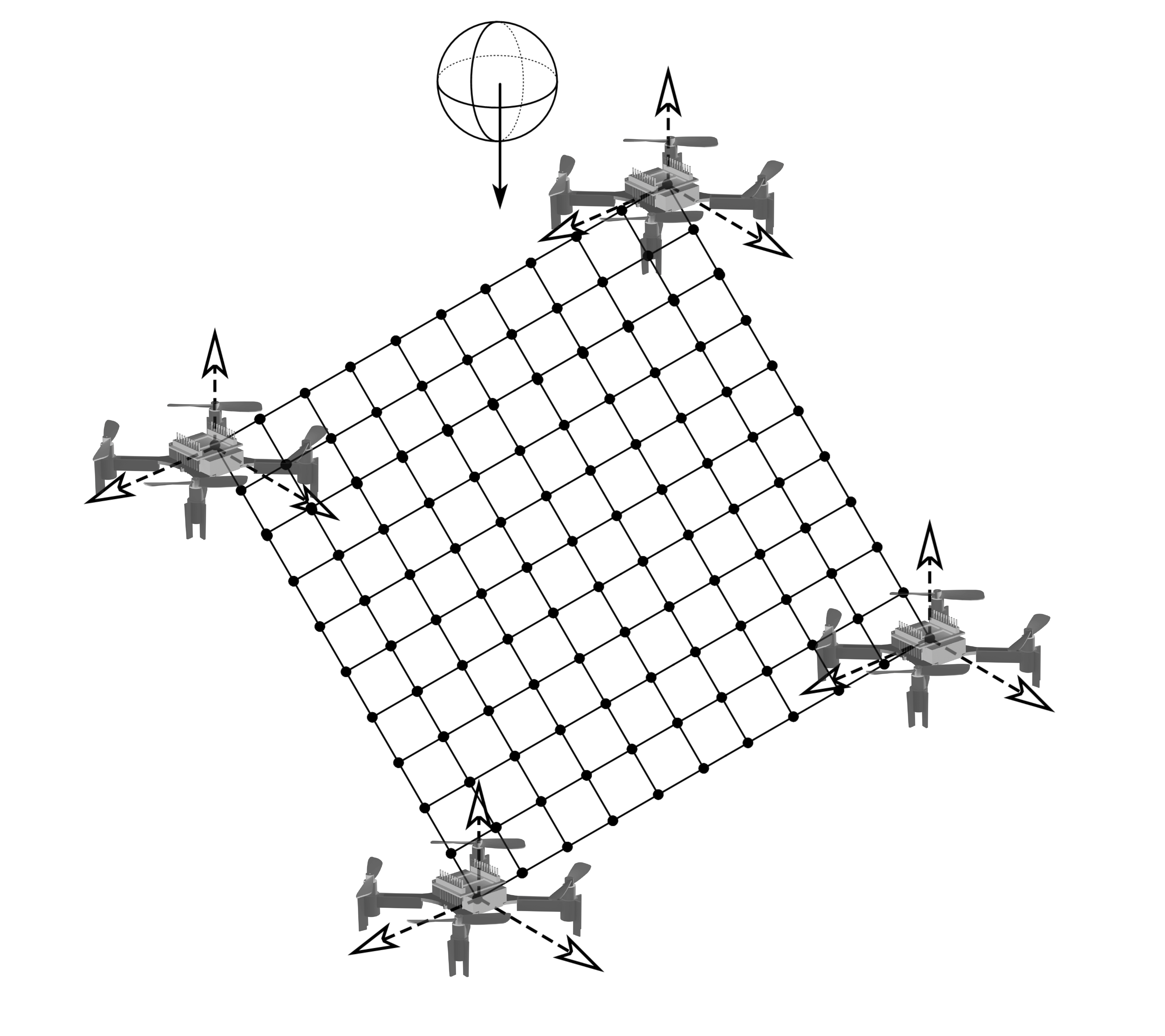}}
\hfill
\subfloat[\centering]{\includegraphics[width=6cm]{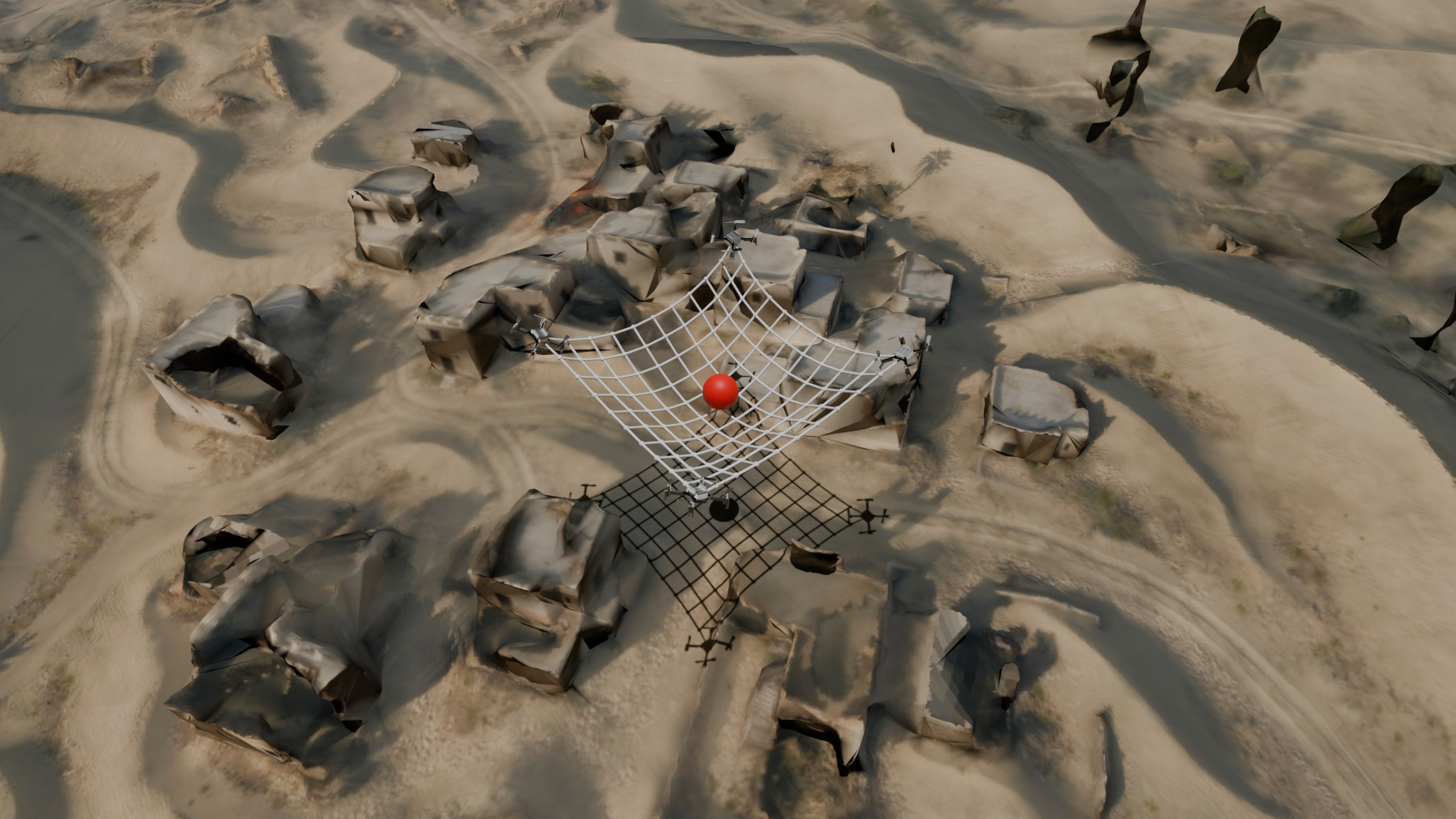}}\\
\subfloat[\centering]{\includegraphics[width=6cm]{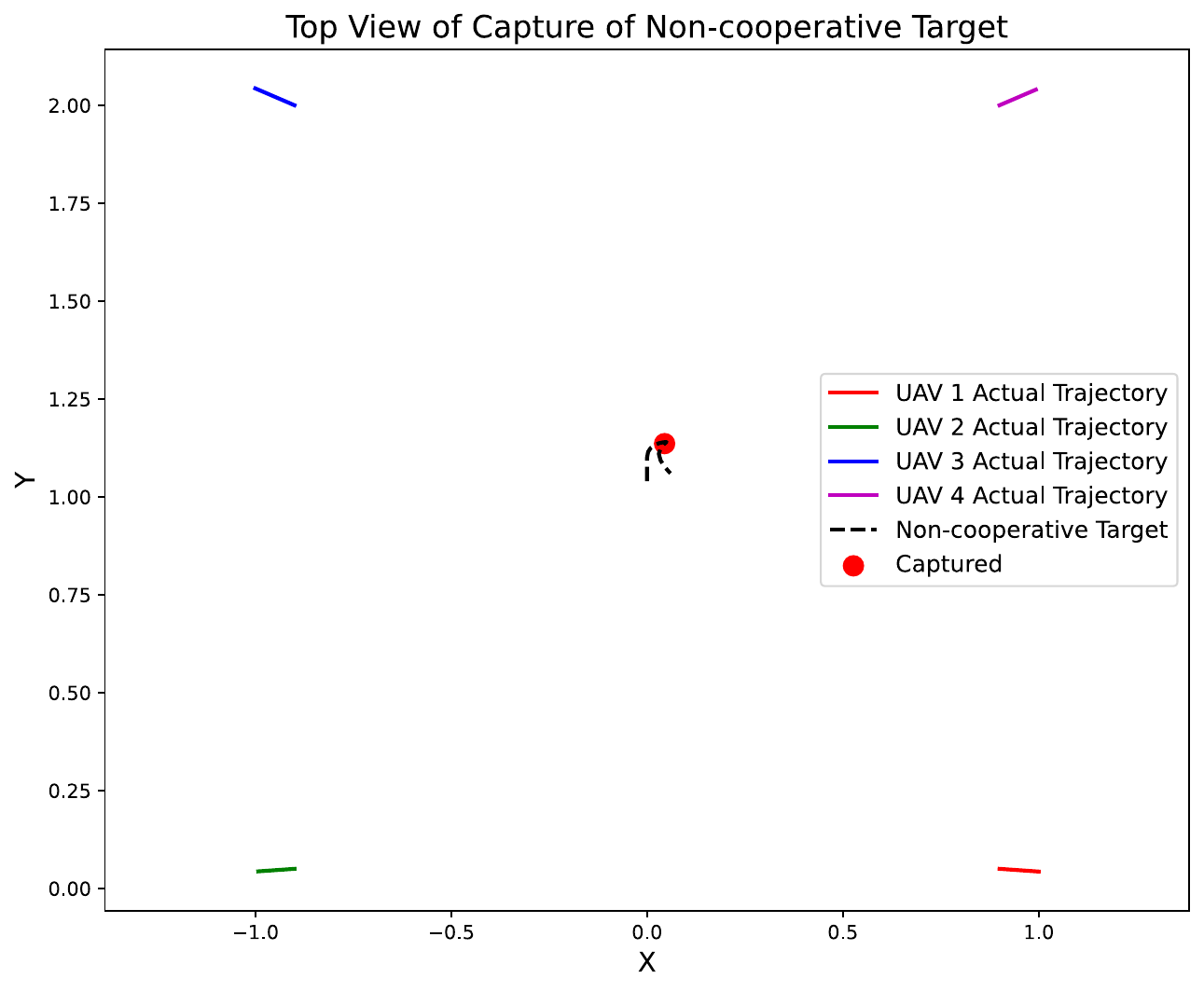}}
\hfill
\subfloat[\centering]{\includegraphics[width=5.5cm]{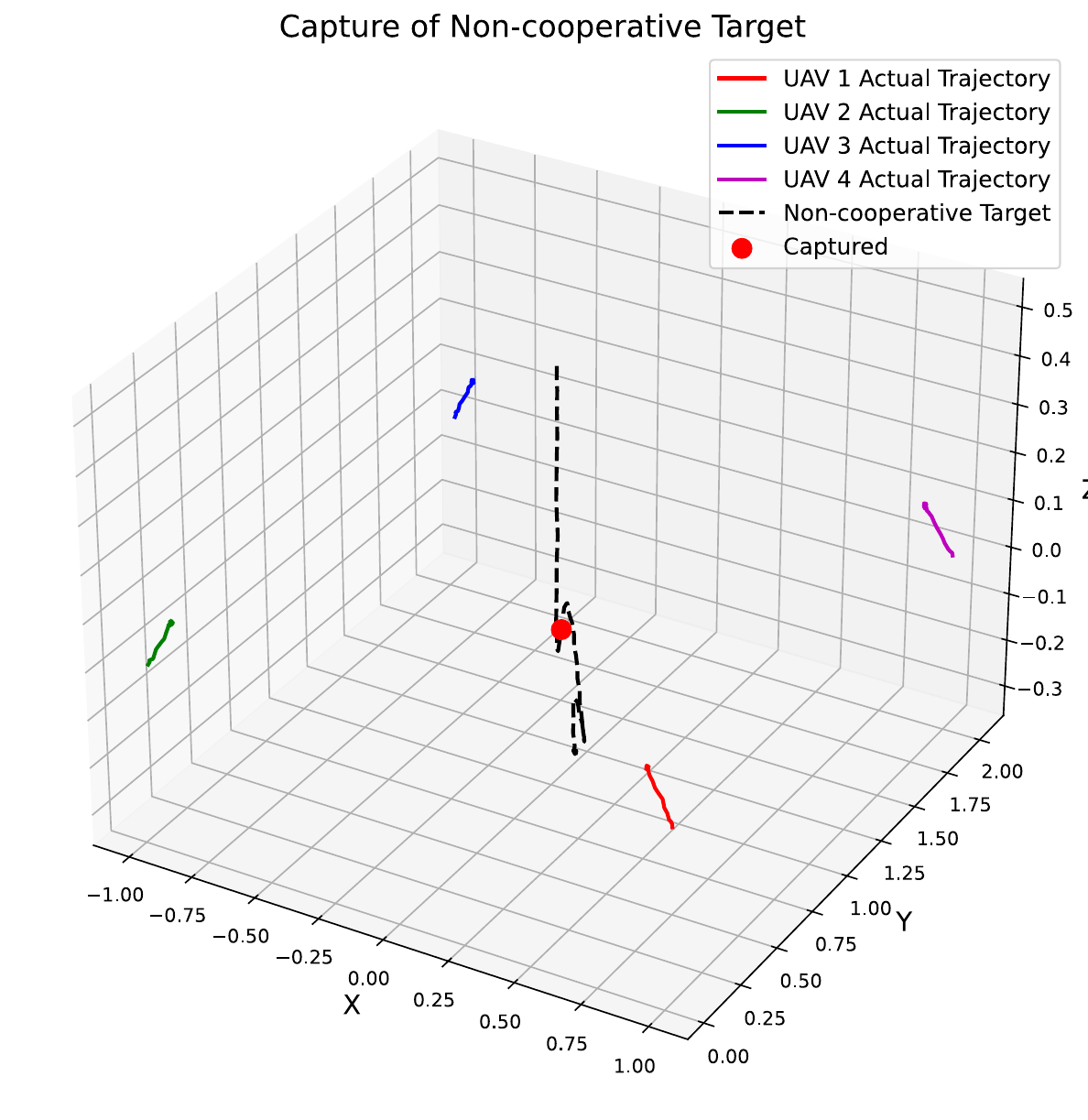}}
\caption{Schematic diagram and visualization of capturing a non-cooperative target. 
(\textbf{a}) is schematic diagram of the UAVs deploying the system to capture a free-falling target. 
(\textbf{b}) presents the visualized result from the simulation. 
(\textbf{c}) displays a top view of the UAVs’ and target’s trajectories. 
(\textbf{d}) shows the side-view position trajectories of the UAVs and target during the capture process.
\label{fig:results_capture_passive}}
\end{figure}

\begin{figure}
\centering
\subfloat[\centering]{\includegraphics[width=7cm]{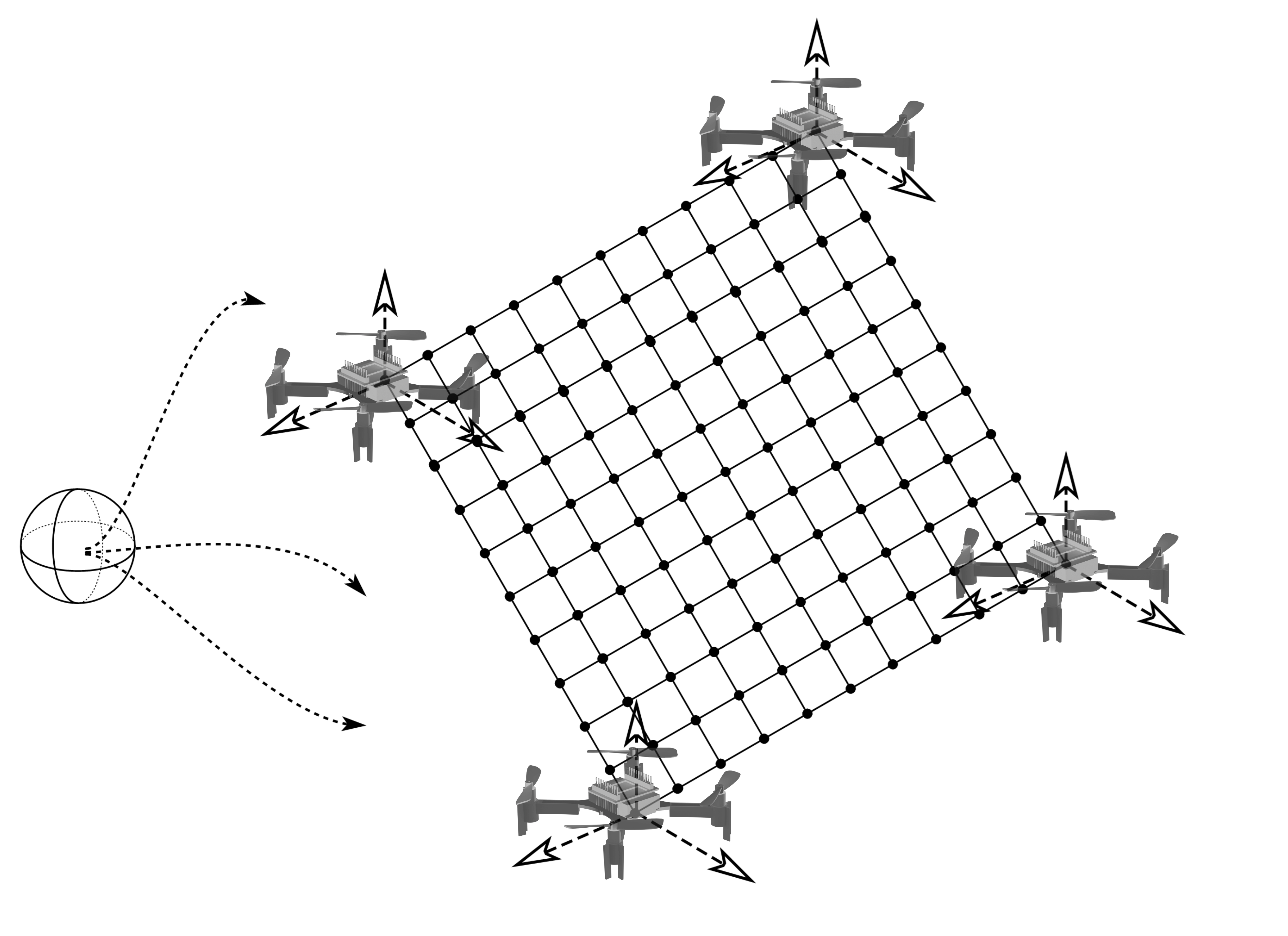}}
\hfill
\subfloat[\centering]{\includegraphics[width=6cm]{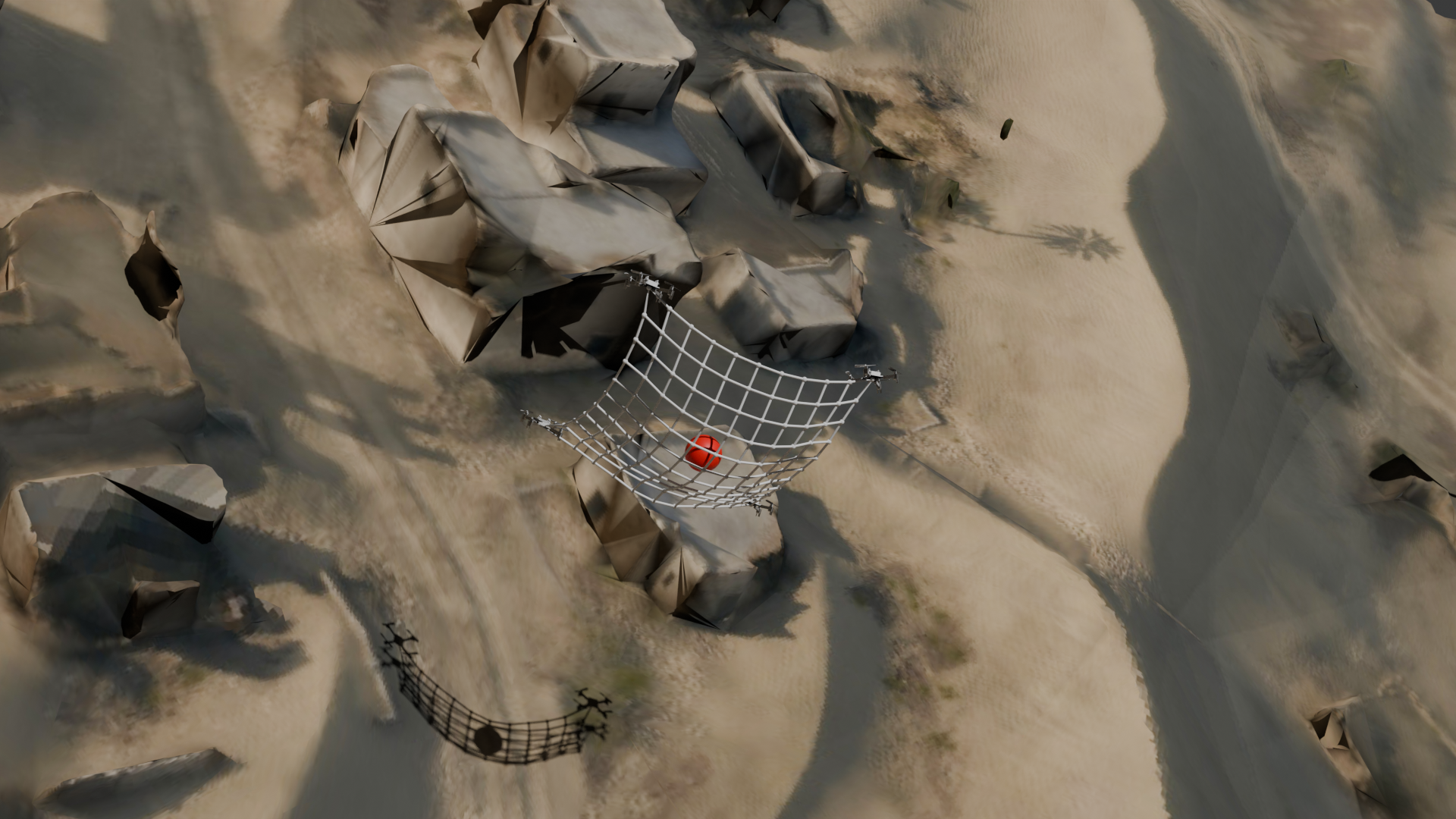}}\\
\subfloat[\centering]{\includegraphics[width=7cm]{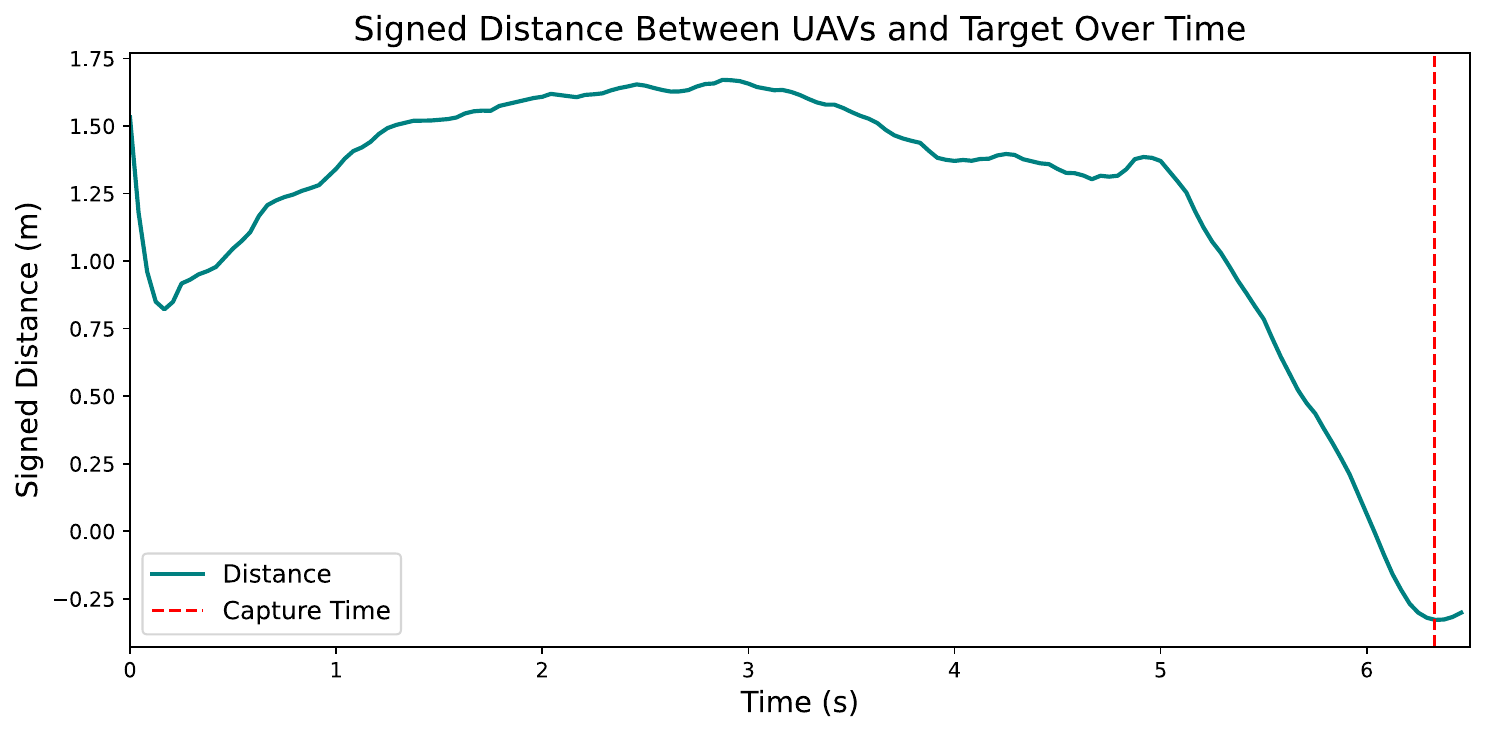}}
\hfill
\subfloat[\centering]{\includegraphics[width=6cm]{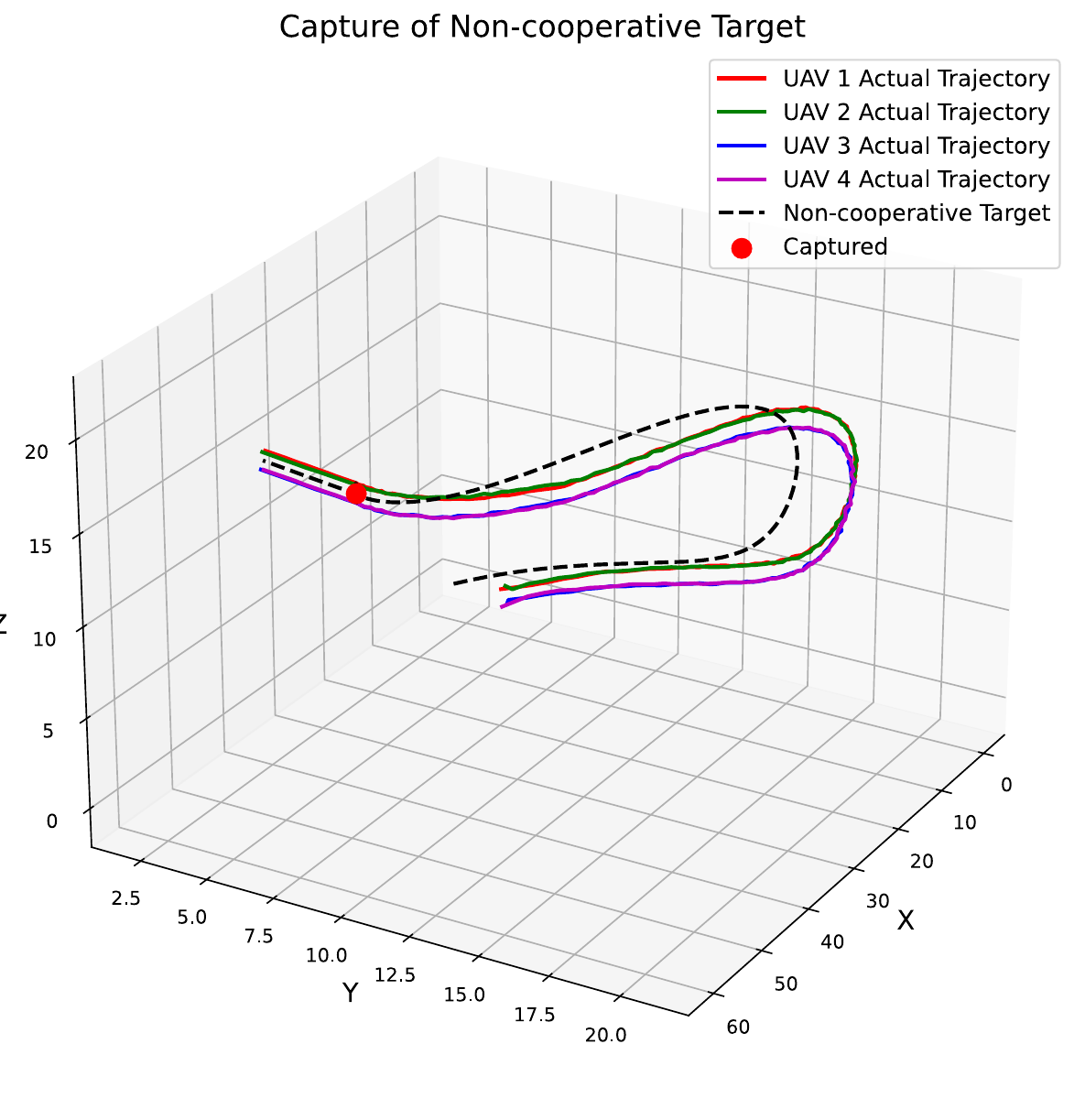}}
\caption{Schematic diagram and visualization of capturing a propelled target. 
(\textbf{a}) is schematic diagram of the UAVs deploying the system to capture a propelled target.  
(\textbf{b}) presents the visualized result from the simulation. 
(\textbf{c}) illustrates the time-varying distances between UAVs and the target. 
(\textbf{d}) shows the full spatial trajectories of the UAVs and the target from a side perspective.
\label{fig:results_capture_active}}
\end{figure}

The first scenario examines a free-falling target without propulsion. Fig.~\ref{fig:results_capture_passive} shows the simulation results. In (\textbf{a}), UAVs deploy the system and coordinate their motion in response to the target’s descent. The corresponding capture process is visualized in (\textbf{b}), where the UAVs surround and capture the target effectively. (\textbf{c}) provides a top view of UAVs and target trajectories. (\textbf{d}) shows the spatial trajectory evolution from another perspective. These results confirm that when the target motion is predictable, the control policy can generate synchronized UAV behavior to form an effective system for capture.

The second case involves a more complex and uncertain setting, where the target exhibits actively driven, unpredictable motion. The system response is shown in Fig.~\ref{fig:results_capture_active}. (\textbf{a}) illustrates the schematic diagram of UAVs adapting to the target’s evasive maneuvers. The capture is visualized in (\textbf{b}). (\textbf{c}) quantifies the UAVs’ distance to the target over time, reflecting how the UAVs dynamically regulate proximity despite disturbances. The trajectory plot in (\textbf{d}) further emphasizes the complex motion and real-time coordination needed for successful capture. The results highlight the robustness of the control strategy under partial observability and uncertain dynamics.

In summary, these two cases demonstrate the generality and effectiveness of the proposed multi-UAV-tethered netted system. Whether dealing with non-propelled or propelled targets, the integrated perception and learning-based control modules enable coordinated and reliable aerial capture, validating the full pipeline across diverse conditions.

\section{Discussion}\label{sec:main_dis}

The results presented in Sect.~\ref{sec:main_res} validate the feasibility and effectiveness of the multi-UAV-tethered netted system in capturing non-cooperative targets within the proposed multibody dynamics-based simulation environment (mySim). The simulation results confirm the system's capability to accurately model the dynamics of both the UAVs and the tethered nets, and the proposed perception and control strategies demonstrate their potential in UAV-based autonomous target tracking and capture. This success in simulation highlights the applicability of the multi-UAV-tethered netted system for real-world use cases, especially in non-cooperative target capture scenarios.

In comparison to existing UAV simulators based on conventional physics engines, mySim provides a more accurate and sophisticated physics simulation built upon multibody dynamics (MBD). This allows for the effective modeling of towed flexible nets, UAV dynamics, and collisions, which are inherently difficult to simulate accurately using standard physics engines that rely on rigid body or soft body approximations. By leveraging the capabilities of MBD, mySim overcomes the limitations typically encountered in other simulation environments, delivering more precise and realistic representations of UAV motion and net interactions.

Additionally, mySim stands out from commercial multibody dynamics software by integrating a complete perception module. This feature allows for seamless integration with vision-based odometry algorithms, enabling accurate tracking and mapping of the UAV’s environment and target trajectories. The system also supports reinforcement learning (RL) integration, providing a robust platform for testing and optimizing UAV control policies in complex mission scenarios. The ability to test control policies and strategies within mySim before real-world deployment offers a significant advantage in designing more proactive and adaptable UAV systems.

Given the successes and strengths demonstrated by mySim, future work could involve exploring other UAV-based systems or expanding the current system to tackle even more complex tasks and environments.

In conclusion, UAV can evolve into a more powerful and versatile tool for advancing multi-UAV collaborative systems and autonomous tasks. The successful design and implementation of the multi-UAV-tethered netted system to capture non-cooperative targets represents a significant step toward deploying these systems in real-world applications. With improvements and maintenance, mySim has the potential to become a foundational platform for designing, simulating and testing UAV-based systems in a wide range of autonomous operations.

\section*{Acknowledgments}
This was was supported in part by...... UNDERCONSTRUCTION

\clearpage
\appendix

\numberwithin{equation}{section}
\setcounter{figure}{0}
\setcounter{table}{0}

\setcounter{algocf}{0}

\renewcommand{\thesection}{\Alph{section}}
\renewcommand{\theequation}{\Alph{section}\arabic{equation}} 
\renewcommand{\thefigure}{\Alph{section}\arabic{figure}} 
\renewcommand{\thetable}{\Alph{section}\arabic{table}} 

\renewcommand{\thealgocf}{\Alph{section}\arabic{algocf}}

\section{Introduction to Marker Technique in Multibody Dynamics}

In multibody dynamics system, adjacent bodies are connected through constraints. The Marker technique allows for the connection between the constraint library and the rigid body dynamics module. A Marker is a local coordinate system attached to a specific rigid body component. By introducing the Marker technique, modules such as the dynamics equation library, constraint library, and force library can be linked with other modules. The constraint library only needs to establish the relationship between the physical quantities of $\mathrm{Marker}_I$ and $\mathrm{Marker}_J$.

For the two types of geometric constraints used in this paper, they can be represented using position vectors and frame matrices. Let:

\begin{linenomath*} 
\begin{equation}\label{eq:A_I_A_J}
    \begin{aligned}
          \mathbf A_I&=\left[x_I,y_I,z_I\right]\\
          \mathbf A_J&=\left[x_J,y_J,z_J\right]
    \end{aligned}
\end{equation}
\end{linenomath*}

For the fixed joint used in Fig.~\ref{fig:camera_fixed}, where there is no relative motion between $\mathrm{Marker}_I$ and $\mathrm{Marker}_J$, the constraint equation is:

\begin{linenomath*} 
\begin{equation}\label{eq:appendix_fixed}
    \begin{aligned}
          \mathbf r_I-\mathbf r_J&=0\\
          x_I\cdot y_J&=0\\
          x_I\cdot z_J&=0\\
          y_I\cdot z_J&=0
    \end{aligned}
\end{equation}
\end{linenomath*}

For the universal joint used in Fig.~\ref{fig:camera_universal}, where $\mathrm{Marker}_I$ and $\mathrm{Marker}_J$ are positioned identically and the Z-axes of the two Markers remain perpendicular to each other, the constraint equation is:
\begin{linenomath*} 
\begin{equation}\label{eq:appendix_universal}
    \begin{aligned}
          \mathbf r_I-\mathbf r_J&=0\\
          z_I\cdot z_J&=0
    \end{aligned}
\end{equation}
\end{linenomath*}

The constraint equations generated using the Marker technique, as shown in Fig.~\ref{fig:multibody_marker}, can be utilized to assemble the dynamics equations of the multibody system.

\section{Algorithms \& Figures}

\begin{algorithm}
\caption{Import Objects and Animate from CSV in Blender}\label{alg:blender_rendering}
\SetAlgoLined
\KwIn{CSV file with timestamps, positions, and rotations\newline
USD files for objects\newline
Blender scene frame rate}
\KwOut{Blender scene updated with animated objects}
Read the CSV file and determine the number of objects\;
Import USD files for each object and store them in a list\;
\ForEach{row in the CSV}{
    Calculate the frame number from the timestamp\;
    \ForEach{object}{
        Update its position and rotation based on the CSV data\;
        Insert keyframes for position and rotation\;
    }
}
Set the animation start and end frames\;
Print completion message\;
\end{algorithm}
\begin{algorithm}
\caption{UAV Pose Estimation Process}\label{alg:drone_state_estimation_process}
\KwIn{Scene Camera Image Sequence\newline 
Simulated IMU Signal\newline 
Visual-Inertial Odometry Algorithm}
\KwOut{Estimated System State}

\SetKwInOut{Input}{Input}
\SetKwInOut{Output}{Output}

\SetKwProg{Fn}{Function}{}{}

\SetKwFunction{FMain}{UAVPoseEstimation}
\Fn{\FMain{}}{
    \textbf{Initialize the VIO system}\;
    Camera intrinsic and extrinsic parameters\;
    IMU calibration parameters\;
    
    \ForEach{frame in the Scene Camera Image Sequence}{
        Fetch corresponding IMU data\;
        Pass the camera frame and IMU data to the VIO system\;
        Estimate the UAV's position and orientation using the VIO algorithm\;
    }
    Store the estimated position and orientation for each frame\;
    \Return{Estimated System State}\;
}
\end{algorithm}
\begin{algorithm}
\caption{Target Tracking Process}\label{alg:target_tracking_process}
\KwIn{Target Camera Image Sequence\newline
Initial Camera Position and Orientation\newline
Target Detection Model\newline
Camera Controller Parameters}
\KwOut{Estimated Target State}

\SetKwInOut{Input}{Input}
\SetKwInOut{Output}{Output}

\SetKwProg{Fn}{Function}{}{}

\SetKwFunction{FMain}{TargetTracking}
\Fn{\FMain{}}{
    Initialize camera position and orientation\;
    
    \ForEach{frame in the Target Camera Image Sequence}{
        Pass the frame to the Target Detection Model\;
        Extract the target's pixel coordinates\;
        \If{target is detected}{
            Compute the error between the target's position and the image center\;
            Adjust the camera's pitch and yaw using the Camera Controller\;
            Update the camera's position and orientation\;
            Render the new frame with the Rotatable Camera\;
        }
        \If{target is not detected}{
            Log the frame as "target lost" or implement recovery logic\;
        }
    }
    Estimate the target's position and orientation in world coordinates\;
    \Return{Estimated Target State}\;
}
\end{algorithm}
\begin{algorithm}
\caption{Multi-UAV Cooperative Capture System using MAPPO}\label{alg:controller_process_mappo}
\KwIn{Target State, System States of all UAVs, Trained MAPPO policy network, Environment simulator (mySim)}
\KwOut{Final system state, Performance metrics}

\SetKwInOut{Input}{Input}
\SetKwInOut{Output}{Output}

\SetKwProg{Fn}{Function}{}{}

\SetKwFunction{FMain}{MultiUAVCooperativeCapture}
\Fn{\FMain{}}{
    Initialize MAPPO controller with shared policy network and centralized critic\;
    
    Set up the environment with initial states of UAVs and the target\;
    
    \While{simulation is running}{
        \ForEach{UAV (agent i $\in$ {1, ..., n})}{
            Observe local state\;
            Pass the observation to the shared policy network to compute an action\;
        }
        Apply all UAVs’ actions to the environment simulator\;
        Update environment state\;
        Compute and log cooperative rewards\;
        Check for termination conditions\;
    }
    Render the updated state for visualization\;
    Log performance metrics\;
    \Return{Final system state, Performance metrics}\;
}
\end{algorithm}
\begin{algorithm}
\caption{Controller Process}\label{alg:controller_process_pd}
\KwIn{Target State\newline
System State\newline
PID Controller Parameters\newline
Multibody Dynamic Model of the UAV}
\KwOut{Control Signals\newline Updated UAV State}

\SetKwInOut{Input}{Input}
\SetKwInOut{Output}{Output}

\SetKwProg{Fn}{Function}{}{}

\SetKwFunction{FMain}{ControllerProcess}
\Fn{\FMain{}}{
    \textbf{Initialize the controller}\;
    Target State, System State, Control parameters\;
    
    \ForEach{time step}{
        Compute the error between the Target State and System State\;
        \If{Position Error and Orientation Error are within acceptable thresholds}{
            Mark the system as stable\;
            Log the current state as "target reached" or "hovering"\;
        }
        Pass the control signals to the Dynamic Physics Model\;
        Simulate the updated state\;
    }
    \Return{Control Signals, Updated UAV State}\;
}
\end{algorithm}

\begin{figure}[h]
    \centering
                \includegraphics[width=10cm]{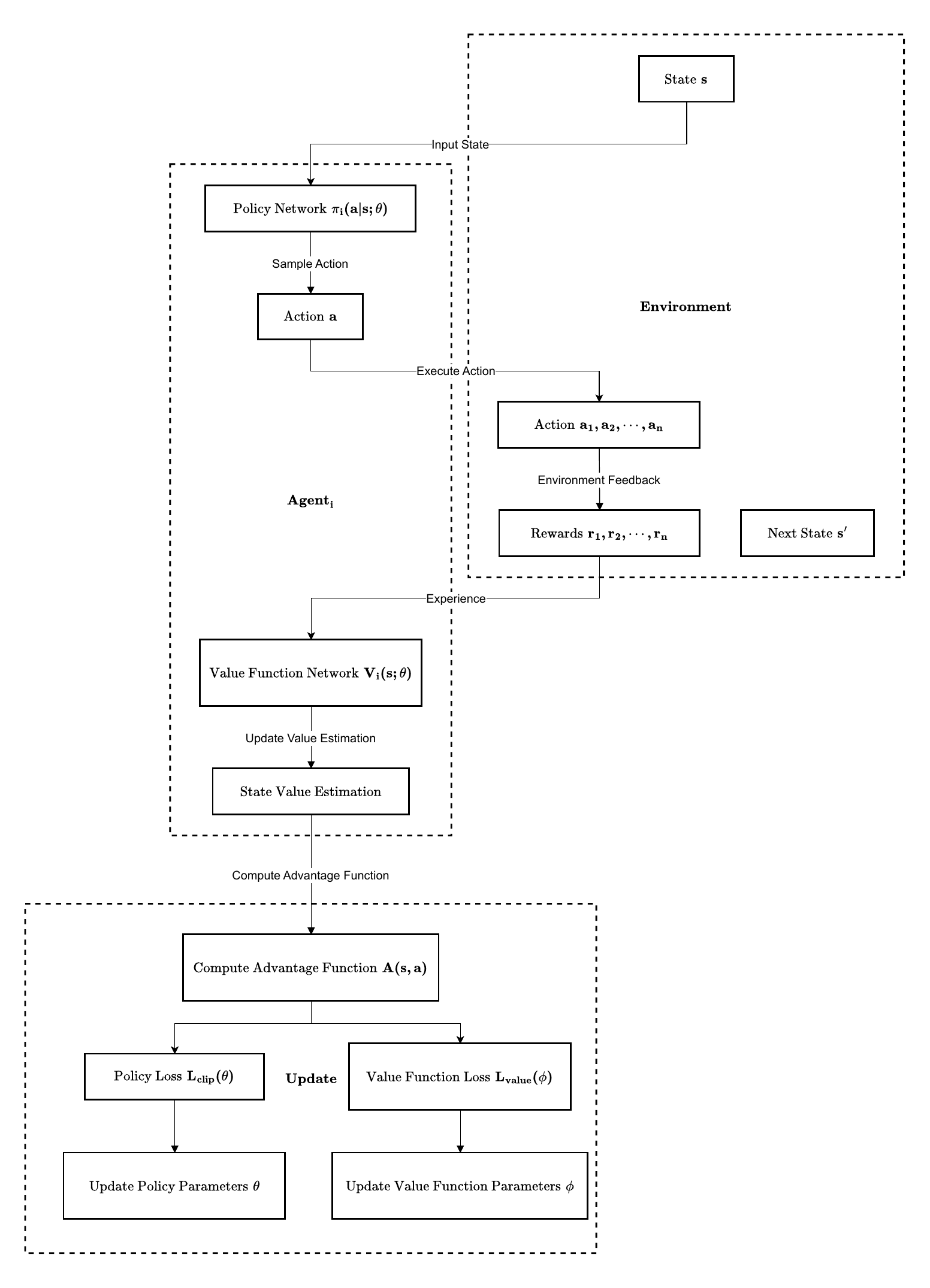}
    \caption{Schematic diagram of the MAPPO algorithm, which optimizes the strategy of each UAV to enable efficient collaboration in a multi-agent environment.}
    \label{fig:ippo}
\end{figure}

%Bibliography
\clearpage
\bibliographystyle{unsrt}  
% \bibliography{references}  

% \reftitle{References}
\bibliography{refs/ref}

\end{document}